\documentclass[preprint,12pt,3p]{elsarticle}
\usepackage[colorlinks,citecolor=red,urlcolor=blue,bookmarks=false,hypertexnames=true]{hyperref} 

\usepackage{graphicx}
\usepackage{caption}
\usepackage{subcaption}
\usepackage{psfrag}
\usepackage{array}
\usepackage{multirow}
\usepackage{comment}
\usepackage{amsmath}%
\newcommand{\norm}[1]{\left\lVert#1\right\rVert}
\DeclareMathOperator*{\argmin}{argmin}
\usepackage{xcolor}
\usepackage{amssymb}
\usepackage{amsfonts}%
\usepackage{amssymb}%%
\usepackage{algorithm}
\usepackage{algpseudocode}
\usepackage{setspace}
\usepackage{setspace}
\usepackage{textcomp}
\usepackage[switch, running]{lineno}

\journal{Advances in Water Resources}

\begin{document}
\sloppy
\begin{frontmatter}

\title{Deep-learning-based coupled flow-geomechanics surrogate model for CO$_2$ sequestration}
% \tnotetext[label0]{This is only an example}

\author[label1,label5]{Meng Tang\corref{cor1}}%\fnref{label3}}
\address[label1]{367 Panama Street, Stanford, CA, 94305}
% \address[label2]{Address Two\fnref{label4}}
\address[label5]{Department of Energy Resources Engineering, Stanford University}
\address[label4]{Atmospheric, Earth, and Energy Division, Lawrence Livermore National Laboratory, Livermore, CA, 94550}

\cortext[cor1]{Corresponding author}
% \fntext[label3]{I also want to inform about\ldots}
% \fntext[label4]{Small city}

\ead{mengtang@stanford.edu}
% \ead[url]{author-one-homepage.com}

\author[label4]{Xin Ju}
\ead{ju1@llnl.gov}

\author[label1,label5]{Louis J. Durlofsky}
\ead{lou@stanford.edu}

% abstraction
\begin{abstract}

A deep-learning-based surrogate model capable of predicting flow and geomechanical responses in CO$_2$ storage operations is presented and applied. The 3D recurrent R-U-Net model combines deep convolutional and recurrent neural networks to capture the spatial distribution and temporal evolution of saturation, pressure and surface displacement fields. The method is trained using high-fidelity simulation results for 2000 storage-aquifer realizations characterized by multi-Gaussian porosity and log-permeability fields. These numerical solutions are expensive because the domain that must be considered for the coupled problem includes not only the storage aquifer but also a surrounding region, overburden and bedrock. The surrogate model is trained to predict the 3D CO$_2$ saturation and pressure fields in the storage aquifer, and 2D displacement maps at the Earth's surface. Detailed comparisons between surrogate model and full-order simulation results for new (test-case) storage-aquifer realizations are presented. The saturation, pressure and surface displacement fields provided by the surrogate model display a high degree of accuracy, both for individual test-case realizations and for ensemble statistics. Finally, the the recurrent R-U-Net surrogate model is applied with a rejection sampling procedure for data assimilation. Although the observations consist of only a small number of surface displacement measurements, significant uncertainty reduction in pressure buildup at the top of the storage aquifer (caprock) is achieved.

\end{abstract}

\begin{keyword}
carbon storage, coupled flow and geomechanics, surrogate model, deep learning, data assimilation, history matching
\end{keyword}
\end{frontmatter}
%Introduction
\section{Introduction}
\label{sect:intro}

Carbon capture and storage (CCS) is expected to play an important role in reducing greenhouse gas emissions to the atmosphere. In CCS, supercritical CO$_2$ is injected into deep brine aquifers or abandoned oil reservoirs for permanent storage. The prediction of the geomechanical response of the formation is very important for the management of CCS operations because excessive pressure buildup can induce fracturing of the caprock, or activate pre-existing faults, through which CO$_2$ or brine can leak. The numerical simulation of coupled flow and geomechanics problems can be computationally prohibitive, however. This leads to significant challenges in the practical application of data assimilation (history matching) and uncertainty quantification, both of which require large numbers of forward simulations. Effective surrogate models for CO$_2$ storage will therefore be very useful in these settings.

Our goal in this work is to develop a deep-learning-based surrogate model to treat coupled flow and geomechanics in carbon storage operations. The resulting model will then be applied for data assimilation, using surface displacement data, to estimate aquifer properties and thus reduce uncertainty in pressure buildup at the caprock. The 3D surrogate model, which is trained to predict pressure and CO$_2$ saturation throughout the storage aquifer along with surface (ground-level) vertical deformation, represents an extension of our recently developed 2D and 3D surrogate models for two-phase subsurface flow in the absence of geomechanical effects \citep{tang2020jcp,tang2021deep}.

A number of investigators have developed and applied numerical models for coupled flow and geomechanics in carbon storage operations \citep{vilarrasa2010coupled, shi2013coupled, talebian2013computational, li2016coupled, fuchs2019geochemical,futhermo,ju2021simple}. Vilarrasa et al.~\citep{vilarrasa2010coupled} modeled an axisymmetric  reservoir-caprock system using hydromechanical coupling to evaluate the probability of reactivation or the creation of fractures, while Eshiet and Sheng~\citep{eshiet2014investigation} introduced a discrete element method for the coupled problem. Shi et al.~\citep{shi2013coupled} conducted a simulation study to match InSAR surface uplift data for the In Salah (Algeria) CO$_2$ storage project. Li and Laloui~\citep{li2016coupled} also studied the In Salah project, using a model that included flow, geomechanical and thermal effects. Their goal was to predict surface deformation and assess the effects of Biot's coefficient and temperature.  Additional studies involving coupled flow and geomechanics in CO$_2$ storage settings were reviewed by Rutqvist et al.~\citep{rutqvist2019numerical}. In this work we apply a computational framework that couples multiphase flow and geomechanics to model poroelastic rock response during CO$_2$ injection. The numerical approach entails a sequential-implicit fixed-stress scheme to couple a multiphase flow module and a mechanics module \citep{futhermo, kim2011stability}.

Surrogate models, also referred to as proxy models, can be used to replace computationally expensive forward models in problems requiring many related simulation runs. Example application areas include data assimilation, optimization, and uncertainty quantification. Surrogate models can be classified into simplified physical/numerical models and statistical (e.g., deep learning) methods. Reduced-order numerical models based on proper orthogonal decomposition (POD), for example, have been developed for problems involving coupled flow and geomechanics by Florez and Gildin~\citep{florez2019model} and Jin et al.~\citep{jin2020reduced}. Neither of these studies, however, was performed in the context of geological carbon storage.

Deep-learning-based procedures approximate the input-output relationship of simulation data using statistical tools. Tang et al.~\citep{tang2020jcp,tang2021deep} developed recurrent R-U-Net surrogate models, for 2D and 3D two-phase subsurface flow problems, which combine a residual U-Net with convLSTM networks to predict the evolution of the saturation and pressure fields. A residual U-Net was additionally used, in an autoregressive strategy, to predict flow response in 2D problems with varying well-control specifications \citep{jiang2021deep}. Deep-learning-based surrogate models have also been implemented for CO$_2$ storage problems. Mo et al.~\citep{mo2019deep} developed a deep neural network framework to forecast CO$_2$ plume movement for random 2D permeability fields, while Wen et al.~\citep{wen2021towards} devised a general framework for predicting CO$_2$ plume location in axisymmetric (single-well) scenarios. To our knowledge, the development of a deep-learning-based surrogate for CO$_2$ storage with coupled flow and geomechanics, which is the goal of this work, has yet to be accomplished.

%and there have been many studies along these lines. Cameron et al.~\citep{cameron2016use}, for example, used pressure observations in the overlying formation to predict the size and location of potential leaks in the caprock. Sun et al.~\citep{sun2019co2data} developed a data-space inversion method to assimilate well monitoring data (pressure and CO$_2$ saturation) and forecast uncertainty associated with CO$_2$ plum locations in the top layer of storage reservoir. Furthermore, utilizing bottom hole pressure (BHP) of the injection wells, CO$_2$ saturation at monitoring wells and and seismic data, ensemble data assimilation method was applied for characterization of CO$_2$ storage reservoirs in \citep{liu2020petrophysical}.

The prediction of the pressure field and plume location, and the identification of leaks, is important for the assessment of risk in CO$_2$ storage operations. Uncertainty in these essential quantities can be reduced through use of surveillance/monitoring data in conjunction with data assimilation procedures. Several studies \citep{gonzalez2015detection,jung2015detection,cameron2016use}, for example, have used pressure monitoring data for leak detection. Chen et al.~\citep{chen2020reducing} used pressure and CO$_2$ saturation data at monitoring wells to reduce uncertainty in plume location. Seismic data can also be incorporated for storage-aquifer characterization \citep{liu2020petrophysical}. The models used in these papers did not include coupled flow and geomechanics. There have, however, also been data assimilation studies for CO$_2$ storage that included geomechanical effects. Guzman et al.~\citep{de2014coupled} considered compartmentalized aquifers and estimated fault transmissibility and near-well permeability using injection well pressure and surface deformation data. Jahandideh et al.~\citep{jahandideh2021inference} used microseismic data to estimate rock properties including permeability and Young's modulus. Both of these studies took properties to be constant within a region/compartment, and both applied ensemble-based methods for the inversion.

%For example, data assimilation of monitoring pressure data near caprock was applied in both \citep{gonzalez2015detection} and \citep{jung2015detection} for detection of brine leaks in geological carbon storage. Pressure and CO$_2$ saturation at monitoring wells are history matched to reduce uncertainties of plume evolution in \citep{chen2020reducing}.  

A number of researchers have addressed data assimilation in other types of coupled flow-geomechanics systems. Wilschut et al.~\citep{wilschut2011joint} used surface subsidence and production data, with an ensemble Kalman filter method, to estimate fault transmissibilities in a pseudo-3D gas-bearing compartmentalized reservoir. Similarly, Jha et al.~\citep{jha2015reservoir} applied an ensemble smoother (ES) to estimate rock properties in a 3D homogeneous gas reservoir by assimilating both surface deformation and well data. ES was also applied by Zoccarato et al.~\citep{zoccarato2016data} to estimate geomechanical parameters such as Poisson's ratio in a 3D homogeneous gas storage model using surface displacement data. Tang~\citep{tang2018history} applied both derivative-free optimization and an ES procedure to invert for (layered) permeability and Young's modulus in 2D models using well production data and surface displacement data. These studies, however, only involved the estimation of a limited number of parameters in homogeneous or layered models; i.e., detailed rock property fields were not considered. In addition, the data assimilation algorithms used in the studies with 3D models were restricted to ensemble-based methods because of the high computational demands associated with the forward modeling. In recent work, Alghamdi et al.~\citep{alghamdi2020bayesian, alghamdi2021bayesian} developed an adjoint-gradient inversion procedure to determine permeability fields in groundwater-aquifer settings from surface deformation data. The permeability fields considered in these studies were areally heterogeneous (and characterized by Gaussian fields), but were assumed to have no variation in the vertical direction.

%[xxx1 -- need 2-3 concluding sentences at the end of the previous paragraph  pointing out the limitations of these studies (e.g., 2D rather than 3D? detailed perm fields not considered? Single-phase flow or oil-water rather than CO2 storage? Etc etc). Along these lines, add some text in the above paragraph to indicate the types of models considered in each study. E.g., ``xxx applied an ensemble smoother (ES) to estimate rock properties (permeability and porosity) in 2D layered models by assimilating ...'' And clarify if the application was CO2 storage or something else.]

In this study, we apply the 3D recurrent R-U-Net surrogate model developed in \citep{tang2021deep} to model coupled flow and geomechanics in CO$_2$ storage settings. The problem domain in the high-fidelity solution includes the storage aquifer, a surrounding region, overburden (which extends up to the Earth's surface), and bedrock. Extensive domains of this type are required for the proper modeling of geomechanical effects. The storage aquifer is characterized by random 3D multi-Gaussian permeability and porosity fields of prescribed correlation structure. The surrogate model is trained to predict the 3D saturation and pressure fields in the storage aquifer as well as the ground (surface-level) 2D vertical displacement field. We apply the overall framework to solve an inverse problem in which surface deformation data are used to predict the detailed permeability and porosity fields in the aquifer. A rigorous rejection sampling algorithm is applied. The resulting posterior models are then used to predict pressure buildup at the top of the storage reservoir.  

This paper proceeds as follows. In Section~\ref{sec:method}, we first present the coupled flow and geomechanics equations applicable for CO$_2$ sequestration. We then describe the recurrent R-U-Net surrogate model and associated data-processing and training procedures. In Section~\ref{sec:eval}, the 3D surrogate model is applied for coupled flow (CO$_2$-brine) and geomechanics problems. In the simulations, CO$_2$ injection is accomplished via four vertical wells. Our specific interest is in predicting solutions for new geomodels characterized by multi-Gaussian random fields. A detailed assessment of surrogate model accuracy is presented. Next, in Section~\ref{sec:hm}, the trained surrogate model is applied for data assimilation. We conclude in Section~\ref{sec:conl} with a summary and suggestions for future research directions.

\section{Governing Equations and Surrogate Model}
\label{sec:method}

In this section, we present the governing equations for coupled multiphase flow and geomechanics used to model CO$_2$ storage operations. The general simulation setup and solution variables are then discussed. Next, the recurrent R-U-Net surrogate model for the coupled system, along with the training process and data preprocessing, are described. 

%subsection 1
\subsection{Governing equations for coupled problem}
\label{sect:goveqns}

In the formulation for the coupled problem, the multiphase fluid and solid skeletons are treated as overlapping continua. The governing equations for the flow problem involve statements of mass conservation for water and CO$_2$ components, which flow in two phases, referred to as the aqueous and gaseous phases. The conservation equations are expressed as:
\begin{equation} 
    \nabla \cdot \left( \sum_{j}\rho_j x_j^r \mathbf{v}_j \right) + q^r = \frac{\partial}{\partial{t}}\left(\sum_{j}\phi \rho_j S_j x_j^r \right),
\label{eq:fluid mass balance}
\end{equation}
where subscript $j=a$ indicates the aqueous phase and $j = g$ the gaseous phase, superscript $r = w$ denotes the water component and $r = c$ the CO$_2$ component, $\rho_j$ is phase density, $x_j^r$ is the mass fraction of component $r$ in phase $j$, $\mathbf{v}_j$ is the Darcy velocity (given below), $q^r$ represents the source/sink term, $t$ is time, $\phi$ is true porosity (defined as the ratio of the pore volume to the bulk volume in the deformed configuration), and $S_j$ is the phase saturation. 

The Darcy velocity for phase $j$, $\mathbf{v}_j$, is given by
\begin{equation} 
     \mathbf{v}_j =  -\frac{\mathbf{k} k_{rj}(S_j)}{\mu_j} \left(\nabla p_j - \rho_j g \nabla z \right),
\label{eq:darcy-eq}
\end{equation}
where $\mathbf{k}$ is the absolute permeability tensor, $k_{rj}$, $\mu_j$ and $p_j$ denote relative permeability, viscosity and fluid pressure of phase $j$, respectively, $g$ is gravitational acceleration, and $z$ is depth. The phase pressures are related to each other through the capillary pressure $p_c$; i.e., $p_c = p_g - p_a$, with $p_c(S_g)$ a prescribed function. The system is closed by enforcing $S_g+S_a=1$.

The governing geomechanical equations are derived from the linear-momentum balance in the matrix, expressed as
\begin{equation}
     \nabla \cdot \boldsymbol \sigma + \rho_m g \nabla z = 0.
\label{eq:linear-moment balance}
\end{equation}
Here $\boldsymbol \sigma$ denotes the Cauchy stress tensor and the composite matrix density $\rho_m$ is a volume weighted average of phase densities $\rho_j$ and rock density~$\rho_{\text{rock}}$, i.e., 
\begin{equation}
    \rho_m = \phi \sum_{j}S_j \rho_j + (1 - \phi) \rho_{\text{rock}}.
\end{equation}
From poroelasticity theory \citep{biot1941general} and the assumption of linearly elastic deformation, $\boldsymbol\sigma$ can be related to fluid pressure and solid displacement via:
\begin{equation}
    \boldsymbol \sigma = \mathbf{C}:\nabla \mathbf{d} - b p_{e} \mathbf{1},
\label{eq:constitutive equation}
\end{equation}
where $\mathbf{C}$ is a fourth-order stiffness tensor for the solid skeleton (associated with the drained-isothermal elastic moduli), $\mathbf{d}$ denotes the solid displacement vector, $b$ is Biot's coefficient, $p_e$ denotes the equivalent pore pressure derived from $p_e = \sum_{j}S_j p_j$ \citep{coussy2004poromechanics}, and $\mathbf{1}$ is the second-order identity tensor.

The fluid mass balance Eq.~\ref{eq:fluid mass balance} and geomechanics momentum conservation Eq.~\ref{eq:linear-moment balance} are tightly coupled through poromechanics \citep{biot1941general, coussy2004poromechanics}. This coupled flow and geomechanics system is solved using a fixed-stress sequential implicit procedure \citep{kim2011stability}, which has been proven to yield accurate results with unconditional stability. In the sequential implicit scheme, the multiphase flow and geomechanical problems are solved iteratively, by updating the true porosity~$\phi$ from its previous state through application of:
\begin{equation}
    d\phi = \frac{b - \phi}{K_{dr}}(dp_e + d\sigma_v),
\end{equation}
where $\sigma_v$ denotes volumetric total stress and $K_{dr}$ is the drained-isothermal bulk modulus.

The flow equations are discretized using a finite volume formulation, with solutions computed for each finite volume block/cell. The geomechanical equations are discretized using a Galerkin finite element formulation. The numerical implementation of the fixed-stress iterative scheme follows the description in \citep{futhermo}. In this work, all high-fidelity simulations (HFS), by which we mean the numerical solution of Eqs.~\ref{eq:fluid mass balance}, \ref{eq:linear-moment balance} and \ref{eq:constitutive equation}, are performed using GEOS, a high-performance computing simulation environment for geoscience applications \citep{settgast2017fully,ju2020gas}.

\subsection{Variables and domains in HFS and surrogate model}
\label{sec:domains}

We can represent the HFS described above as
\begin{align}
    [\mathbf{p}_f, \mathbf{S}_f, \mathbf{d}_f] = f(\mathbf{m}_f),
    \label{eq-f}
\end{align}
where $f$ denotes the numerical simulation, $\mathbf{m}_f \in \mathbb{R}^{n_b}$ is the specified geomodel, and $\mathbf{p}_f \in \mathbb{R}^{n_b \times n_{ts}}$, $\mathbf{S}_f \in \mathbb{R}^{n_b \times n_{ts}}$ and $\mathbf{d}_f \in \mathbb{R}^{3 \times n_d \times n_{ts}}$ represent the dynamic gaseous-phase pressure, gaseous-phase saturation and displacement fields (the displacement field has a component in each coordinate direction). Here $n_b$ is the total number of finite volume grid blocks in the full-order overall model, $n_d$ is the total number of finite element nodes in the full-order model, and $n_{ts}$ is the number of time steps in the HFS. Note that $\mathbf{p}_f$, $\mathbf{S}_f$ and $\mathbf{d}_f$ are the primary variables computed in the HFS (i.e., all other dependent quantities can be determined from these variables). 

\begin{figure}[htbp]
  \centering
  \includegraphics[trim={0cm 0cm 0cm 0cm},clip, scale=0.75]{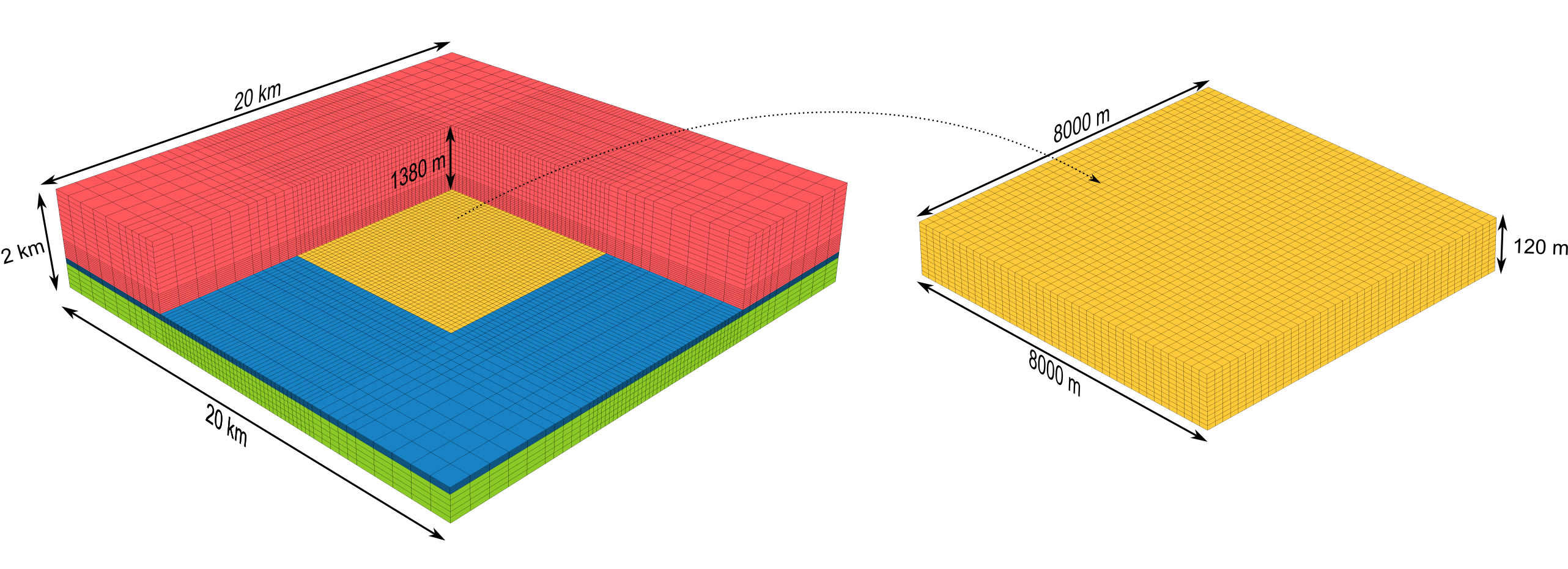}
   \caption{Illustration of the overall model (left) and storage aquifer (right). The yellow, red, blue and green regions denote the storage aquifer, overburden rock, surrounding domain, and bedrock, respectively. }
  \label{fig:reservoir-fig}
\end{figure}

In the numerical setup for coupled flow and geomechanics problems in carbon storage simulations, the overall model domain is much larger than the storage aquifer domain. This is necessary in order to correctly simulate the geomechanical response. The problem specification used in this study, shown in Fig.~\ref{fig:reservoir-fig}, illustrates the setup. Here the storage aquifer domain is of size $8000~\text{m} \times 8000~\text{m} \times 120~\text{m}$, and is represented by $40 \times 40 \times 12$ grid blocks. The full domain is much larger -- $20~\text{km} \times 20~\text{km} \times 2~\text{km}$, represented by $60 \times 60 \times 37$ grid blocks. The overburden rock (red region in Fig.~\ref{fig:reservoir-fig}), bedrock (green region), and the surrounding domain (blue region) are the additional domains required for the geomechanical solution.

When data assimilation is performed, flow simulations for a large number (e.g., $O(10^3-10^6$) of candidate geomodels $\mathbf{m}_f$ are required. In this work we consider the porosity and (isotropic) permeability fields in the storage aquifer to be uncertain, and we fix all other geomodel parameters. This includes porosity, permeability, Young's modulus and Poisson's ratio in the surrounding domain, overburden and bedrock, and Young's modulus and Poisson's ratio in the storage aquifer. These properties could also be varied if necessary, but we would then require more training simulations for the surrogate model. In addition, most of these properties are expected to have a relatively small impact on the key quantities of interest, such as plume location and pressure buildup in the storage aquifer.

Consistent with the discussion above, new storage-aquifer geomodels are denoted $\mathbf{m}_s \in \mathbb{R}^{n_s}$, and the corresponding saturation and pressure states are defined as $\mathbf{S}_s \in \mathbb{R}^{n_s\times n_{ts}}$ and $\mathbf{p}_s \in \mathbb{R}^{n_s\times n_{ts}}$, where $n_s$ indicates the number of grid blocks in the storage aquifer. Another quantity of interest is the vertical displacement map at the Earth's surface, directly above the storage aquifer. This is denoted as $\mathbf{d}_g \in \mathbb{R}^{n_g\times n_{ts}}$, where $n_g$ is the number of surface nodes aligned with (i.e., directly above) the storage aquifer. Although displacement is computed in the $x$, $y$ and $z$ directions, $\mathbf{d}_g$ only includes the vertical displacement, consistent with the data available from InSAR \citep{burgmann2000synthetic} measurements.

The storage aquifer geomodels $\mathbf{m}_s$ considered in this study are characterized by multi-Gaussian permeability and porosity fields. For a given geomodel, the permeability $\mathbf{k}_s \in \mathbb{R}^{n_s}$ and porosity $\boldsymbol \phi_s \in \mathbb{R}^{n_s}$ descriptions are derived from the same multi-Gaussian field $\mathbf{m}_s \in \mathbb{R}^{n_s}$. These fields are generated using the geomodeling tool SGeMS~\citep{remy2009applied}. For a particular grid block~$i$ in the storage aquifer, the isotropic permeability $(k_s)_i$ and porosity $(\phi_s)_i$ are given by $(k_s)_i = \exp (a (m_s)_i + b)$ and $(\phi_s)_i = c (m_s)_i + d$, where $a$, $b$, $c$ and $d$ are specified constants. Other modeling parameters are kept fixed, though some of these parameters are specified to have different values in different domains, as discussed in Section~\ref{sect:setup}.

%We have $n_d < n_r \ll n_b$, where $n_b$ is the total number of grid blocks in the coupled flow and geomechanics model. 

Despite the fact that the dynamic quantities of interest ($\mathbf{S}_s$, $\mathbf{p}_s$, $\mathbf{d}_g$) are restricted to the storage aquifer region and  the ground surface directly above the storage aquifer, the coupled problem must still be solved over the full domain (which leads to high computational expense). An advantage of the deep-learning-based surrogate model developed here is that it can be trained to predict only the quantities of interest in the domains of interest. This leads to savings in the time required for training and predictions.

Given a number of HFS training `samples' collected from the numerical simulator, a 3D recurrent R-U-Net is trained to serve as a surrogate. 
%to replace expensive numerical simulations~$f$ and provide prediction for dynamic state fields in any study regions at any selected time steps. 
We use $\hat{f}$ to denote the recurrent R-U-Net surrogate and $\hat{\mathbf{S}}_s \in \mathbb{R}^{n_s\times n_{t}}$, $\hat{\mathbf{p}}_s \in \mathbb{R}^{n_s\times n_{t}}$ and $\hat{\mathbf{d}}_g \in \mathbb{R}^{n_g\times n_{t}}$ to represent the surrogate-model predictions for gaseous-phase saturation, gaseous-phase pressure, and surface displacement. Note that these quantities are provided at $n_t$ time steps rather than at the $n_{ts}$ HFS time steps (with $n_t$ typically much less than $n_{ts}$), as the surrogate is applied to predict the states only at key time steps. The recurrent R-U-Net surrogate model is thus expressed as
\begin{align}
    % \mathbf{x} = f(\mathbf{m}, \mathbf{u}),
    [\hat{\mathbf{p}}_s, \hat{\mathbf{S}}_s, \hat{\mathbf{d}}_g] = \hat{f}(\mathbf{m}_s).
    \label{eq-f-hat}
\end{align}
For conciseness, in our discussion below we will use $\hat{\mathbf{x}}$ to denote $\hat{\mathbf{S}}_s$, $\hat{\mathbf{p}}_s$ or $\hat{\mathbf{d}}_g$, and $\mathbf{x}$ to indicate the corresponding reference solutions generated from the HFS.

%Again, a dominating advantage of the surrogate is predicting dynamic states only in the study domains, whose dimension~$n_r$ or $n_d$ is much smaller than that of the overall simulated domain~$n_b$.

\subsection{Recurrent R-U-Net surrogate model for coupled problems}
\label{sec:surr_model}

The 3D recurrent R-U-Net implemented in this work is similar to the network used in our previous work for two-phase flow in the absence of geomechanical effects \citep{tang2021deep}. Here we describe the overall procedure, with emphasis on the new treatments required for the coupled problem.

\begin{figure}[htbp]
  \centering
  \includegraphics[trim={1cm 3cm 3cm 3cm},clip, scale=0.6]{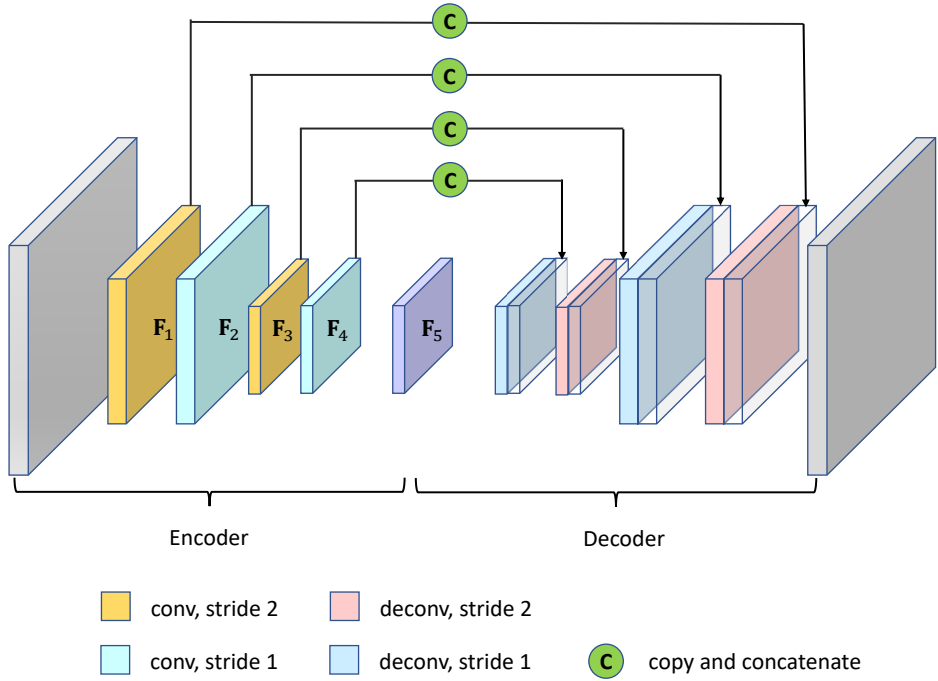}
   \caption{Schematic of 3D residual U-Net (R-U-Net) used in this work. This network entails encoding and decoding nets. Feature maps $\mathbf{F}_1$ to $\mathbf{F}_4$ extracted in the encoding net are concatenated with the upsampled features in the decoding net to generate predictions for the states.}
  \label{fig:r-u-net}
\end{figure}

The residual U-Net (referred to as R-U-Net), illustrated in Fig.~\ref{fig:r-u-net}, is a key component of the recurrent R-U-Net architecture. U-Nets provide robust architectures for high-dimensional matrix-wise regression problems \citep{ronneberger2015u}. In our setting, this network is used to capture the spatial correlation between the input rock properties $\mathbf{m}_s$ and the output dynamic states. 
%$\mathbf{S}_s$, $\mathbf{p}_s$, $\mathbf{d}_g$. 
U-Nets contain encoding and decoding nets, with the extracted feature maps $\mathbf{F}_1$ to $\mathbf{F}_4$ in the encoding net concatenated with the upsampled features in the decoding net, as shown in Fig.~\ref{fig:r-u-net}. This linkage between the encoding and decoding pathways facilitates information flow in training, and leads to improved accuracy in the predicted states \citep{tang2020jcp, tang2021deep}. Residual layers \citep{he2016deep} are added to process the most compressed feature $\mathbf{F}_5$. The residual U-Net architecture thus described can provide high accuracy for steady-state predictions, but the network in Fig.~\ref{fig:r-u-net} is not designed to capture temporal dynamics. 

%Next we described the incorporation of recurrency, which enables dynamic predictions.

In order to capture the time evolution of the saturation, pressure and displacement states, we incorporate the R-U-Net into a convolutional long-short term memory network (convLSTM). The convLSTM network \citep{xingjian2015convolutional} introduces recurrency, enabling us to capture dynamics associated with high-dimensional features. The overall framework, referred to as the 3D recurrent R-U-Net, is thus able to represent the temporal evolution of the 3D (saturation and pressure) and 2D (surface displacement) fields of interest. 

\begin{figure}[htbp]
  \centering
  \includegraphics[trim={3cm 2.5cm 1cm 2.5cm},clip, scale=0.6]{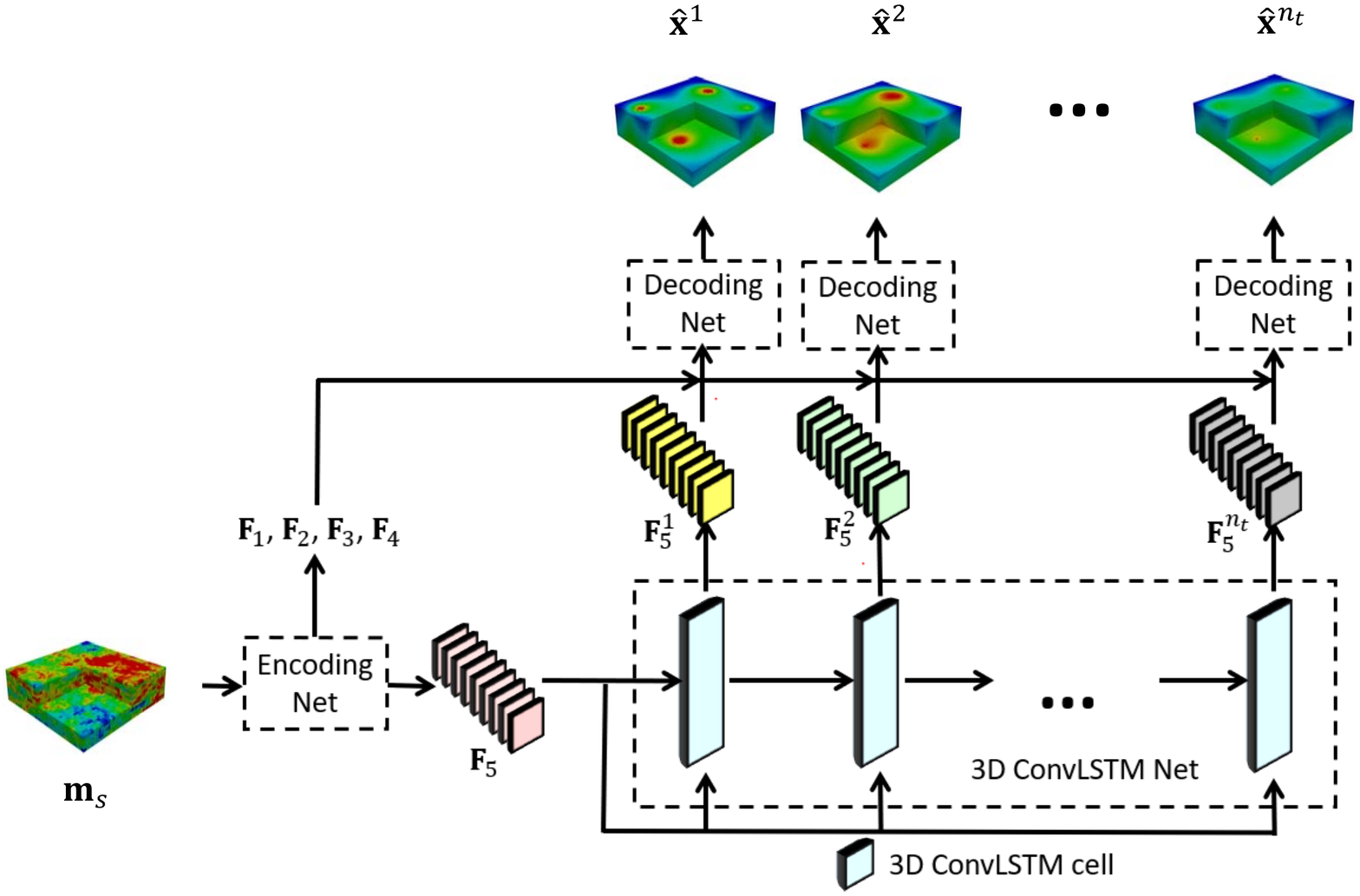}
  \caption{Schematic of the recurrent R-U-Net architecture. The convLSTM network accepts the global feature map $\mathbf{F}_5$ from the encoding net and generates a sequence of feature maps $\mathbf{F}_5^t$, $t=1, \ldots, n_t$. These are decoded, separately, into a sequence of predictions for the states $\hat{\mathbf{x}}^{t}$, $t=1,\ldots, n_t$, using the same decoding net. Here $\hat{\mathbf{x}}$ represents the predicted states $\hat{\mathbf{S}}_s$, $\hat{\mathbf{p}}_s$ and $\hat{\mathbf{d}}_g$. Figure modified from \citep{tang2021deep}.}
  \label{fig:recurrent-r-u-net}
\end{figure}

The architecture of the 3D recurrent R-U-Net is shown in Fig.~\ref{fig:recurrent-r-u-net}. The encoding net extracts feature maps $\mathbf{F}_1$ to $\mathbf{F}_5$. These maps incorporate effects at a range of scales, with $\mathbf{F}_5$ representing the most compressed/global set of features. Importantly, the convLSTM propagates only $\mathbf{F}_5$, meaning this is the only feature map that varies in time. The resulting representations are denoted $\mathbf{F}_5^t$, $t = 1, \ldots, n_t$, where $n_t$ is the number of time steps at which the surrogate model provides predictions (recall $n_t < n_{ts}$). The $\mathbf{F}_5^t$ maps are then combined with the features $\mathbf{F}_1$ to $\mathbf{F}_4$ and upsampled to provide the predicted state maps $\hat{\mathbf{x}}^t$, $t = 1, \ldots, n_t$. The use of the same $\mathbf{F}_1$ to $\mathbf{F}_4$ at all time steps leads to a reduction in the number of tunable network parameters and thus to savings in training time.

As discussed in Section~\ref{sec:domains}, the domain for the state predictions $\hat{\mathbf{x}}^t$ need not coincide precisely with that from the HFS. Thus, rather than predict vertical displacement at the $n_g$ surface nodes lying above the storage aquifer, we predict this quantity at the tops of the blocks in the uppermost layer of the overburden domain. These displacements (for the HFS), denoted $\mathbf{d}_{gb} \in \mathbb{R}^{n_{gb} \times n_t}$, are determined from $\mathbf{d}_{g} \in \mathbb{R}^{n_{g} \times n_t}$ through a simple averaging. More specifically, the block-top displacement value is computed as the average of the four nodal values at the top surface. We thus have $n_{gb}=n_g-n_{dx}-n_{dy}+1$, where $n_{dx}$ and $n_{dy}$ are the number of nodes in the $x$ and $y$ directions at the top surface (above the storage aquifer), with $n_g=n_{dx} \times n_{dy}$. Note we need these maps at only $n_t$ (rather than $n_{ts}$) time steps.

Our prediction of $\hat{\mathbf{d}}_{gb} \in \mathbb{R}^{n_{gb} \times n_t}$ is accomplished using the same recurrent R-U-Net architecture that is used for saturation and pressure. This means that we actually predict a 3D field, and then project it to the target 2D map $\hat{\mathbf{d}}_{gb}^t$ at each time step. The recurrent R-U-Net could, in principle, be modified to predict the 2D maps directly, but this would require additional exploration of network architecture and associated hyperparameters. Our approach entails copying and concatenating the surface deformation map $\mathbf{d}_{gb}^t$ $n_l$ times, where $n_l$ is the number of layers in the storage aquifer. With this treatment, the displacement data are of the same dimension as $\mathbf{S}_s^t$ and $\mathbf{p}_s^t$. The requisite 2D map $\hat{\mathbf{d}}_{gb}^t$ is then constructed as the average of the result over the $n_l$ layers. As we will see, this approach provides accurate predictions, while maintaining consistency in the network for all of the predicted quantities. 

For full details on the recurrent R-U-Net architecture used in this study, please see Appendix~B in \citep{tang2021deep}. A key advantage of this network is its flexibility in terms of reuse for different problems. This includes the ability to predict 2D maps along with 3D fields, which enables the convenient handling of surface displacement described above.

\subsection{Data preprocessing and network training}
\label{sec:training}

A set of 2000 high-fidelity simulations of the coupled problem, with each run corresponding to a different storage-aquifer realization $\mathbf{m}_{s}$, is performed to provide the training data. The output state maps at prescribed time steps are saved. The boundary conditions, along with the well locations and CO$_2$ injection-rate specifications, are the same in all runs.

Appropriate data preprocessing is necessary to achieve optimal network performance. In our previous work \citep{tang2020jcp, tang2021deep}, we considered binary channelized geomodels, which were characterized by $m_i=0$ (background mud) and $m_i=1$ (sand). Thus no additional scaling of the geomodel was required. Saturation values fall between 0 and 1, so no processing was required for these fields. Pressure fields were normalized through use of `detrending' and `min-max' scaling, which were found to provide better results than time-independent normalization. 

In this study, the input multi-Gaussian fields $\mathbf{m}_s$, which are used to construct $\log k$ and $\phi$, are standard normal variables and are thus $O(1)$. Thus no preprocessing is applied. Saturation fields $\mathbf{S}_s$, again between 0 and 1, can also be used directly. For the pressure and surface displacement fields, we apply detrending and standardization, which we found to provide slightly better accuracy than our previous treatments \citep{tang2020jcp, tang2021deep}. The specific preprocessing for pressure and surface displacement is given by
  \begin{equation}
     \tilde{\mathbf{x}}^{t}_i = \frac{\mathbf{x}^{t}_i -  \text{mean}([\mathbf{x}_1^t, \ldots,\mathbf{x}_{n_{smp}}^t ])}{\text{std}([\mathbf{x}_1^t, \ldots,\mathbf{x}_{n_{smp}}^t ]}, \quad i=1,\ldots, n_{smp}, \ \ t = 1, \ldots, n_t,
     \label{eq:p-0}
\end{equation}
where $n_{smp}$ is the number of training samples. With this approach, each pressure and displacement field at time step $t$, $\mathbf{x}^{t}_i$, is normalized by subtracting the mean field (over all $n_{smp}$ samples) and dividing each element by its standard deviation at that time step.

%The recurrent R-U-Net weights, denoted by $\boldsymbol{\theta}$, are then optimized to minimize the difference between the neural network predictions $\hat{\mathbf{x}}^t$ and the HFS reference $\mathbf{x}^t$. 

The recurrent R-U-Net weights, denoted by $\boldsymbol{\theta}$, are then optimized to minimize the difference between the direct neural network predictions $\tilde{\hat{\mathbf{x}}}^t$ (before inverse transform) and the normalized HFS reference $\tilde{\mathbf{x}}^t$. This optimization can be expressed as
%[xxx1 -- should previous sentence be ``The recurrent R-U-Net weights, denoted by $\boldsymbol{\theta}$, are then optimized to minimize the difference between the normalized neural network predictions ${\tilde {\hat {\mathbf{x}}}}^t$ and the HFS reference $\tilde{\mathbf{x}}^t$.'' Note addition of normalized. And should it be ${\tilde {\hat {\mathbf{x}}}}^t_i$ and $\tilde{\mathbf{x}}^t_i$ in Eq.~\ref{eq:loss-function}? Note the hats and tildes.] 
%
\begin{equation}
\boldsymbol{\theta}^* = \argmin_{\boldsymbol\theta} \frac{1}{n_{smp}}\frac{1}{n_t} \sum_{i=1}^{n_{smp}} \sum_{t=1}^{n_t} ||{\tilde{\hat {\mathbf x}}}^t_i - \tilde{\mathbf{x}}_{i}^{t}||_{2}^2,
\label{eq:loss-function}
\end{equation}
% \begin{equation}
% \boldsymbol{\theta}^* = \argmin_{\boldsymbol\theta} \frac{1}{n_{smp}}\frac{1}{n_t} \sum_{i=1}^{n_{smp}} \sum_{t=1}^{n_t} ||{\hat {\mathbf x}}^t_i - \mathbf{{x}}_{i}^{t}||_{2}^2,
% \label{eq:loss-function}
% \end{equation}
%
where $\boldsymbol{\theta}^*$ indicates the optimized neural network parameters. Separate $\boldsymbol{\theta}^*$ are determined for the saturation, pressure and displacement fields (i.e., we train three networks). Note that the mismatch in all quantities is measured in the $L^2$ norm, in contrast to our approach in \citep{tang2021deep}.

The neural network parameters are determined using the adaptive moment estimation (ADAM) optimizer \citep{kingma2014adam}. The batch size is set to 4, and a total of 300~epochs are used to assure convergence. Each of the three trainings converges within 5~hours on a Nvidia Tesla V100 GPU. These trainings can be performed in parallel on a single GPU given the small batch size. 

% Surrogate model validation

\section{Surrogate Model Evaluation}
\label{sec:eval}

In this section, we first describe the setup for the carbon storage problem with coupled flow and geomechanics used in our evaluations. Results for 3D saturation and pressure fields and 2D surface displacement maps will then be presented. Error statistics and percentile results (P$_{10}$, P$_{50}$, P$_{90}$) for key quantities will also be provided.

\subsection{Coupled problem setup}
\label{sect:setup}

As discussed in Section~\ref{sec:method} and shown in Fig.~\ref{fig:reservoir-fig}, the overall problem corresponds to a domain of dimensions 20~km $\times$ 20~km $\times$ 2~km (in the $x$, $y$ and $z$ directions). The overall model is defined on a $60 \times 60 \times 37$ grid. The grid-block size is constant within the storage aquifer region (corresponding to $40 \times 40 \times 12$ blocks), but it increases in the surrounding region as we move outwards. The overburden rock and bedrock are of thickness 1380~m and 500~m respectively. 

The model contains four injection wells, which are open to flow in all 12 layers in the storage aquifer. Each well injects fully-saturated supercritical CO$_2$ at a constant mass flow rate of 30~kg CO$_2$ per second. This is equivalent to $0.946 \times 10^9$~kg CO$_2$ per year, or almost 1~Mta (megatonnes per anum), per well. Thus the total injection into the system is nearly 4~Mta, which corresponds to a large-scale storage operation. The top of the overburden rock is subject to atmospheric pressure. The simulation time frame is 30~years.

\begin{table}
 \caption{Parameters used in the coupled models}
  \centering
  \begin{tabular}{ll}
    \hline
    \textbf{Storage aquifer parameters}     &  \textbf{Value}     \\
    \hline
    Thickness & 120~m \\
    Permeability & Multi-Gaussian field \\
    Porosity &  Multi-Gaussian field  \\
    Young's modulus & 5~GPa \\
    \hline
       \textbf{Overburden parameters}     &  \textbf{Value}     \\
    \hline
    Thickness & 1380~m \\
    Permeability & $10^{-7}$~md \\
    Porosity & 0.3\\
    Young's modulus & 1~GPa \\
    \hline
    \textbf{Bedrock parameters}     &  \textbf{Value}     \\
    \hline
    Thickness & 500~m \\
    Permeability & $10^{-7}$~md \\
    Porosity & 0.3\\
    Young's modulus & 50~GPa \\
    \hline
    \textbf{Surrounding region parameters}   &  \textbf{Value}     \\
    \hline
    Permeability & 300~md \\
    Porosity & 0.3 \\
    \hline
    \textbf{Capillary pressure -- Van Genuchten model} & \textbf{Value}  \\
    \hline
    $\lambda$ & 0.254 \\
    $S_{ar}$ & 0.11   \\
    $P_\text{max}$ & 12,500~Pa   \\
    \hline
    \textbf{Relative permeability -- Corey model} & \textbf{Value}  \\
    \hline
    $S_{ar}$ & 0.11    \\
    $S_{gr}$ & 0.01 \\
    $n_a$ & 4 \\
    $n_g$ & 2 \\
    \hline
  \end{tabular}
  \label{table: simulation-param}

\end{table}

\begin{figure}
     \centering
     \begin{subfigure}[b]{0.32\textwidth}
         \centering
         \includegraphics[trim={0cm 0cm, 0cm, 0cm},clip,scale=0.22]{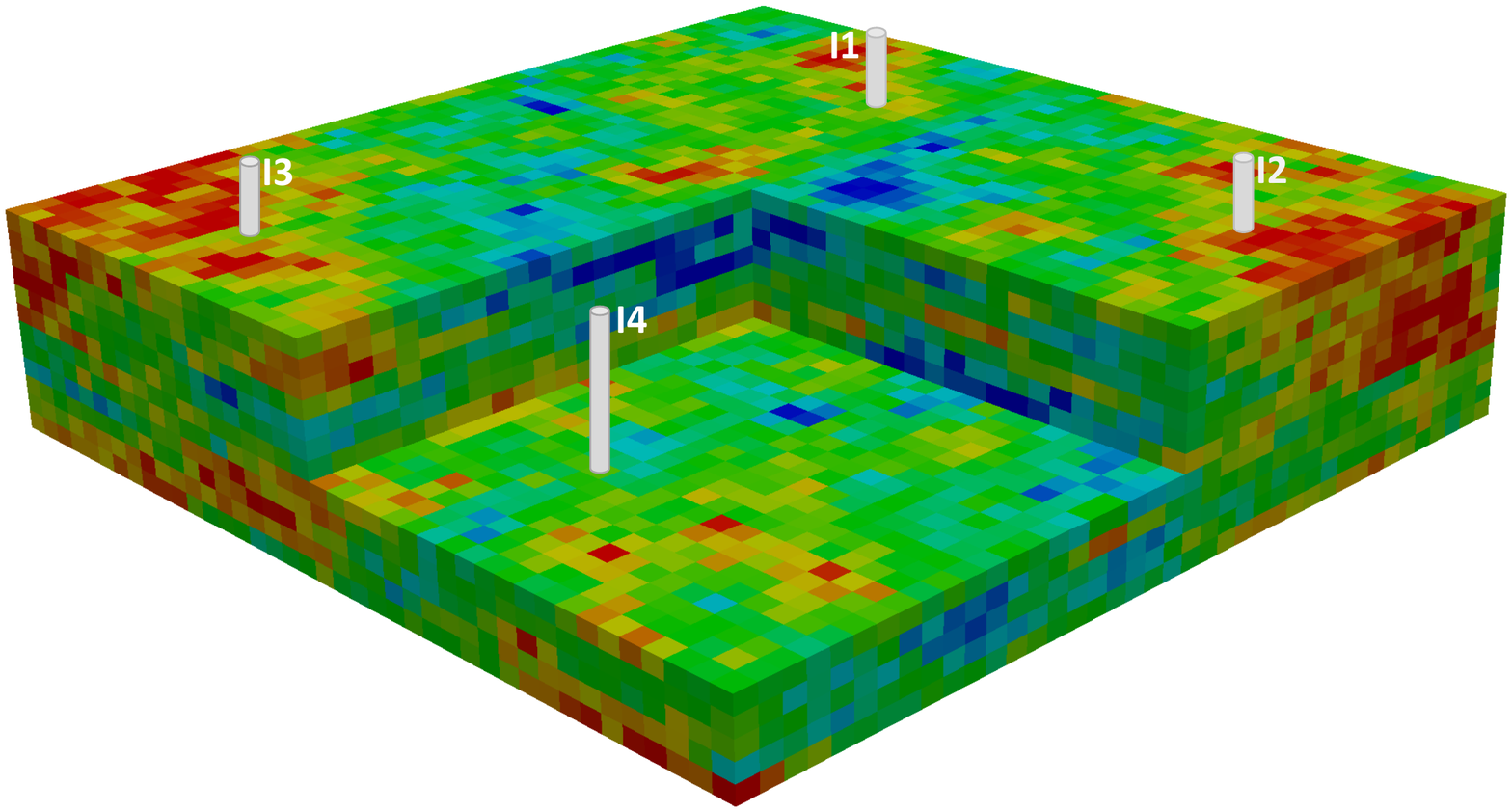}
         \caption{}
         \label{test-sat-surr-case-1}
     \end{subfigure}
     %\hfill
    \begin{subfigure}[b]{0.32\textwidth}
         \centering
         \includegraphics[trim={0cm 0cm, 0cm, 0cm},clip,scale=0.22]{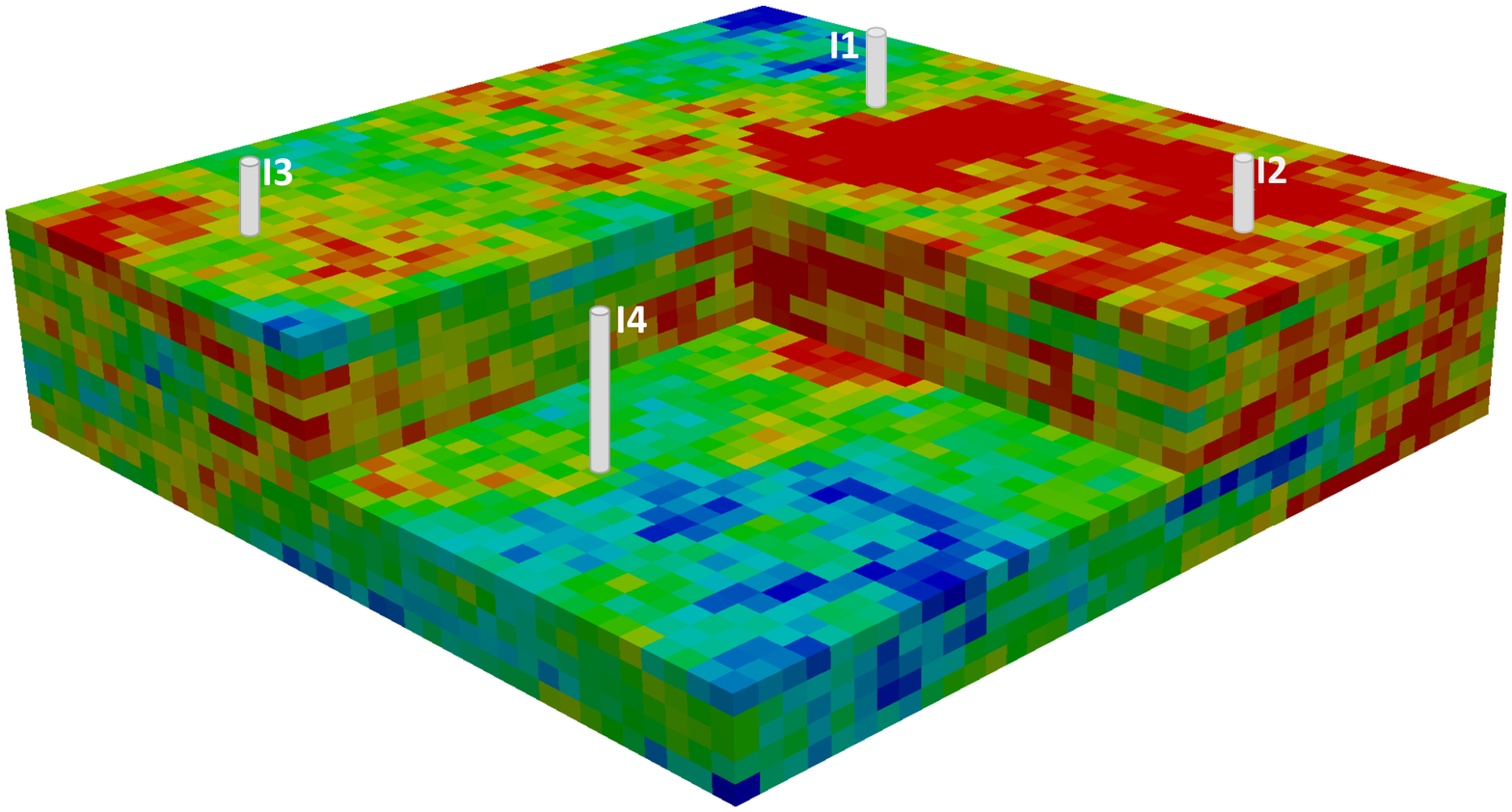}
         \caption{}
         \label{test-sat-surr-case-2}
     \end{subfigure}
     %\hfill
      \begin{subfigure}[b]{0.32\textwidth}
         \centering
         \includegraphics[trim={0cm 0cm, 0cm, 0cm},clip,scale=0.22]{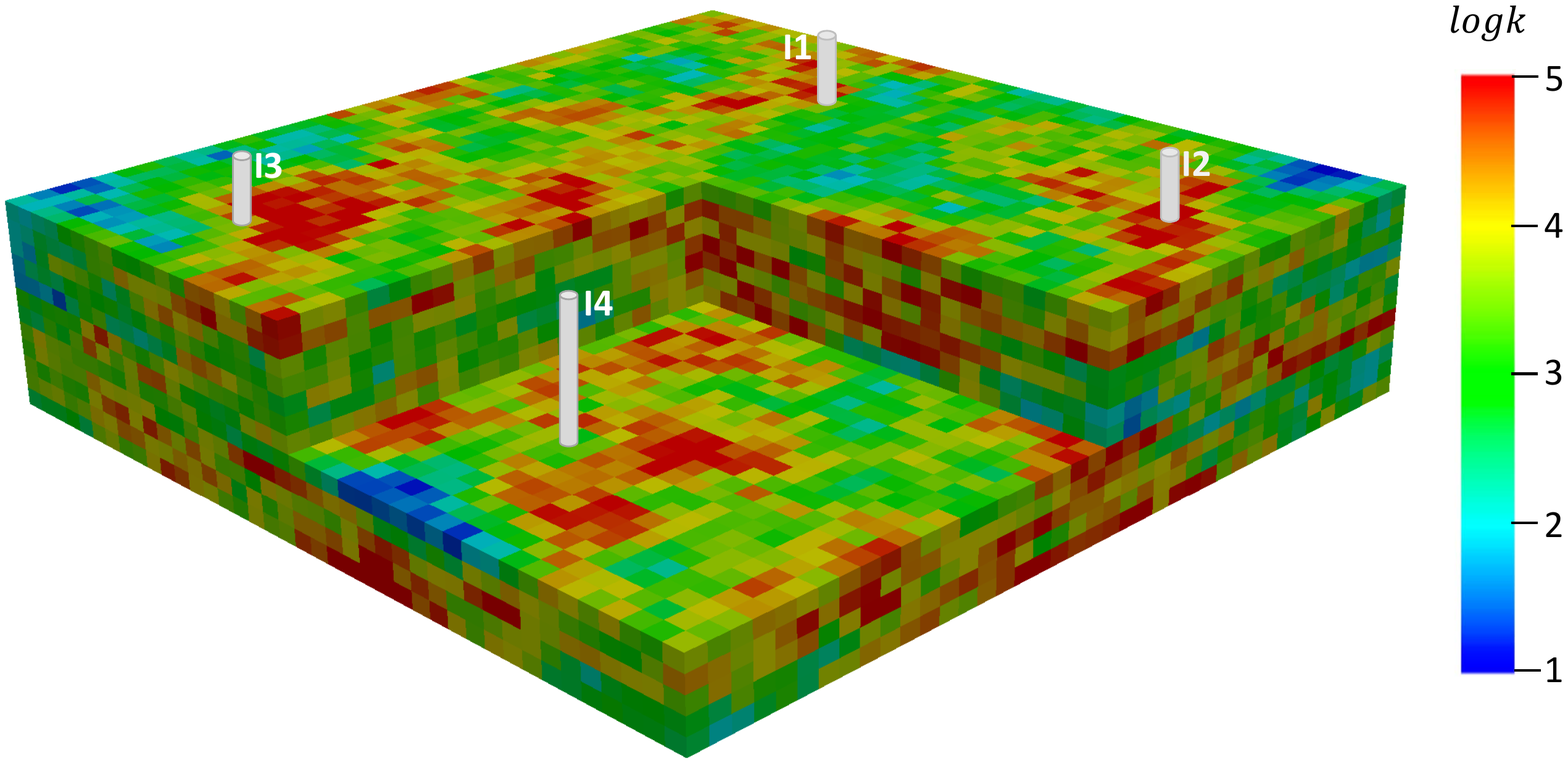}
         \caption{}
         \label{test-sat-surr-case-3}
     \end{subfigure}
\caption{Multi-Gaussian log-permeability realizations honoring hard data at the four injection wells (indicated by white cylinders). The realization in (c) is the `true' model used for data assimilation in Section~\ref{sec:hm}.}   

\label{fig:sgems-gaussian-reals}
\end{figure}

As noted in Section~\ref{sec:method}, the storage aquifer is characterized by multi-Gaussian fields generated using SGeMS \citep{remy2009applied}. Three realizations of the resulting log-permeability fields are displayed in Fig.~\ref{fig:sgems-gaussian-reals}. An exponential variogram model was used, and the correlation lengths were specified to be 20 grid blocks in the $x$ and $y$ directions and 5 grid blocks in the $z$ direction. The mean and standard deviation of log-permeability are 2.5 and 1 respectively, which leads to an (arithmetic) average permeability of 23~md in the storage aquifer. Porosity is computed (block by block) directly from log-permeability, with a mean of 0.3 and a standard deviation of 0.05. Permeability and porosity are conditioned to well data, meaning all realizations display the same properties in well blocks. 

Relevant parameters for the four domains are summarized in Table~\ref{table: simulation-param}. Note that Poisson's ratio is set to 0.2 and Biot's coefficient to 1.0 for all blocks in the overall model. As is evident from the table, the overburden and bedrock are essentially impermeable ($k=10^{-7}$~md), while the surrounding region is characterized by $k=300$~md. Porosity is set to 0.3 in overburden rock, bedrock and surrounding regions (the porosity values in the impermeable overburden and bedrock have virtually no effect on the solution). Young's modulus is set to 5~GPa in the storage aquifer, 50~GPa in the bedrock and 1~GPa in the overburden. The bedrock serves as a rigid boundary and thus has a high Young's modulus value, while the overburden is susceptible to deformation in our model, and is therefore characterized by a low Young's modulus.

The fluid properties for CO$_2$ and water are obtained from correlations developed by Altunin et al.~\citep{altunin1968thermophysical} and steam table equations according to IAPWAS-IF97 \citep{wagner2008iapws}, respectively. At a pressure of 136~bar (corresponding to the top of the storage aquifer, at a depth of 1380~m), CO$_2$ viscosity and density are 0.035~cp and 484~kg/m$^3$, and water viscosity and density are 0.43~cp and 986~kg/m$^3$. Capillary pressure and relative permeability parameters are provided in Table~\ref{table: simulation-param}. Here, $\lambda$ is the exponent coefficient in the Van Genuchten capillary pressure model, $S_{ar}$ is the irreducible aqueous-phase saturation (required for the capillary pressure and relative permeability models), and $P_\text{max}$ is the maximum capillary pressure. The quantity $S_{gr}$ is the gaseous-phase residual saturation, and $n_a$ and $n_g$ are the exponents in the Corey relative permeability model.

\subsection{Recurrent R-U-Net saturation, pressure and displacement fields}

We generate a total of 2000 random realizations of the multi-Gaussian storage-aquifer permeability and porosity fields using SGeMS. These fields are `inserted' into the overall model description (which also includes the other three domains), and high-fidelity numerical simulation is then performed for each model. All runs are conducted using the GEOS simulator \citep{settgast2017fully}. The 3D saturation and pressure fields in the storage aquifer, and the 2D surface displacement maps, are saved at 10 time steps (0.5, 2, 5, 8, 11, 15, 19, 23, 27, 30~years). These solution states constitute the training data.

\begin{figure}
     \centering
     \begin{subfigure}[b]{0.32\textwidth}
         \centering
         \includegraphics[trim={0cm 0cm, 0cm, 0cm},clip,scale=0.22]{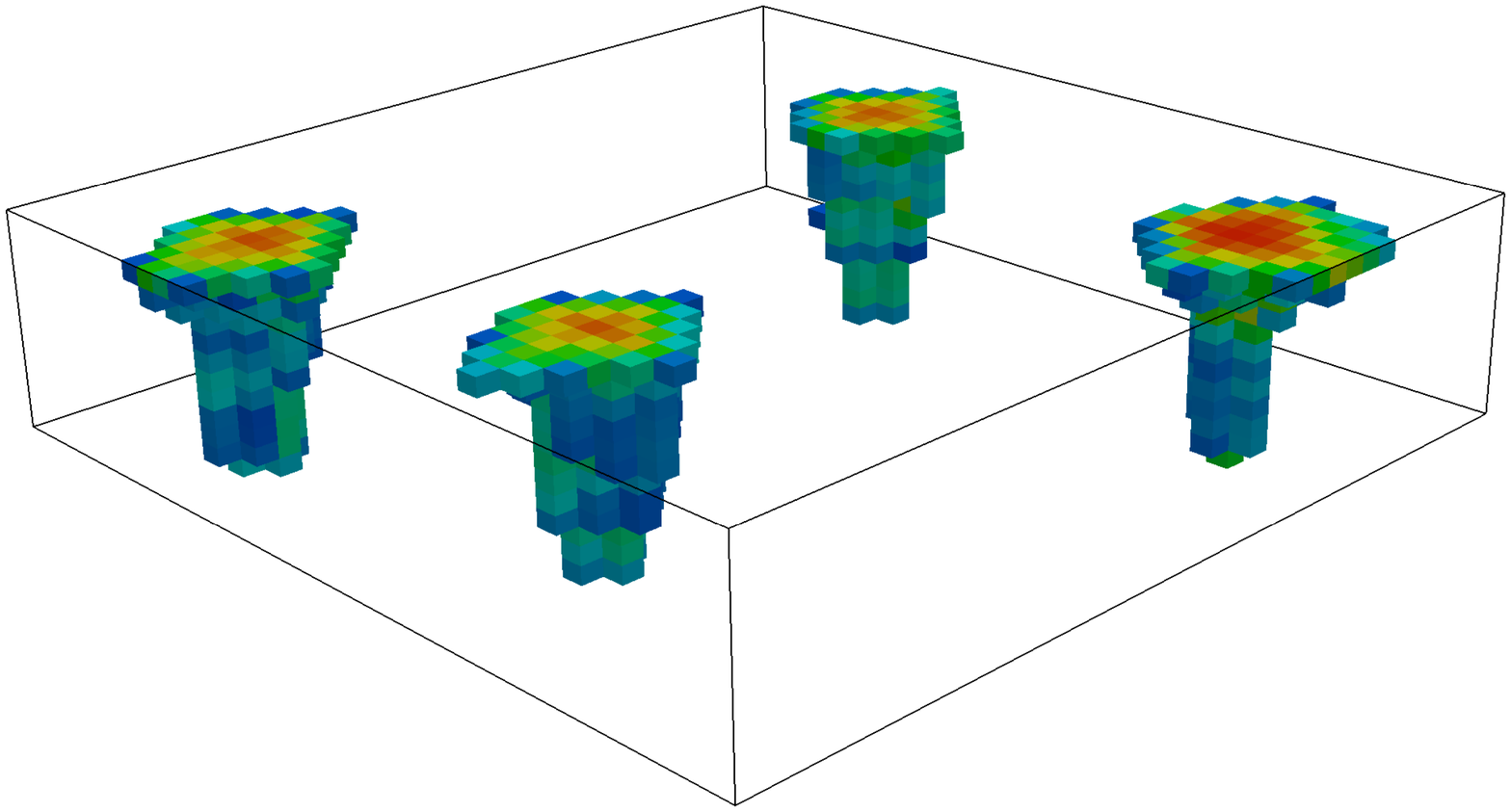}
         \caption{5 years (surr)}
         \label{test-sat-surr-case-1}
     \end{subfigure}
     %\hfill
    \begin{subfigure}[b]{0.32\textwidth}
         \centering
         \includegraphics[trim={0cm 0cm, 0cm, 0cm},clip,scale=0.22]{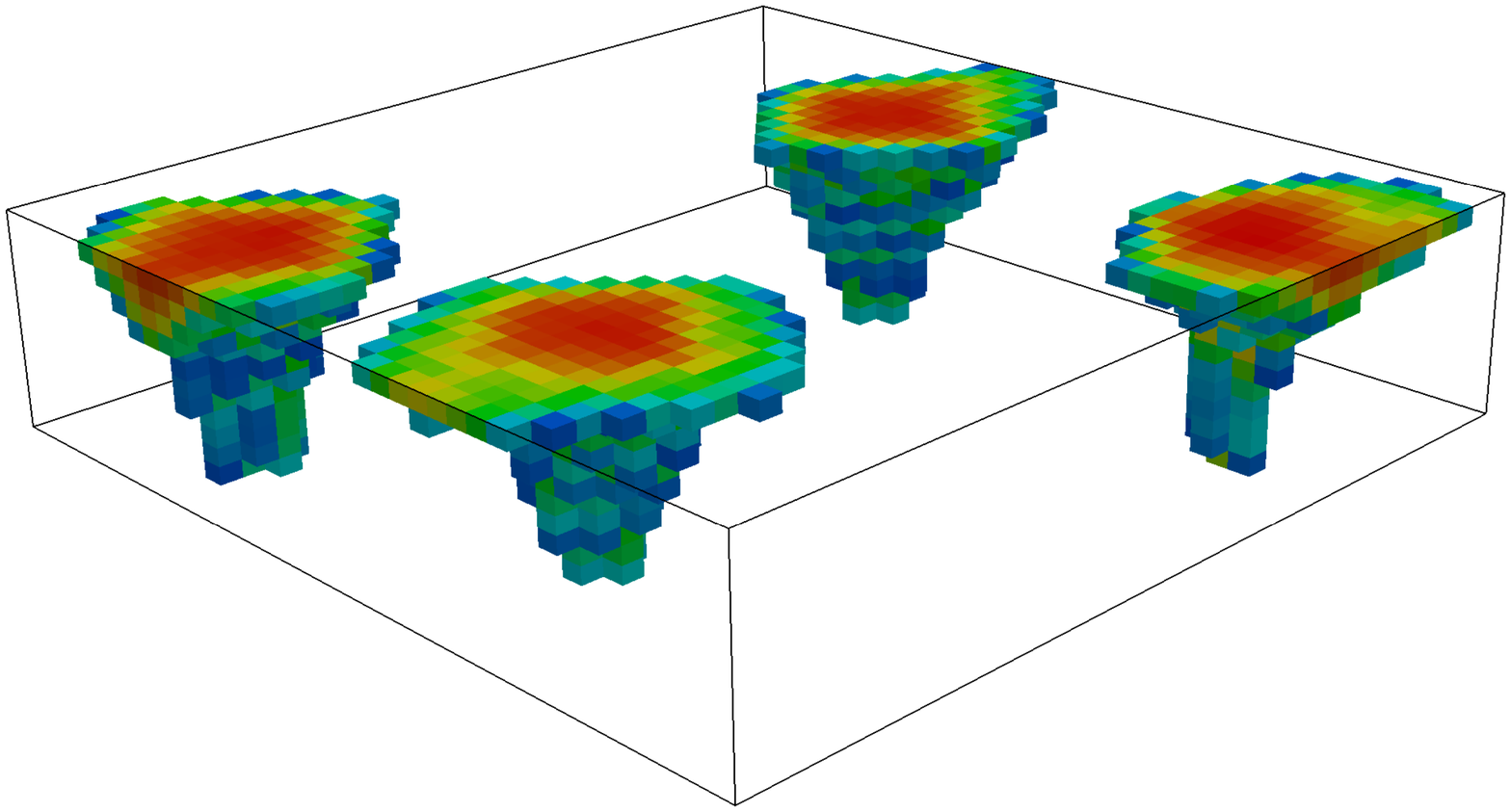}
         \caption{15 years (surr)}
         \label{test-sat-surr-case-2}
     \end{subfigure}
     %\hfill
      \begin{subfigure}[b]{0.32\textwidth}
         \centering
         \includegraphics[trim={0cm 0cm, 0cm, 0cm},clip,scale=0.22]{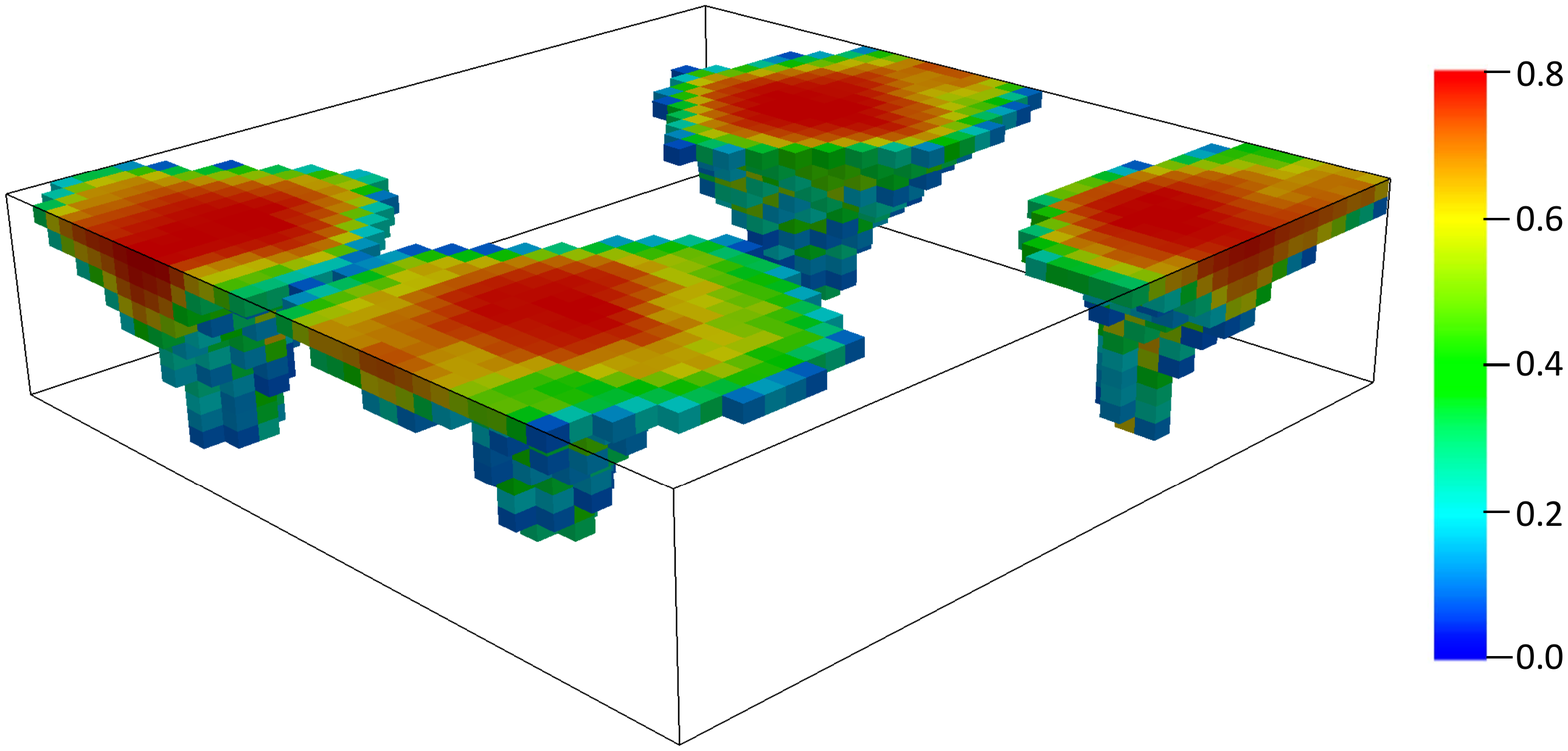}
         \caption{30 years (surr)}
         \label{test-sat-surr-case-3}
     \end{subfigure}
          \begin{subfigure}[b]{0.32\textwidth}
         \centering
         \includegraphics[trim={0cm 0cm, 0cm, 0cm},clip,scale=0.22]{figs-ccs/fields-pdf/sat-surr-case0-t1.pdf}
         \caption{5 years (sim)}
         \label{test-sat-sim-case-1}
     \end{subfigure}
     %\hfill
    \begin{subfigure}[b]{0.32\textwidth}
         \centering
         \includegraphics[trim={0cm 0cm, 0cm, 0cm},clip,scale=0.22]{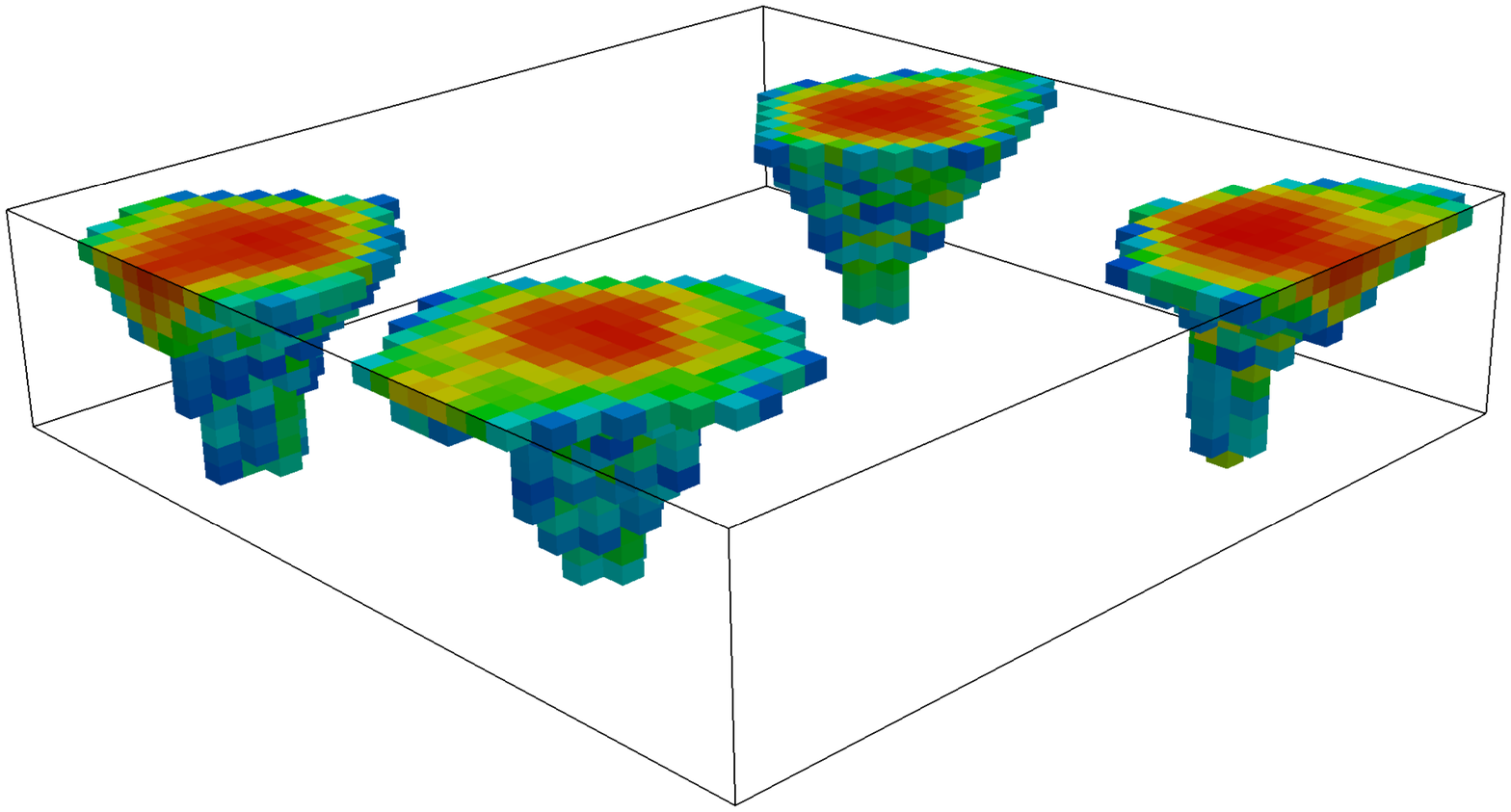}
         \caption{15 years (sim)}
         \label{test-sat-sim-case-2}
     \end{subfigure}
     %\hfill
      \begin{subfigure}[b]{0.32\textwidth}
         \centering
         \includegraphics[trim={0cm 0cm, 0cm, 0cm},clip,scale=0.22]{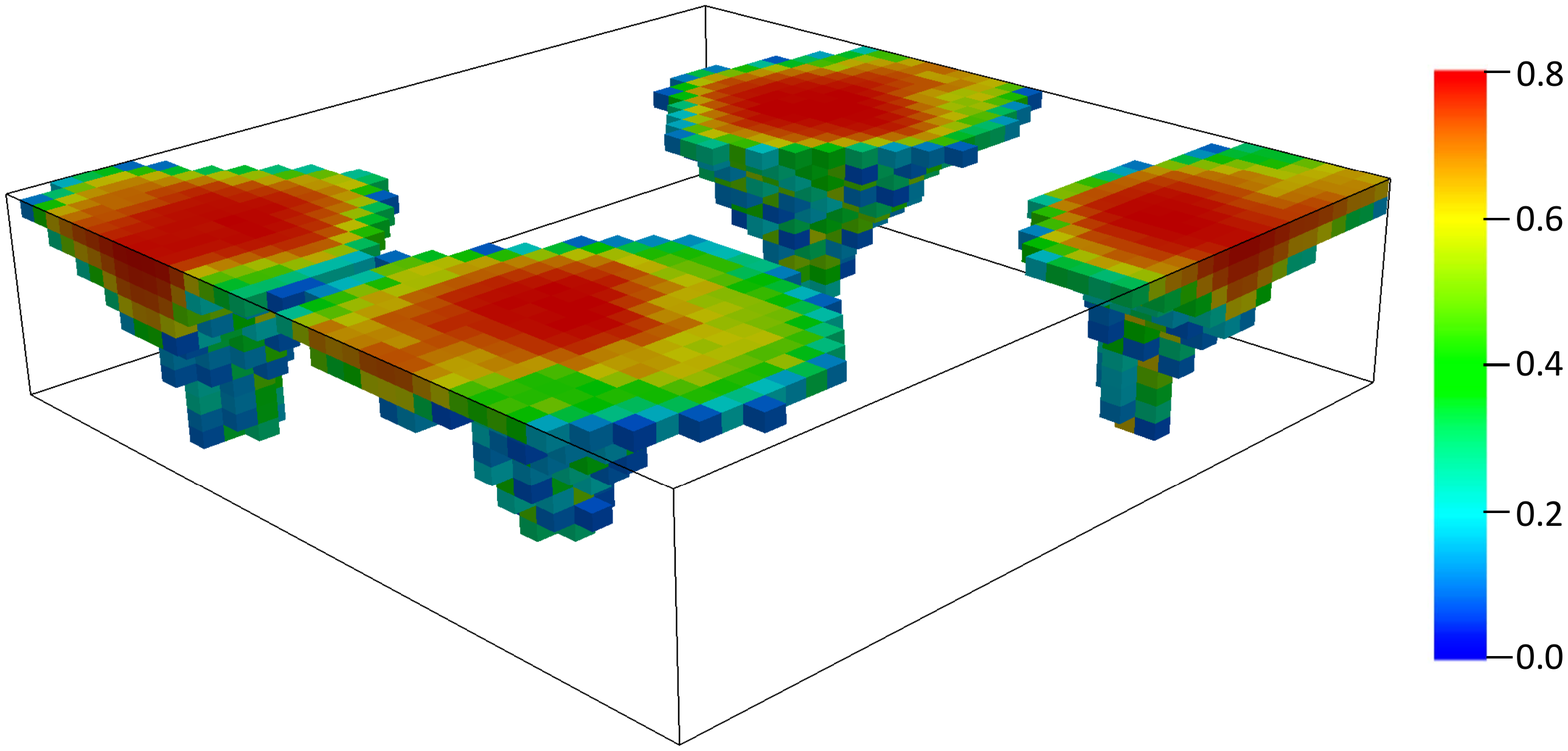}
         \caption{30 years (sim)}
         \label{test-sat-sim-case-3}

     \end{subfigure}
\caption{CO$_2$ saturation fields from recurrent R-U-Net surrogate model (upper row) and HFS (lower row) for a representative test case at three different times.}
\label{fig:sat-varying-time}
\end{figure}

A new set of 500 geomodels is generated and simulated (at high fidelity). These results represent the test cases used to assess the performance of the surrogate model. Saturation fields within the storage aquifer for a particular test case, at three different time steps, are shown in Fig.~\ref{fig:sat-varying-time}. For visual clarity, only grid blocks with CO$_2$ saturation greater than 0.01 are displayed (note that we use the term CO$_2$ saturation to refer to gaseous-phase saturation). The saturation error for this case (quantified later in Eq.~\ref{eq:sat-relative-error}) is greater than the median error over the 500 test cases, so these results can be considered to be representative. The upper row in Fig.~\ref{fig:sat-varying-time} shows the recurrent R-U-Net surrogate predictions, while the lower row displays the reference HFS results. We observe variability in the CO$_2$ saturation distributions around the different injection wells, despite the fact that they all inject at the same rate. There is, nonetheless, very close agreement between the surrogate and HFS results. It is evident from Fig.~\ref{fig:sat-varying-time}c and f that some of the injected CO$_2$ has left the storage aquifer (and entered the surrounding region, which is not shown in the figure).

In terms of timing, it takes about 0.8~hours of parallel computation (on 32 cores) to complete a single high-fidelity simulation. The trained surrogate, by contrast, can provide batch-wise predictions for 100 geomodels in less than 1~second, or $\lesssim 0.01$~second per realization. Thus the recurrent R-U-Net surrogate does indeed provide dramatic speedups for this coupled problem.

\begin{figure}
     \centering
     \begin{subfigure}[b]{0.32\textwidth}
         \centering
         \includegraphics[trim={0cm 0cm, 0cm, 0cm},clip,scale=0.22]{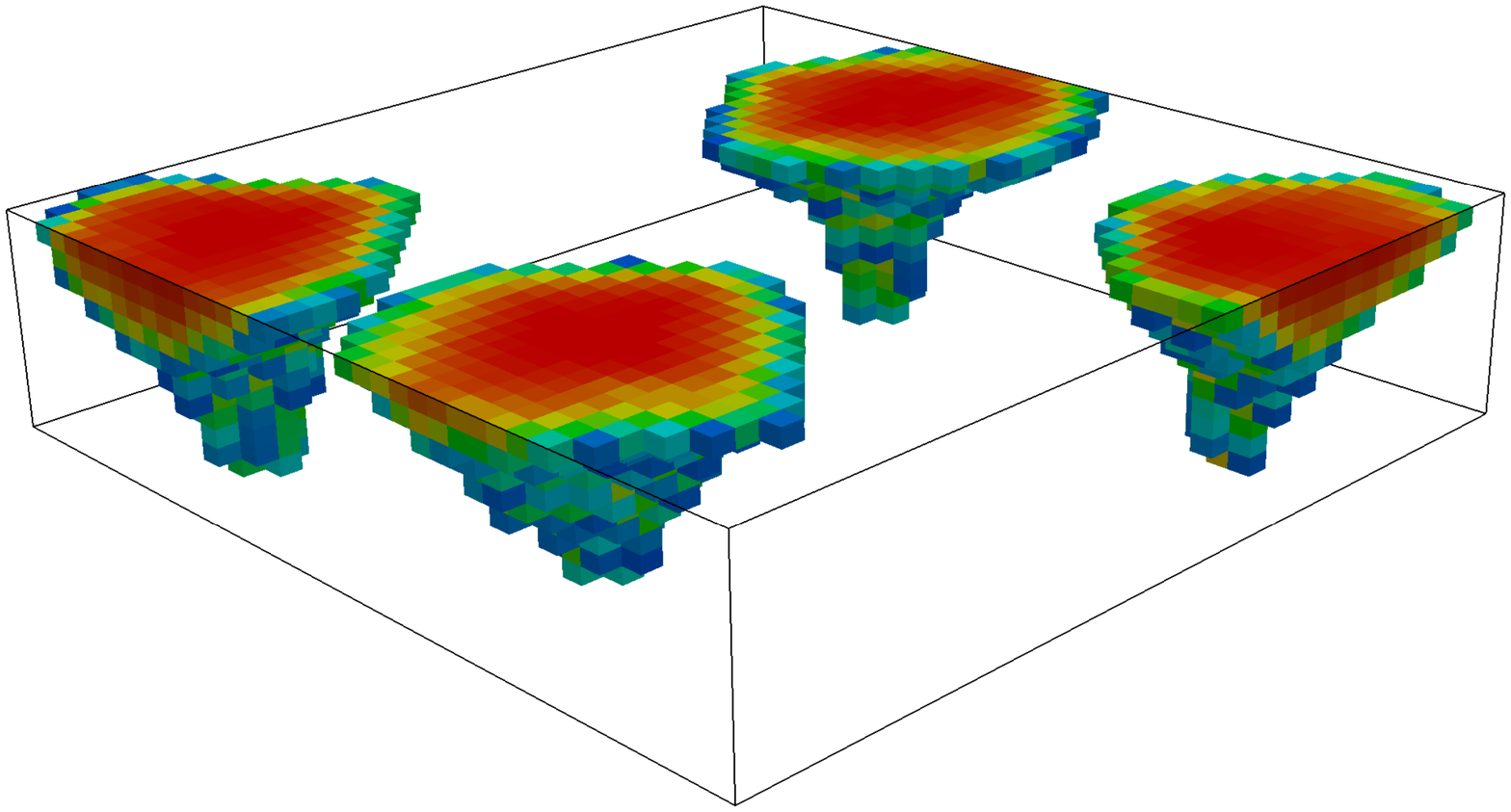}
         \caption{Realization 1 (surr)}
         \label{test-sat-surr-case-1}
     \end{subfigure}
     %\hfill
    \begin{subfigure}[b]{0.32\textwidth}
         \centering
         \includegraphics[trim={0cm 0cm, 0cm, 0cm},clip,scale=0.22]{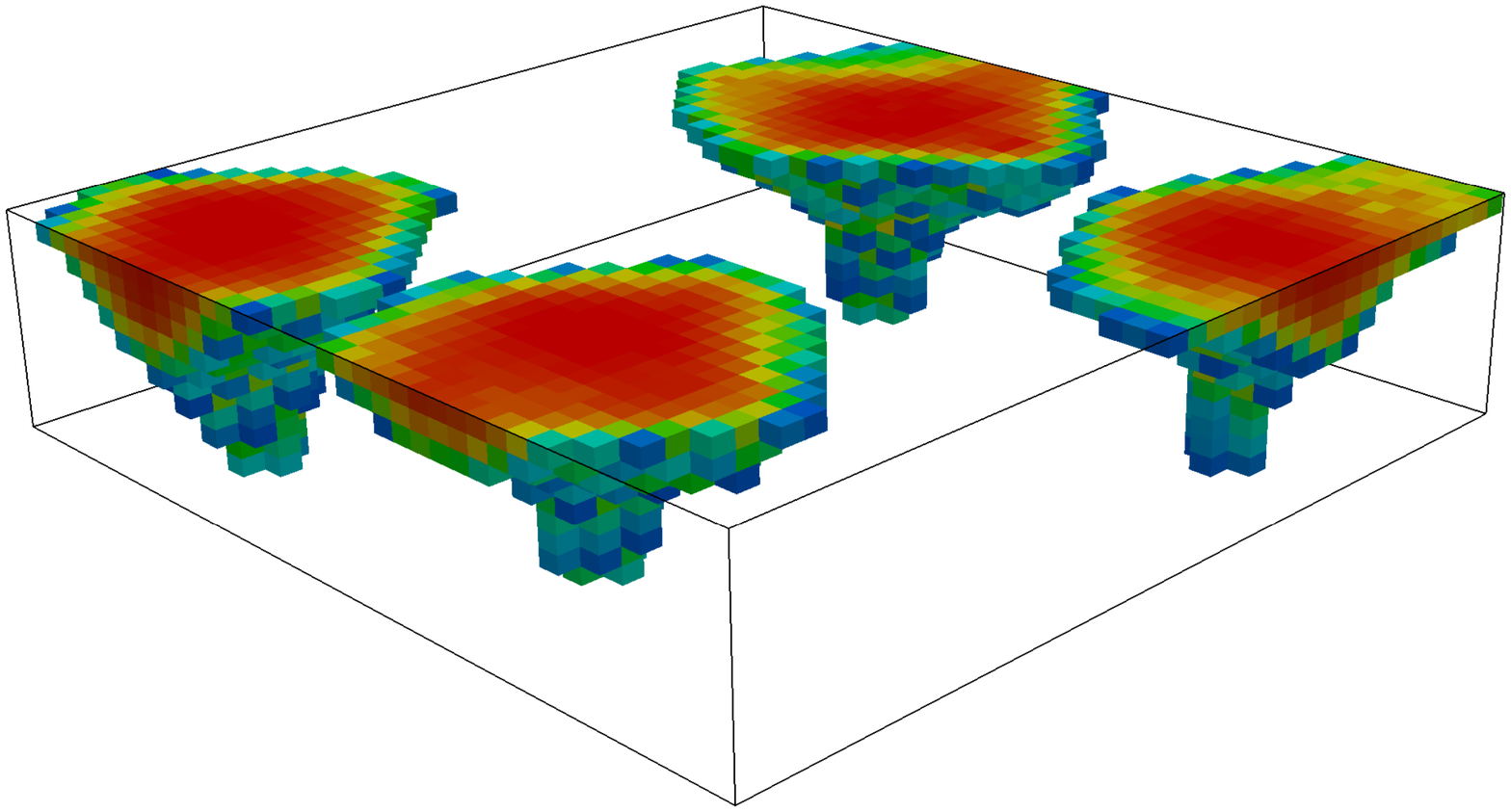}
         \caption{Realization 2 (surr)}
         \label{test-sat-surr-case-2}
     \end{subfigure}
     %\hfill
      \begin{subfigure}[b]{0.32\textwidth}
         \centering
         \includegraphics[trim={0cm 0cm, 0cm, 0cm},clip,scale=0.22]{figs-ccs/fields-pdf/sat-surr-case0-t3.pdf}
         \caption{Realization 3 (surr)}
         \label{test-sat-surr-case-3}
     \end{subfigure}
          \begin{subfigure}[b]{0.32\textwidth}
         \centering
         \includegraphics[trim={0cm 0cm, 0cm, 0cm},clip,scale=0.22]{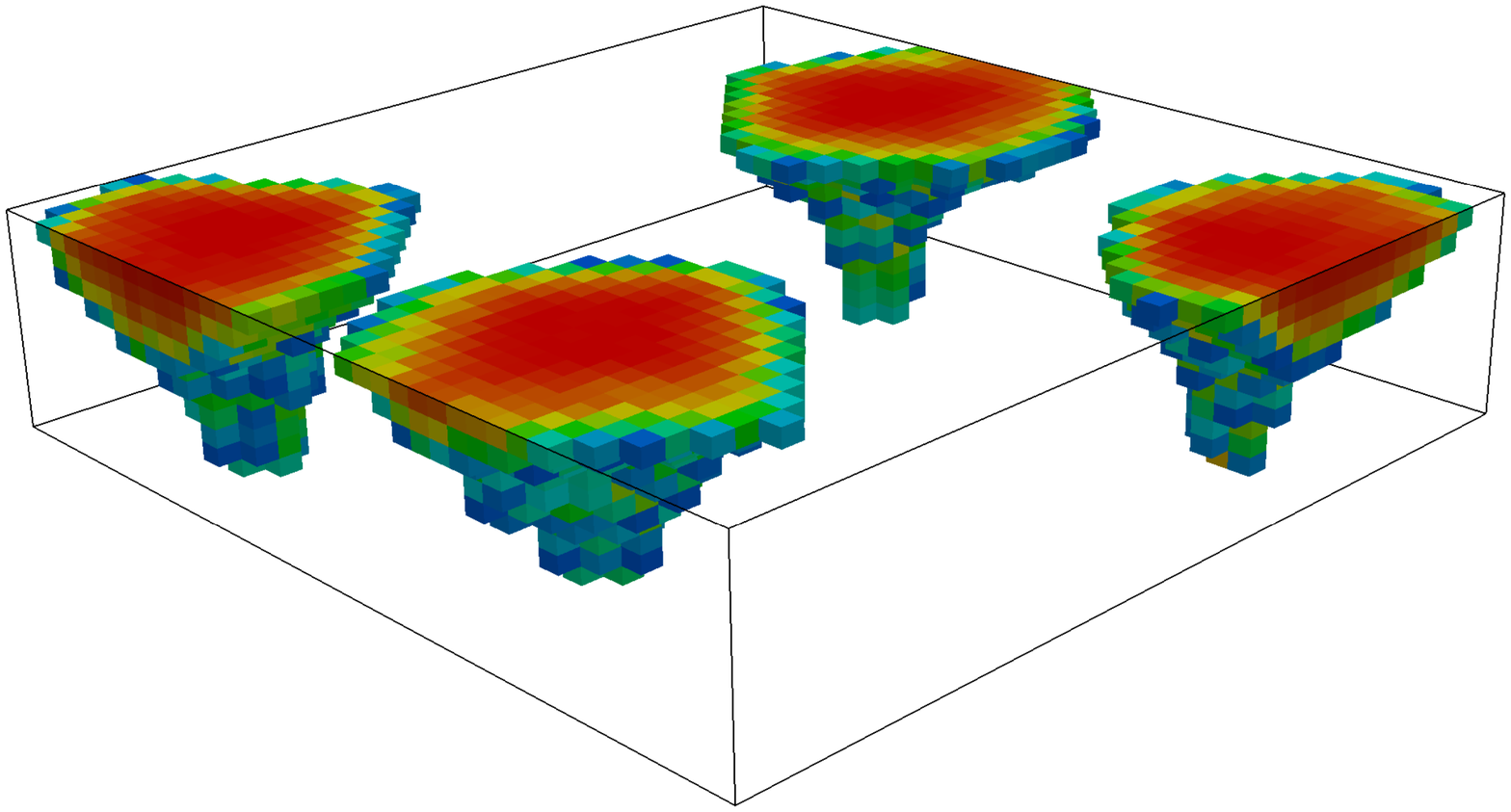}
         \caption{Realization 1 (sim)}
         \label{test-sat-sim-case-1}
     \end{subfigure}
     %\hfill
    \begin{subfigure}[b]{0.32\textwidth}
         \centering
         \includegraphics[trim={0cm 0cm, 0cm, 0cm},clip,scale=0.22]{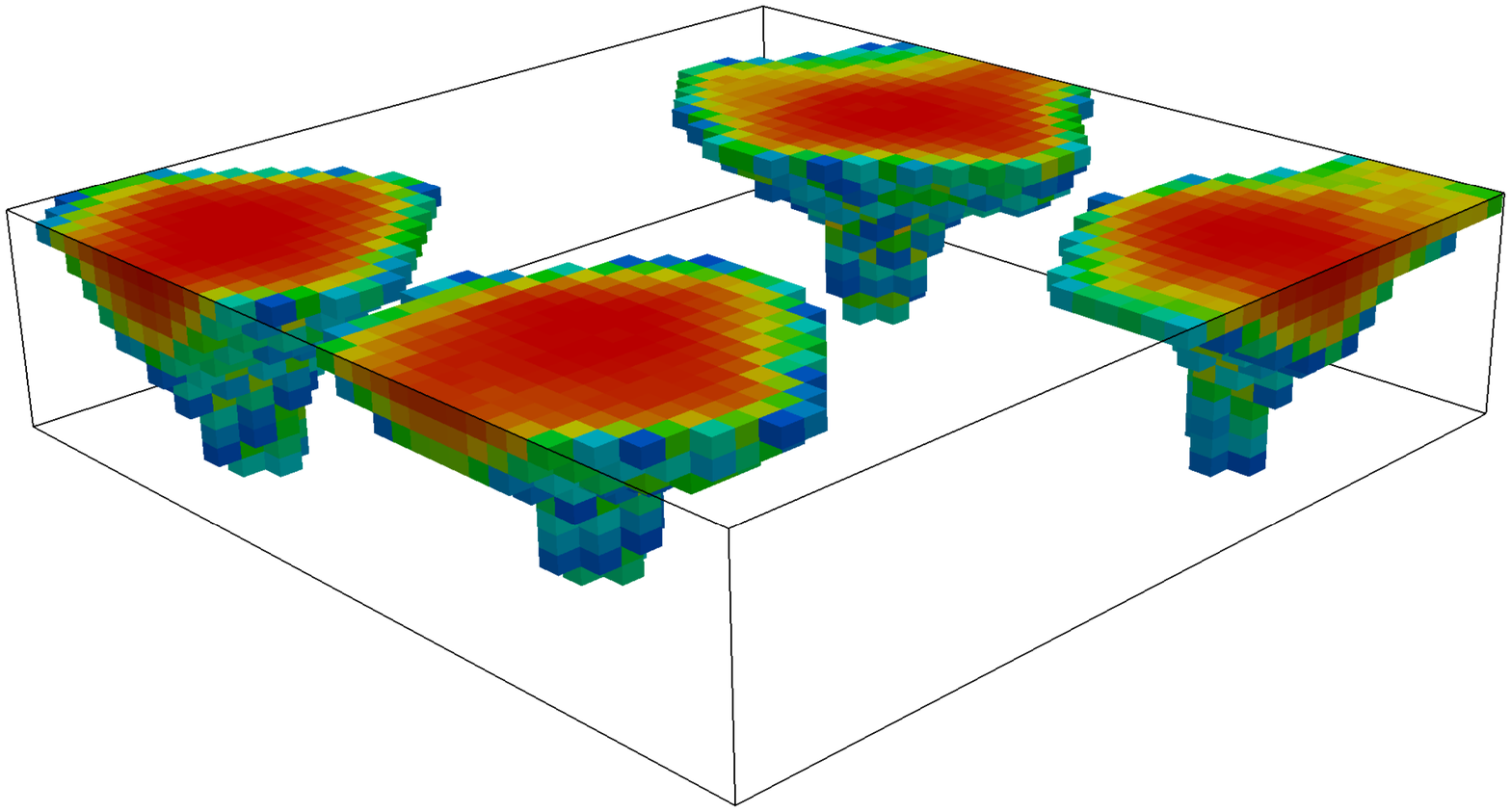}
         \caption{Realization 2 (sim)}
         \label{test-sat-sim-case-2}
     \end{subfigure}
     %\hfill
      \begin{subfigure}[b]{0.32\textwidth}
         \centering
         \includegraphics[trim={0cm 0cm, 0cm, 0cm},clip,scale=0.22]{figs-ccs/fields-pdf/sat-true-case0-t3.pdf}
         \caption{Realization 3 (sim)}
         \label{test-sat-sim-case-3}

     \end{subfigure}
\caption{CO$_2$ saturation fields from recurrent R-U-Net surrogate model (upper row) and HFS (lower row) for three different test cases at 30~years.}    

\label{fig:sat-variability}
\end{figure}

Saturation fields at the end of the 30-year simulation time frame, for three different realizations, are presented in Fig.~\ref{fig:sat-variability}. The errors for these models are all around or above the median error for the full set of test cases. Consistent with Fig.~\ref{fig:sat-varying-time}, we again observe clear agreement between the HFS and surrogate model results. There is a reasonable amount of variation from case to case, which is evident upon close inspection. For example, the top of the plume from the front-most injector is much larger in Fig.~\ref{fig:sat-variability}c than in Fig.~\ref{fig:sat-variability}a, and the top of the plume from the right-most injector is larger in Fig.~\ref{fig:sat-variability}b than in Fig.~\ref{fig:sat-variability}a or c. The 3D shapes of the plumes also vary from realization to realization. The (sometimes subtle) variations between realizations are well captured by the recurrent R-U-Net model.

\begin{figure}
     \centering
     \begin{subfigure}[b]{0.32\textwidth}
         \centering
         \includegraphics[trim={0cm 0cm, 0cm, 0cm},clip,scale=0.22]{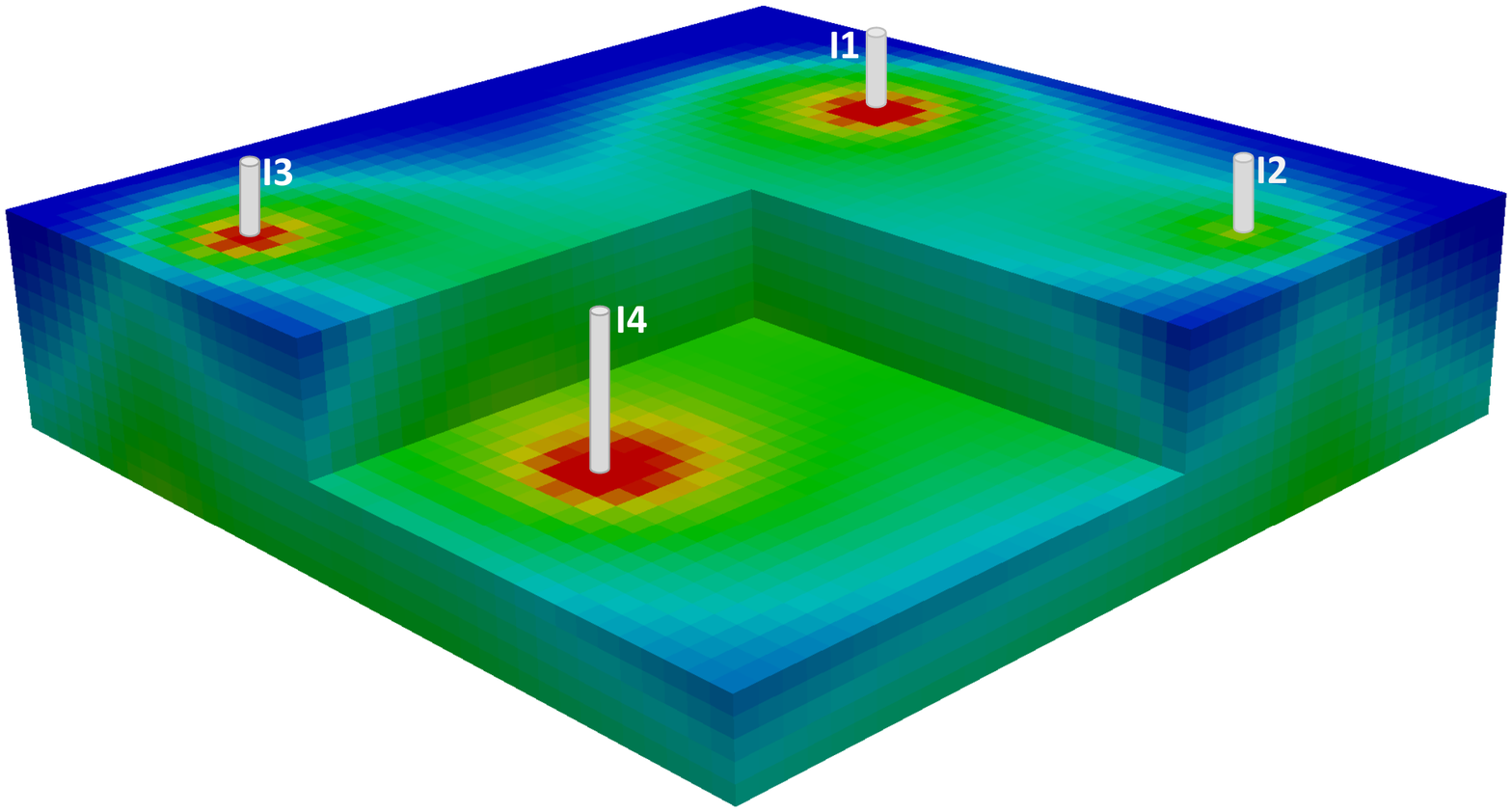}
         \caption{2 years (surr)}
         \label{test-sat-surr-case-1}
     \end{subfigure}
     %\hfill
    \begin{subfigure}[b]{0.32\textwidth}
         \centering
         \includegraphics[trim={0cm 0cm, 0cm, 0cm},clip,scale=0.22]{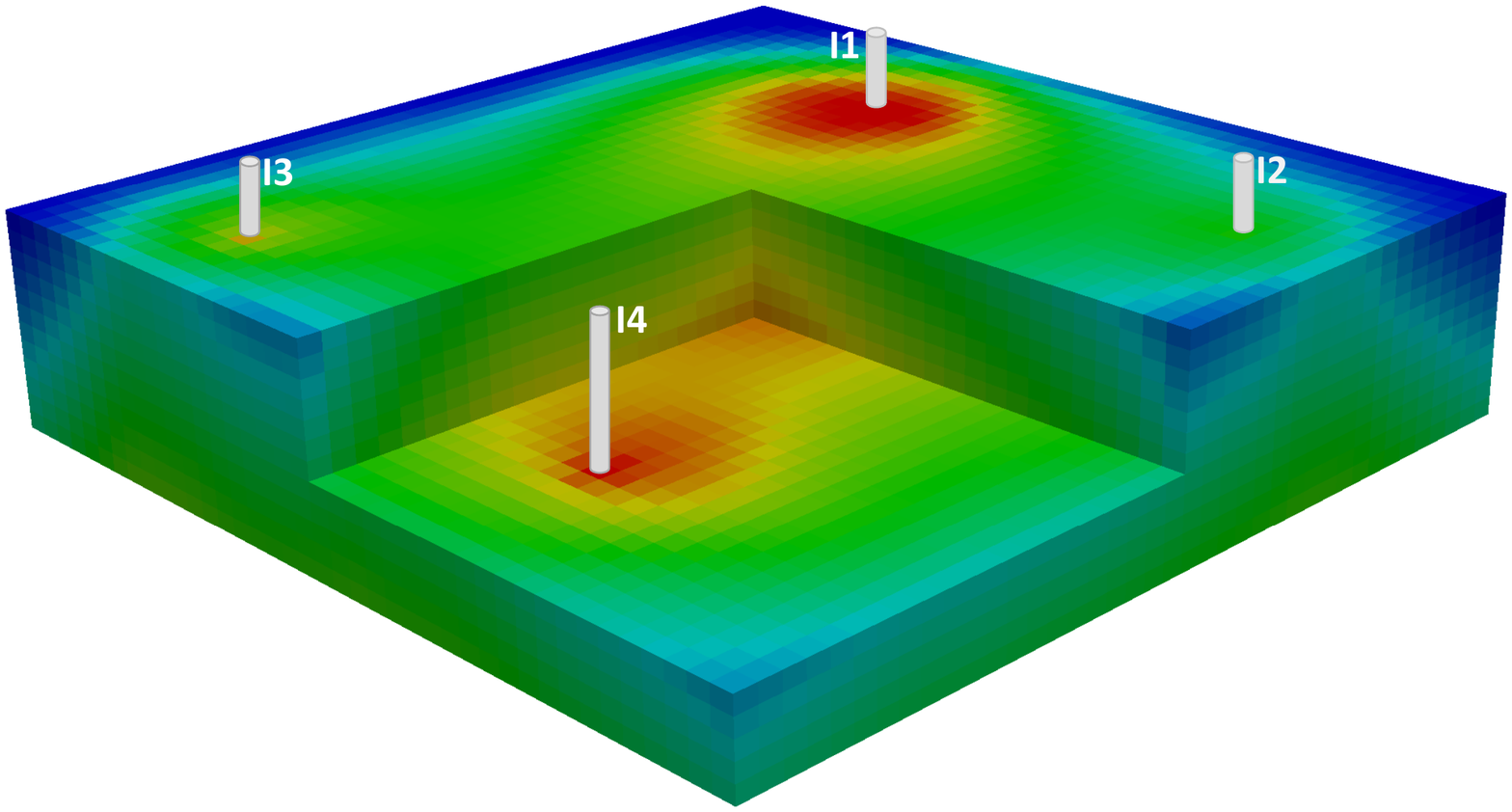}
         \caption{15 years (surr)}
         \label{test-sat-surr-case-2}
     \end{subfigure}
     %\hfill
      \begin{subfigure}[b]{0.32\textwidth}
         \centering
         \includegraphics[trim={0cm 0cm, 0cm, 0cm},clip,scale=0.22]{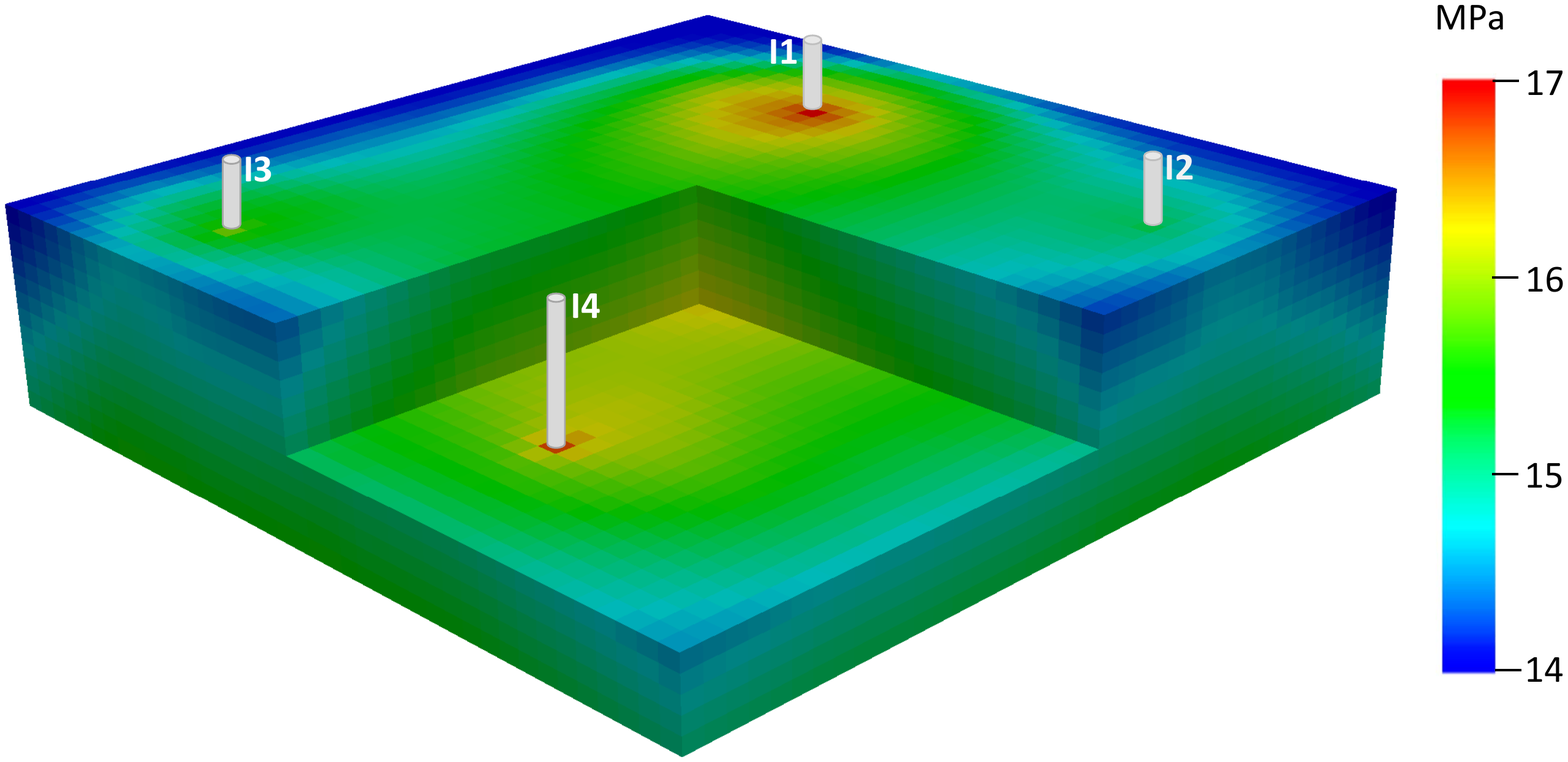}
         \caption{30 years (surr)}
         \label{test-sat-surr-case-3}
     \end{subfigure}
          \begin{subfigure}[b]{0.32\textwidth}
         \centering
         \includegraphics[trim={0cm 0cm, 0cm, 0cm},clip,scale=0.22]{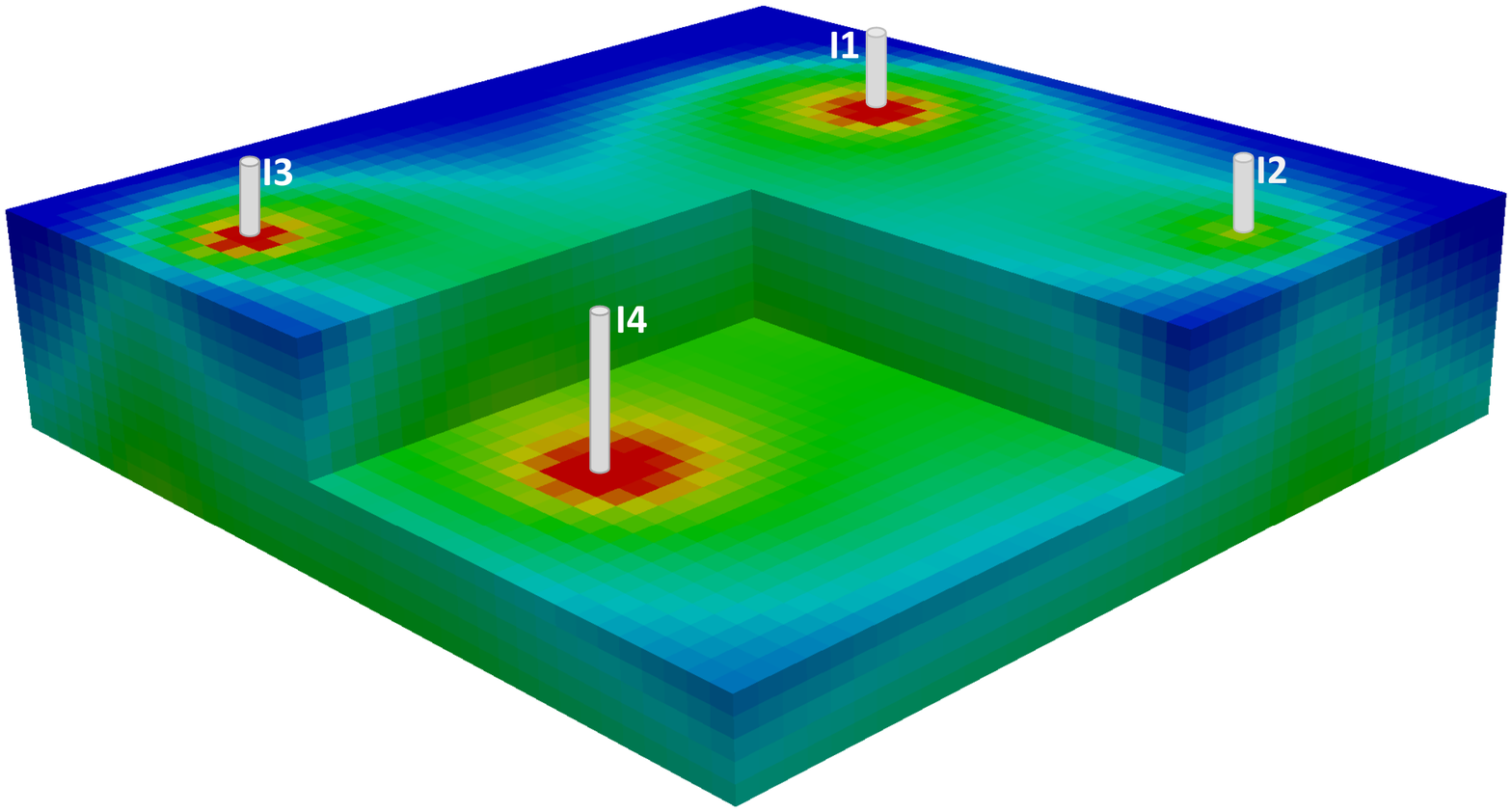}
         \caption{2 years (sim)}
         \label{test-sat-sim-case-1}
     \end{subfigure}
     %\hfill
    \begin{subfigure}[b]{0.32\textwidth}
         \centering
         \includegraphics[trim={0cm 0cm, 0cm, 0cm},clip,scale=0.22]{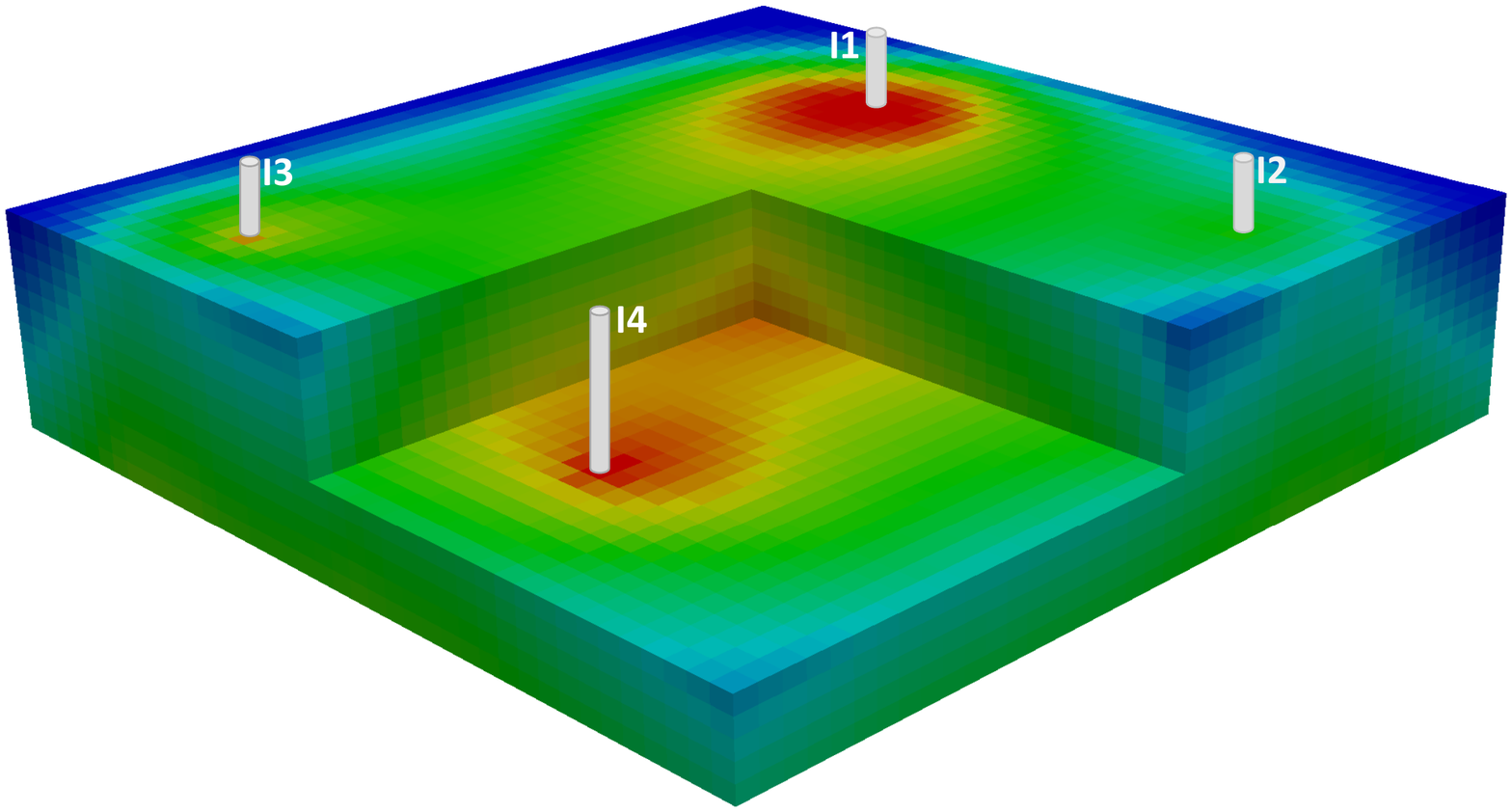}
         \caption{15 years (sim)}
         \label{test-sat-sim-case-2}
     \end{subfigure}
     %\hfill
      \begin{subfigure}[b]{0.32\textwidth}
         \centering
         \includegraphics[trim={0cm 0cm, 0cm, 0cm},clip,scale=0.22]{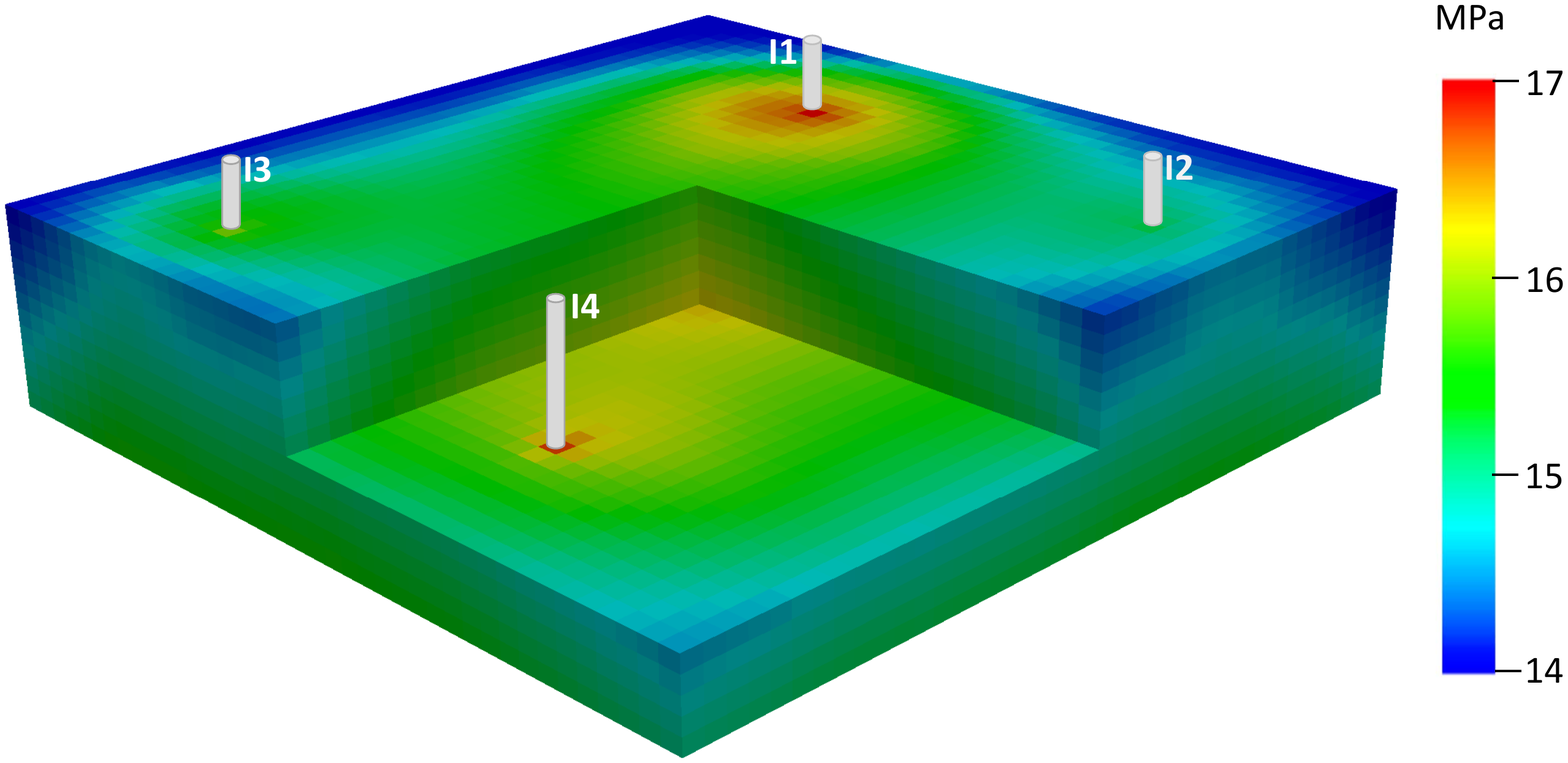}
         \caption{30 years (sim)}
         \label{test-sat-sim-case-3}

     \end{subfigure}
\caption{Pressure fields from recurrent R-U-Net surrogate model (upper row) and HFS (lower row) for a representative test case at three different times.}
\label{fig:pressure-evolution}
\end{figure}

Results for the 3D pressure field in the storage aquifer, for a representative test case (with error above the median), are shown in Fig.~\ref{fig:pressure-evolution}. Because we have nonzero capillary pressure in this problem, pressure is different in the aqueous and gaseous phases. In all of the pressure results in this paper, we present gaseous-phase pressures, which we refer to simply as `pressure.' From Fig.~\ref{fig:pressure-evolution} we see that pressure displays higher local variation at early time (2~years) than at later times. Pressure does not continuously increase with time because of the very large surrounding region included in the overall model. In the absence of this region, pressure would steadily rise with time. The correspondence between the surrogate model and the reference HFS results is again very close.

\begin{figure}
     \centering
     \begin{subfigure}[b]{0.32\textwidth}
         \centering
         \includegraphics[trim={0cm 0cm, 0cm, 0cm},clip,scale=0.22]{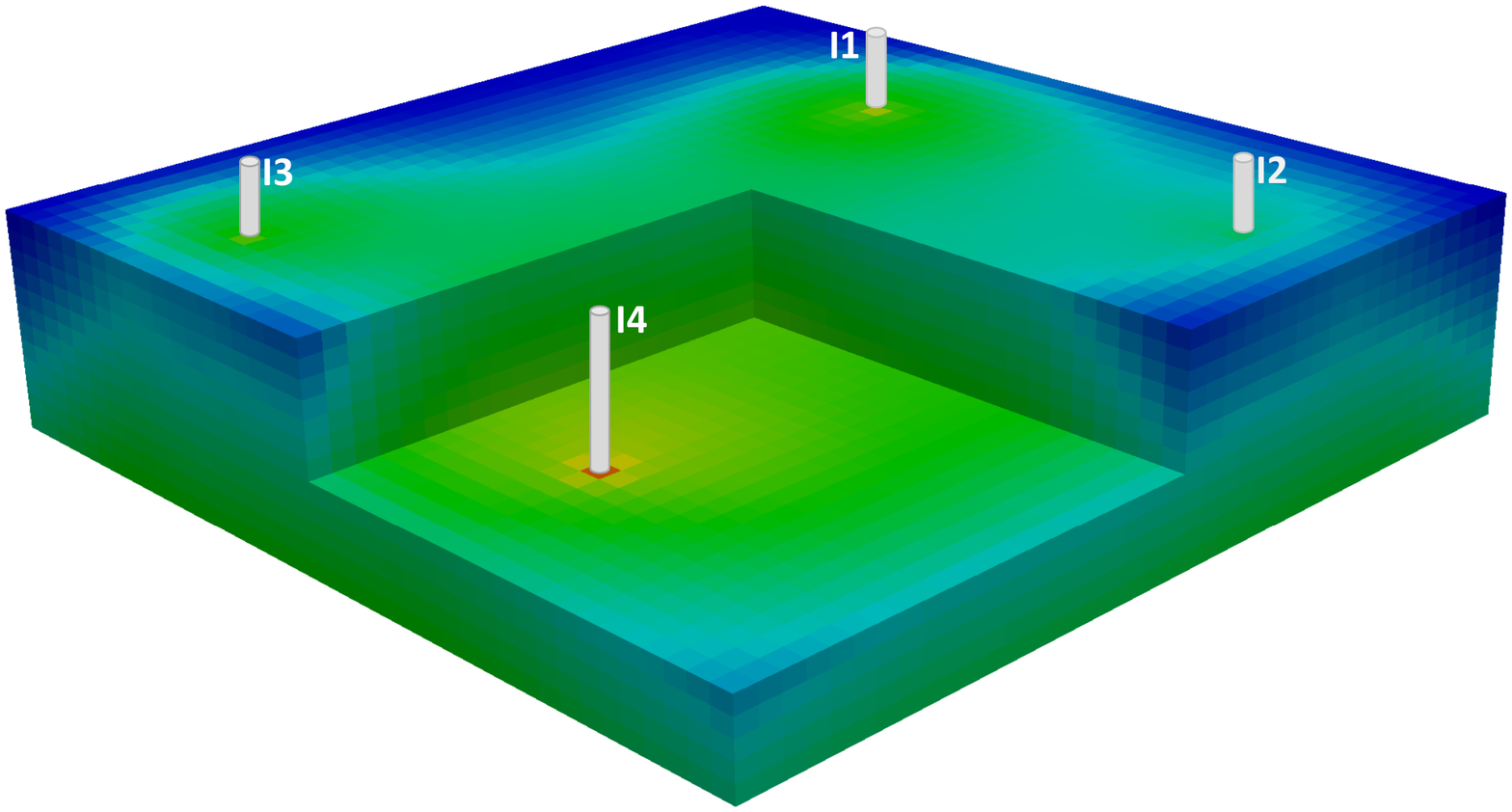}
         \caption{Realization 1 (surr)}
         \label{test-p-surr-case-1}
     \end{subfigure}
     %\hfill
    \begin{subfigure}[b]{0.32\textwidth}
         \centering
         \includegraphics[trim={0cm 0cm, 0cm, 0cm},clip, scale=0.22]{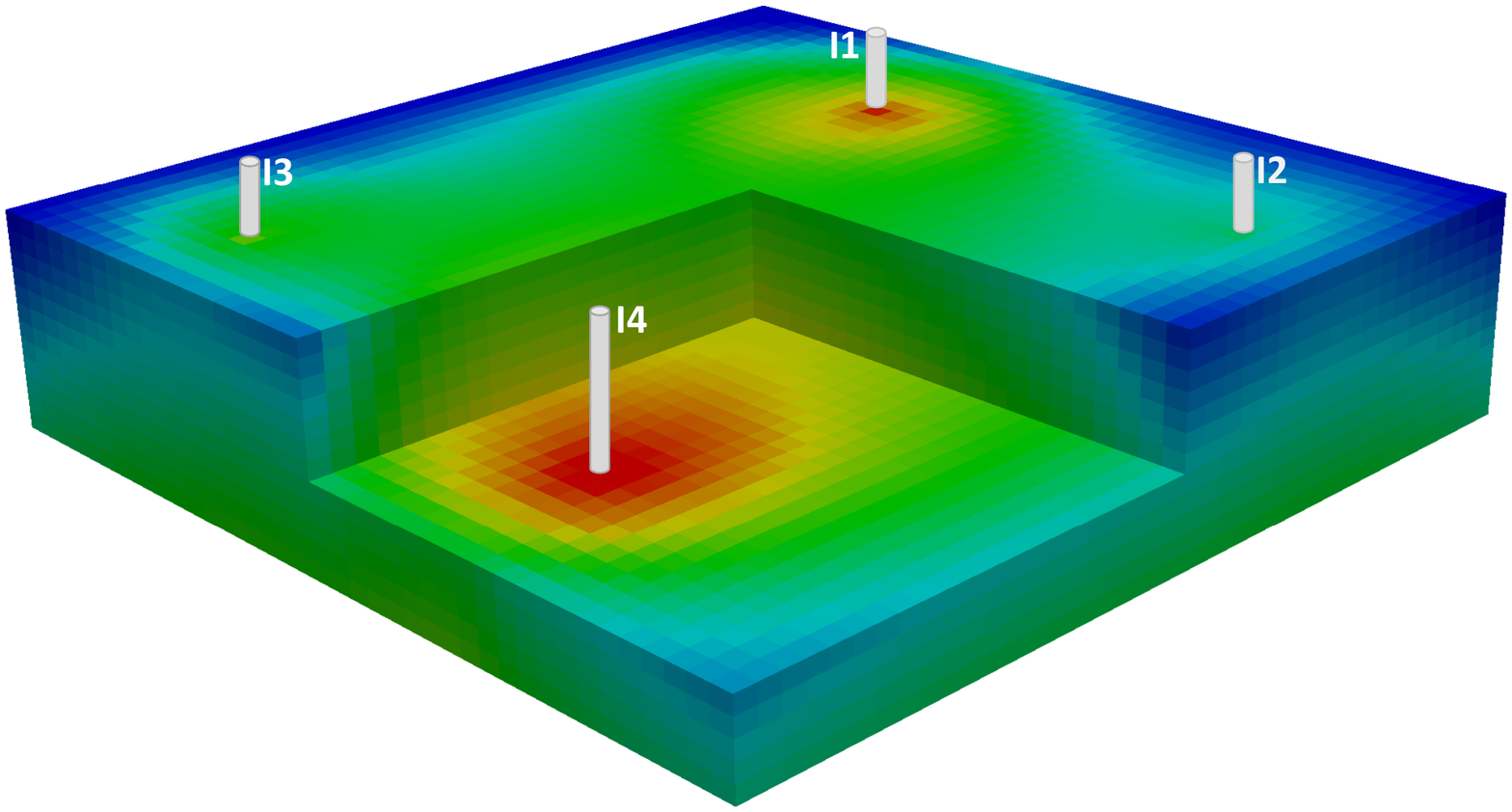}
         \caption{Realization 2 (surr)}
         \label{test-p-surr-case-2}
     \end{subfigure}
     %\hfill
      \begin{subfigure}[b]{0.32\textwidth}
         \centering
         \includegraphics[trim={0cm 0cm, 0cm, 0cm},clip,scale=0.22]{figs-ccs/fields-pdf/p-surr-case1-t3.pdf}
         \caption{Realization 3 (surr)}
         \label{test-p-surr-case-3}
     \end{subfigure}
          \begin{subfigure}[b]{0.32\textwidth}
         \centering
         \includegraphics[trim={0cm 0cm, 0cm, 0cm},clip,scale=0.22]{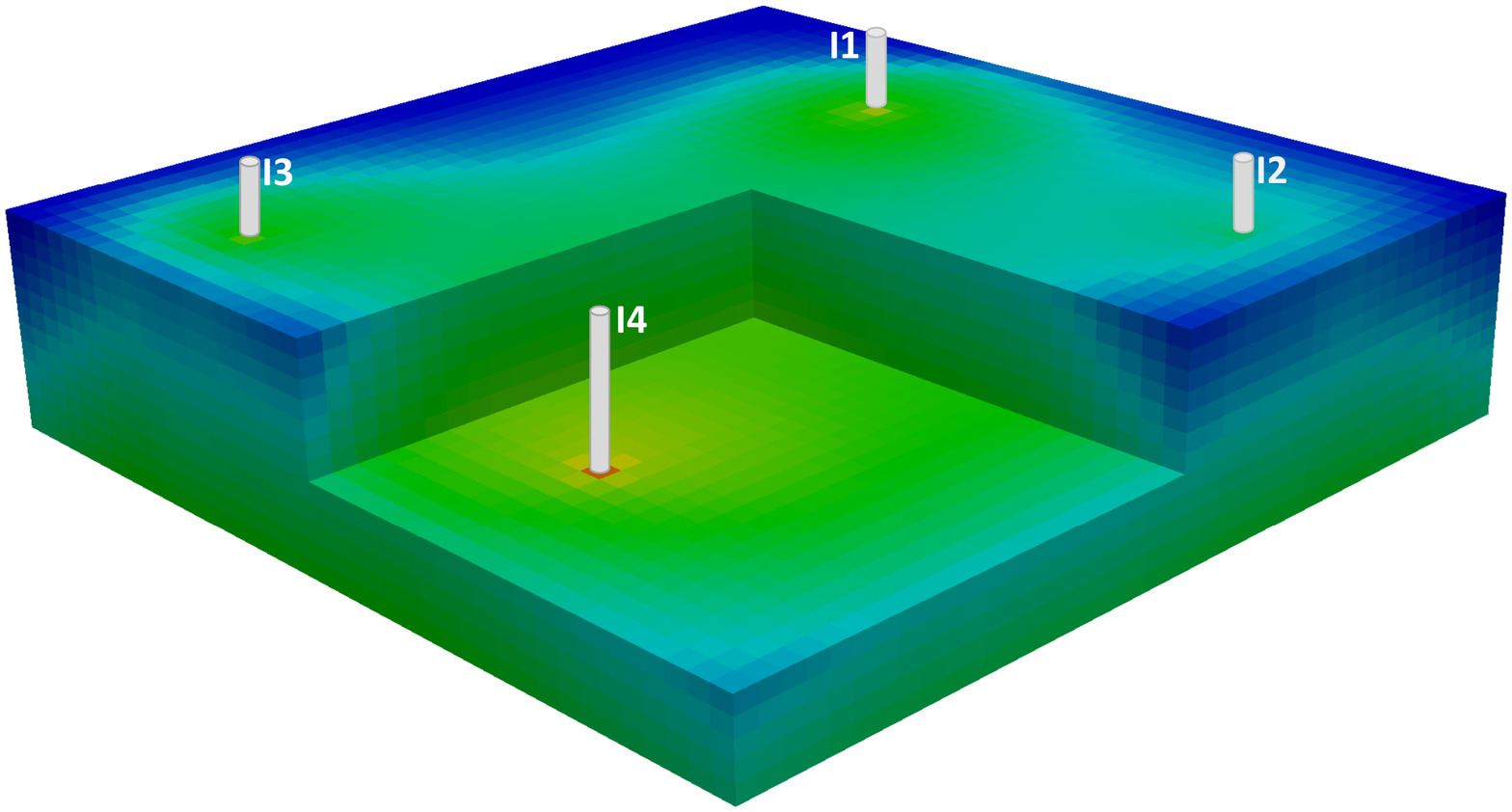}
         \caption{Realization 1 (sim)}
         \label{test-p-sim-case-1}
     \end{subfigure}
     %\hfill
    \begin{subfigure}[b]{0.32\textwidth}
         \centering
         \includegraphics[trim={0cm 0cm, 0cm, 0cm},clip,scale=0.22]{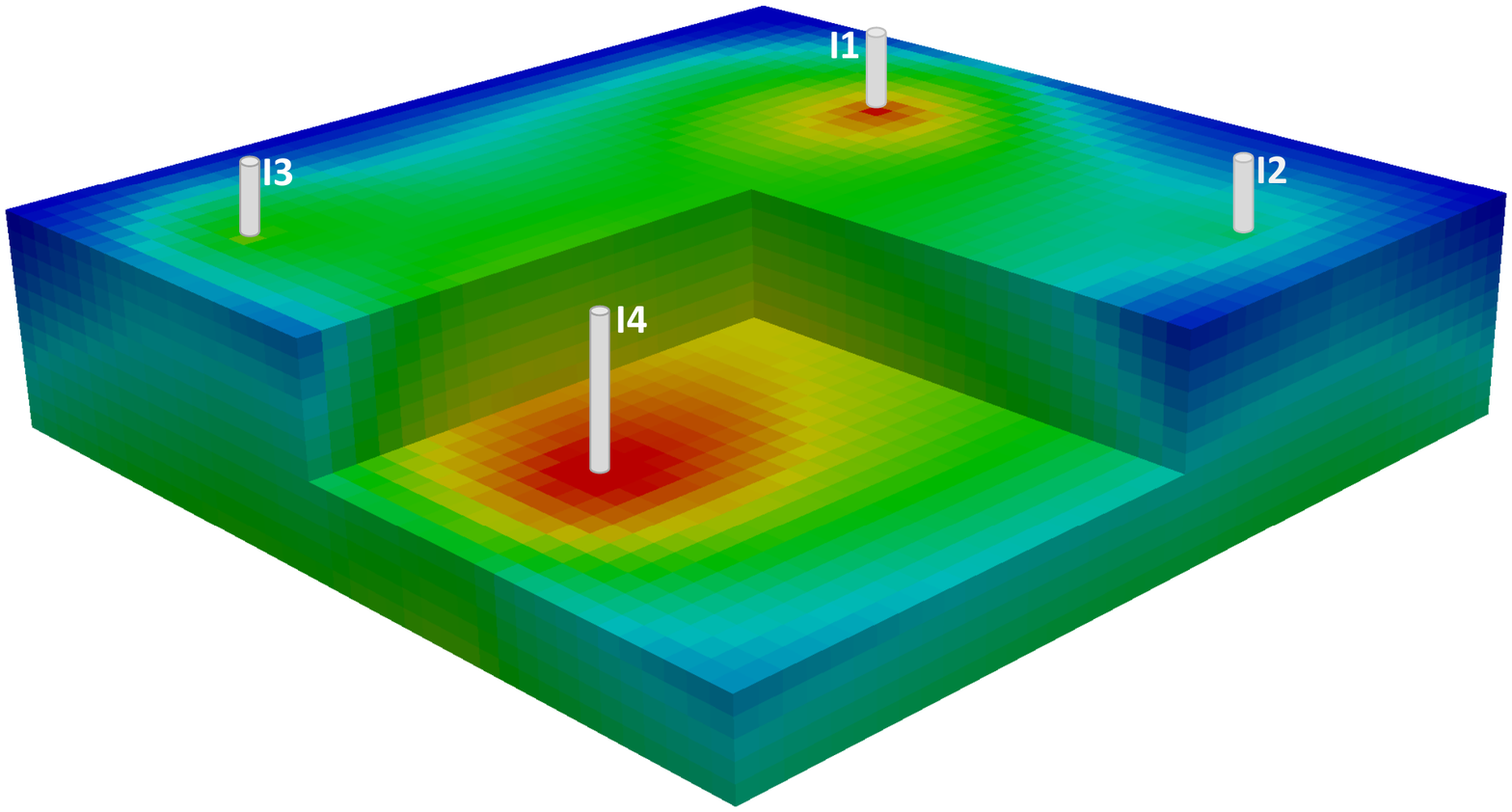}
         \caption{Realization 2 (sim)}
         \label{test-p-sim-case-2}
     \end{subfigure}
     %\hfill
      \begin{subfigure}[b]{0.32\textwidth}
         \centering
         \includegraphics[trim={0cm 0cm, 0cm, 0cm},clip,scale=0.22]{figs-ccs/fields-pdf/p-true-case1-t3.pdf}
         \caption{Realization 3 (sim)}
         \label{test-p-sim-case-3}

     \end{subfigure}
    
\caption{Pressure fields from recurrent R-U-Net surrogate model (upper row) and HFS (lower row) for three different test cases at 30~years.}
\label{fig:p-variability}
\end{figure}

Pressure fields at 30~years, for three different test-case realizations (with error near or above the median test-case error), are displayed in Fig.~\ref{fig:p-variability}. We see substantial differences from case to case in near-well pressure buildup, with Realization~2 displaying high pressure around wells I1 and I4. Despite the variability between and within realizations, consistent agreement between the surrogate and HFS results is again observed.

\begin{figure}
     \centering
     \begin{subfigure}[b]{0.32\textwidth}
         \centering
         \includegraphics[trim={3cm 0cm, 0cm, 0cm},clip,scale=0.42]{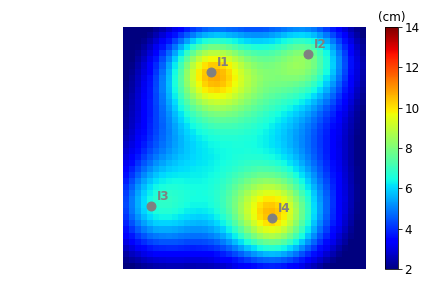}
         \caption{2 years (surr)}
         \label{test-disp-pred-case-1}
     \end{subfigure}
     %\hfill
    \begin{subfigure}[b]{0.32\textwidth}
         \centering
         \includegraphics[trim={3cm 0cm, 0cm, 0cm},clip,scale=0.42]{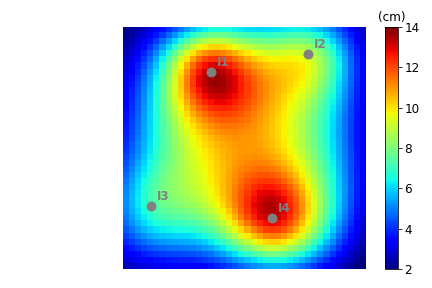}
         \caption{15 years (surr)}
         \label{test-disp-pred-case-1}
     \end{subfigure}
     %\hfill
      \begin{subfigure}[b]{0.32\textwidth}
         \centering
         \includegraphics[trim={3cm 0cm, 0cm, 0cm},clip,scale=0.42]{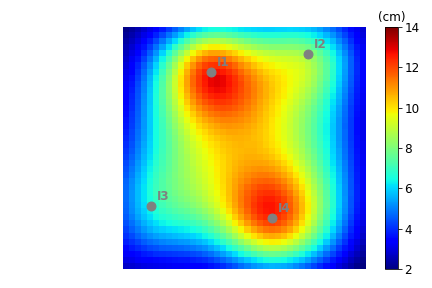}
         \caption{30 years (surr)}
         \label{test-disp-pred-case-1}
     \end{subfigure}
     
          \begin{subfigure}[b]{0.32\textwidth}
         \centering
         \includegraphics[trim={3cm 0cm, 0cm, 0cm},clip,scale=0.42]{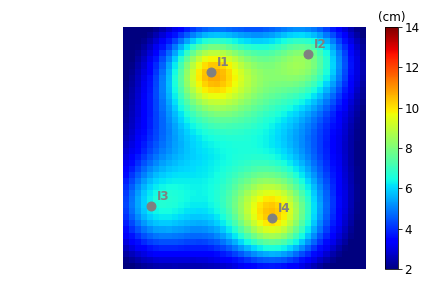}
         \caption{2 years (sim)}
         \label{test-disp-pred-case-1}
     \end{subfigure}
     %\hfill
    \begin{subfigure}[b]{0.32\textwidth}
         \centering
         \includegraphics[trim={3cm 0cm, 0cm, 0cm},clip,scale=0.42]{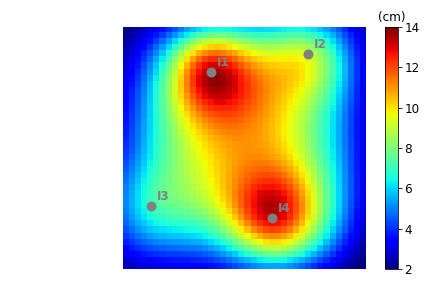}
         \caption{15 years (sim)}
         \label{test-disp-pred-case-1}
     \end{subfigure}
     %\hfill
      \begin{subfigure}[b]{0.32\textwidth}
         \centering
         \includegraphics[trim={3cm 0cm, 0cm, 0cm},clip,scale=0.42]{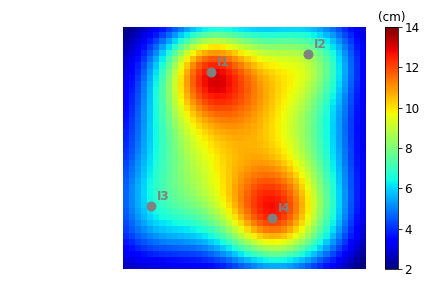}
         \caption{30 years (sim)}
         \label{test-disp-pred-case-1}
     \end{subfigure}
    
\caption{Vertical displacement maps from recurrent R-U-Net surrogate model (upper row) and HFS (lower row) for a test case at three different times. These results are for the same realization as in Fig.~\ref{fig:pressure-evolution}.}
    \label{fig:disp-evolution}
\end{figure}

As discussed in Section~\ref{sec:domains}, vertical displacements for the grid blocks lying directly above the storage aquifer, at the Earth's surface, are computed from the (nodal) finite element geomechanics solution. These are the quantities the surrogate model is trained to predict. Results for this ground-level vertical displacement are shown in Fig.~\ref{fig:disp-evolution}. These results are for the same realization (and time steps) as in  Fig.~\ref{fig:pressure-evolution} (which displayed pressure fields). The upper and lower rows again show the surrogate and HFS results, and we observe a high level of accuracy in the surrogate model predictions. The surface displacements reach a maximum of 14~cm, which is a substantial amount of deformation. Of particular interest is the correspondence between the displacement and pressure fields, evident by comparing Figs.~\ref{fig:pressure-evolution} and \ref{fig:disp-evolution}. Specifically, the maximum pressures are observed near wells I1 and I4, and it is above these wells that the maximum displacements occur. 

\begin{figure}
     \centering
     \begin{subfigure}[b]{0.32\textwidth}
         \centering
         \includegraphics[trim={3cm 0cm, 0cm, 0cm},clip,scale=0.42]{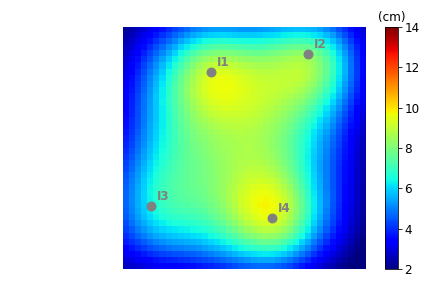}
         \caption{Realization 1 (surr)}
         \label{test-disp-pred-case-1}
     \end{subfigure}
     %\hfill
    \begin{subfigure}[b]{0.32\textwidth}
         \centering
         \includegraphics[trim={3cm 0cm, 0cm, 0cm},clip,scale=0.42]{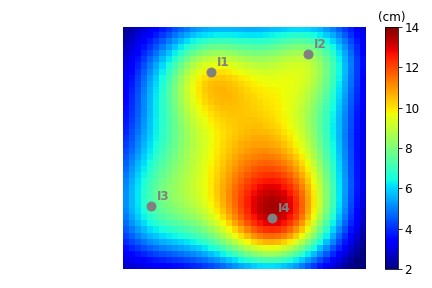}
         \caption{Realization 2 (surr)}
         \label{test-disp-pred-case-1}
     \end{subfigure}
     %\hfill
      \begin{subfigure}[b]{0.32\textwidth}
         \centering
         \includegraphics[trim={3cm 0cm, 0cm, 0cm},clip,scale=0.42]{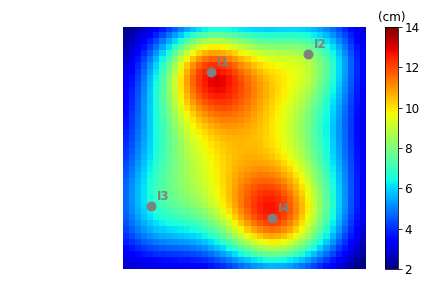}
         \caption{Realization 3 (surr)}
         \label{test-disp-pred-case-1}
     \end{subfigure}
     
          \begin{subfigure}[b]{0.32\textwidth}
         \centering
         \includegraphics[trim={3cm 0cm, 0cm, 0cm},clip,scale=0.42]{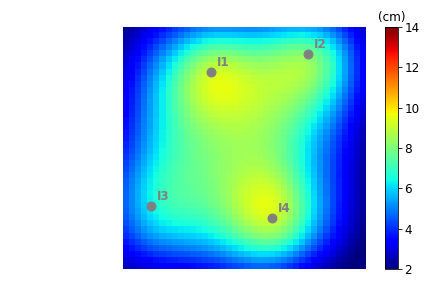}
         \caption{Realization 1 (sim)}
         \label{test-disp-pred-case-1}
     \end{subfigure}
     %\hfill
    \begin{subfigure}[b]{0.32\textwidth}
         \centering
         \includegraphics[trim={3cm 0cm, 0cm, 0cm},clip,scale=0.42]{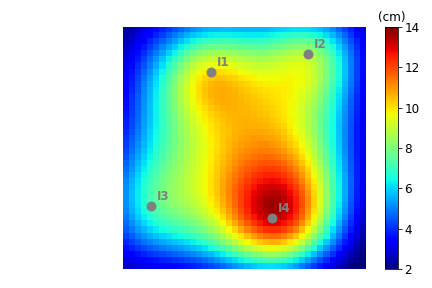}
         \caption{Realization 2 (sim)}
         \label{test-disp-pred-case-1}
     \end{subfigure}
     %\hfill
      \begin{subfigure}[b]{0.32\textwidth}
         \centering
         \includegraphics[trim={3cm 0cm, 0cm, 0cm},clip,scale=0.42]{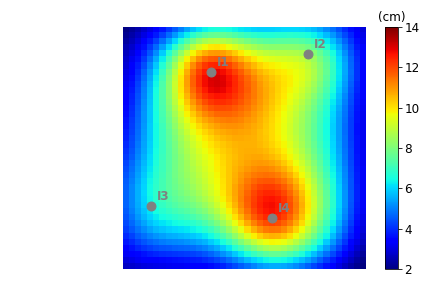}
         \caption{Realization 3 (sim)}
         \label{test-disp-pred-case-1}
     \end{subfigure}
    
\caption{Vertical displacement maps from recurrent R-U-Net surrogate model (upper row) and HFS (lower row) for three different test cases at 30~years. These results are for the same realizations as in Fig.~\ref{fig:p-variability}.}
    \label{fig:disp-variability}
\end{figure}

Surface displacement results for the three test-case models presented in Fig.~\ref{fig:p-variability} (for pressure) are shown in Fig.~\ref{fig:disp-variability}. These results are all at a time of 30~years. Despite the significant differences between the various cases, we continue to observe close correspondence between the surrogate model and HFS results. The highest near-well pressures are observed for well~I4 in Realization~2, and the largest vertical displacements are also seen above this well for this realization. Analogously, lower near-well pressures and surface displacements are evident for Realization~1.

\subsection{Surrogate model errors and percentile predictions}

We now quantify the saturation, pressure and surface displacement errors associated with the surrogate model predictions. We compute relative errors for these three quantities, defined as
\begin{equation}
{\delta_S = \frac{1}{n_{e}n_s n_t}\sum_{i=1}^{n_{e}}\sum_{j=1}^{n_s}\sum_{t=1}^{n_t} \frac{\norm{{(\hat{S}_s)}_{i, j}^{t} - {(S_s)}_{i,j}^{t}}}{{(S_s)}_{i,j}^{t} + \epsilon}},
\label{eq:sat-relative-error}
\end{equation}

\begin{equation}
{\delta_p = \frac{1}{n_{e}n_s n_t}\sum_{i=1}^{n_{e}}\sum_{j=1}^{n_s}\sum_{t=1}^{n_t} \frac{\norm{(\hat{p}_s)_{i,j}^{t} - (p_s)_{i,j}^{t}}}{(p_s)_{i,\text{max}}^t - (p_s)_{i,\text{min}}^t}},
\label{eq:pressure-relative-error}
\end{equation}

\begin{equation}
{\delta_d = \frac{1}{n_{e}n_{gb} n_t}\sum_{i=1}^{n_{e}}\sum_{j=1}^{n_{gb}}\sum_{t=1}^{n_t} \frac{\norm{{(\hat{d}_{gb})}_{i, j}^{t} - {(d_{gb})}_{i,j}^{t}}}{{(d_{gb})}_{i,j}^{t}}},
\label{eq:disp-relative-error}
\end{equation}
where $n_e$ is the number of test cases, $n_s$ is the total number of grid blocks in the storage aquifer, $n_t$ is the number of time steps considered, and $n_{gb}$ is the number of grid blocks on the surface above the storage aquifer. A constant $\epsilon = 0.01$ is introduced in the denominator of Eq.~\ref{eq:sat-relative-error} to avoid division by very small values. Note that the pressure error is normalized by the difference between the maximum and minimum pressure values for test case $i$ at time step $t$ ($(p_s)_{i,\text{max}}^t - (p_s)_{i,\text{min}}^t$). This treatment leads to larger pressure errors than would be computed if we used the absolute pressure value $(p_s)_{i,j}^t$.

Applying Eqs.~\ref{eq:sat-relative-error}, \ref{eq:pressure-relative-error} and \ref{eq:disp-relative-error}, we obtain $\delta_S = 5.3\%$, $\delta_p = 0.31\%$ and $\delta_d = 1.2\%$. These small relative errors demonstrate that the recurrent R-U-Net surrogate predicts storage aquifer saturation and pressure, along with surface-level vertical displacement, quite accurately. The observation that $\delta_S>\delta_p$ is consistent with our previous findings \citep{tang2020jcp, tang2021deep}. Saturation errors are generally largest in the vicinity of fronts.

%In data assimilation settings, the mismatch between measured data and simulation results at particular locations is used to evaluate candidate realizations. 

\begin{figure}[htbp]
  \centering
  \includegraphics[trim={1cm 3cm 3cm 3cm},clip, scale=0.35]{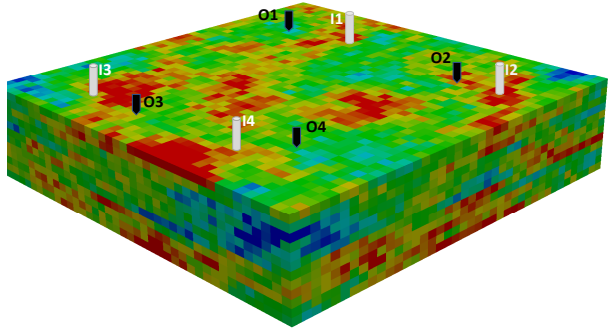}
   \caption{Storage aquifer realization showing injection wells I1 -- I4 (white cylinders) and observation locations O1 -- O4 (black arrows).}
  \label{fig:reservoir-obs-loc}
\end{figure}

We next assess the statistical correspondence in block-wise results between the surrogate and HFS models. This evaluation enables us to see whether the surrogate model maintains accuracy over a range of possible test-case responses. Specifically, for four observation locations in the top layer of the storage aquifer (shown in Fig.~\ref{fig:reservoir-obs-loc} and denoted O1 -- O4), we compute the 10th, 50th and 90th percentile results for saturation and pressure as a function of time. These quantities are referred to as the P$_{10}$, P$_{50}$ and P$_{90}$ responses. Each curve can correspond to a different realization from time step to time step.

\begin{figure}
     \centering
     \begin{subfigure}[b]{0.4\textwidth}
         \centering
         \includegraphics[width=\textwidth]{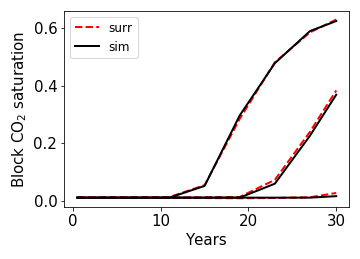}
         \caption{Saturation at O1}
         \label{pfs-orate-w1}
     \end{subfigure}
     %\hfill
     \begin{subfigure}[b]{0.4\textwidth}
         \centering
         \includegraphics[width=\textwidth]{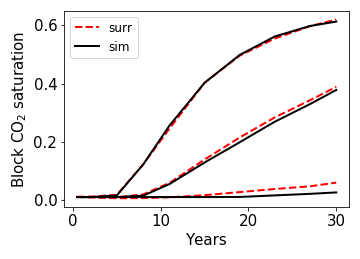}
         \caption{Saturation at O2}
         \label{pfs-wrate-w1}
     \end{subfigure}
     %\hfill
     
     \begin{subfigure}[b]{0.4\textwidth}
         \centering
         \includegraphics[width=\textwidth]{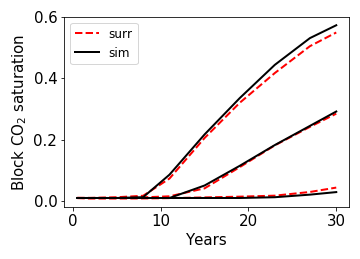}
         \caption{Saturation at O3}
         \label{pfs-orate-w2}
     \end{subfigure}
     %\hspace{1mm}
     \begin{subfigure}[b]{0.4\textwidth}
         \centering
         \includegraphics[width=\textwidth]{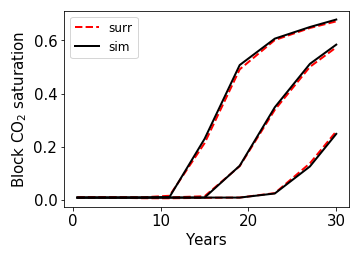}
         \caption{Saturation at O4}
         \label{pfs-wrate-w15}
     \end{subfigure}
    
\caption{Comparison of saturation statistics at four observation locations over the full ensemble of 500 test cases. Red and black curves represent $\text{P}_{10}$ (lower), $\text{P}_{50}$ (middle) and $\text{P}_{90}$ (upper) results from the recurrent R-U-Net surrogate model and the high-fidelity simulator, respectively.}
% [xxx1 -- the font size on the axis numbers in (a) is different than in (b)-(d) -- please fix this]}
    \label{fig:saturation-statistics}
\end{figure}

\begin{figure}
     \centering
     \begin{subfigure}[b]{0.4\textwidth}
         \centering
         \includegraphics[width=\textwidth]{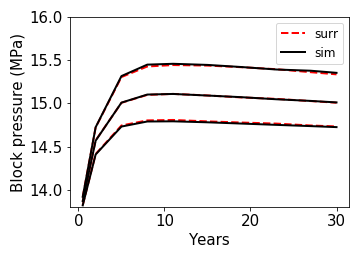}
         \caption{Pressure at O1}
         \label{pfs-orate-w1}
     \end{subfigure}
     %\hfill
     \begin{subfigure}[b]{0.4\textwidth}
         \centering
         \includegraphics[width=\textwidth]{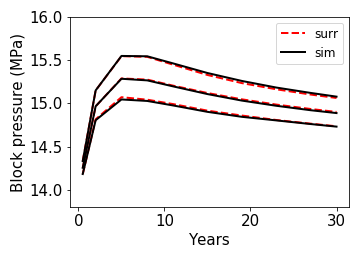}
         \caption{Pressure at O2}
         \label{pfs-wrate-w1}
     \end{subfigure}
     %\hfill
     
     \begin{subfigure}[b]{0.4\textwidth}
         \centering
         \includegraphics[width=\textwidth]{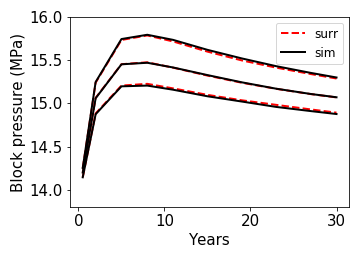}
         \caption{Pressure at O3}
         \label{pfs-orate-w2}
     \end{subfigure}
     %\hspace{1mm}
     \begin{subfigure}[b]{0.4\textwidth}
         \centering
         \includegraphics[width=\textwidth]{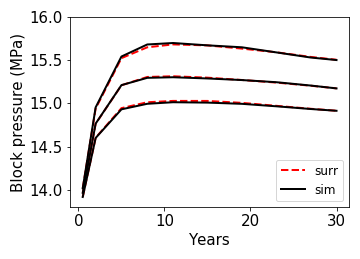}
         \caption{Pressure at O4}
         \label{pfs-wrate-w15}
     \end{subfigure}
    
\caption{Comparison of pressure statistics at four observation locations over the full ensemble of 500 test cases. Red and black curves represent $\text{P}_{10}$ (lower), $\text{P}_{50}$ (middle) and $\text{P}_{90}$ (upper) results from the recurrent R-U-Net surrogate model and the high-fidelity simulator, respectively.}
    \label{fig:pressure-statistics}
\end{figure}

\begin{figure}
     \centering
     \begin{subfigure}[b]{0.4\textwidth}
         \centering
         \includegraphics[width=\textwidth]{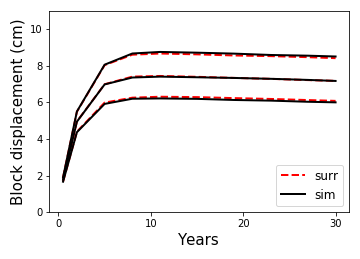}
         \caption{Displacement at the surface above O1}
         \label{pfs-orate-w1}
     \end{subfigure}
     %\hfill
     \begin{subfigure}[b]{0.4\textwidth}
         \centering
         \includegraphics[width=\textwidth]{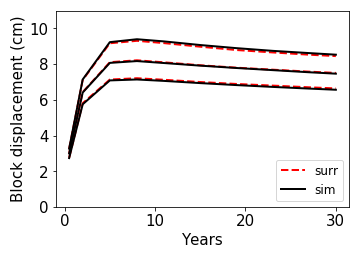}
         \caption{Displacement at the surface above O2}
         \label{pfs-wrate-w1}
     \end{subfigure}
     %\hfill
     
     \begin{subfigure}[b]{0.4\textwidth}
         \centering
         \includegraphics[width=\textwidth]{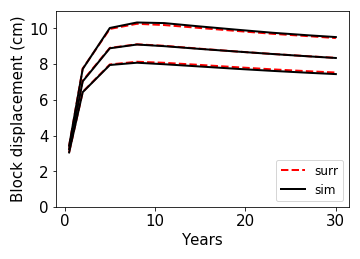}
         \caption{Displacement at the surface above O3}
         \label{pfs-orate-w2}
     \end{subfigure}
     %\hspace{1mm}
     \begin{subfigure}[b]{0.4\textwidth}
         \centering
         \includegraphics[width=\textwidth]{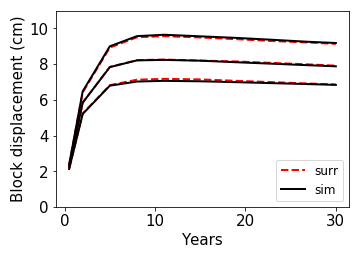}
         \caption{Displacement at the surface above O4}
         \label{pfs-wrate-w15}
     \end{subfigure}
    
\caption{Comparison of vertical displacement statistics at four observation locations over the full ensemble of 500 test cases. Red and black curves represent $\text{P}_{10}$ (lower), $\text{P}_{50}$ (middle) and $\text{P}_{90}$ (upper) results from the recurrent R-U-Net surrogate model and the high-fidelity simulator, respectively.}
    \label{fig:disp-statistics}
\end{figure}

Saturation and pressure results are presented in Figs.~\ref{fig:saturation-statistics} and \ref{fig:pressure-statistics}. The solid black curves are the target HFS results, and the dashed red curves are the surrogate model results. The lower curve represents the P$_{10}$ result, the middle curve is P$_{50}$, and the upper curve P$_{90}$. Agreement between the HFS and surrogate model P$_{10}$, P$_{50}$ and P$_{90}$ results, for both saturation and pressure at all four observation locations, is excellent. This suggests that the surrogate model is accurate over the full P$_{10}$--P$_{90}$ range for these  quantities. The consistent agreement in saturation results, despite the wide variability for this quantity (evident in the difference between the P$_{10}$ and P$_{90}$ curves in Fig.~\ref{fig:saturation-statistics}), is particularly noteworthy. 

For surface displacement, the observation locations are at ground level, directly above the O1 -- O4 storage-aquifer locations. The P$_{10}$, P$_{50}$ and P$_{90}$ results at the four observation locations are shown in Fig.~\ref{fig:disp-statistics}. Agreement between HFS and surrogate predictions is again very close, though there is relatively little variation between the P$_{10}$ and P$_{90}$ surface displacement curves. Nonetheless, as we will see in Section~\ref{sec:hm}, data of this type can still be used in history matching to reduce prediction uncertainty.

\section{Data Assimilation using Recurrent R-U-Net Surrogate Model}
\label{sec:hm}

In this section we apply the recurrent R-U-Net surrogate model for data assimilation. A rigorous rejection sampling (RS) procedure is applied to provide posterior (history matched) samples conditioned to surface displacement data. We first describe the RS procedure, and then present data assimilation results.

\subsection{Rejection sampling using PCA-based geomodels}
\label{sec:RS}
RS is a formal posterior sampling method that can correctly quantify posterior uncertainty. It has the drawback of requiring very large numbers of samples and function evaluations, which renders it impractical in most settings when HFS is used for the function evaluations. With the surrogate model applied in place of the high-fidelity simulator, however, RS can be applied in a much wider range of cases.

In our example, we construct and evaluate 500,000 realizations of the geomodel. Rather than apply the geological modeling software SGeMS to generate each realization, we construct these models very quickly using principal component analysis (PCA). Following the description in \citep{liu20203d}, we construct the PCA basis matrix, in a preprocessing step, from a set of 3000 SGeMS models. Then, new PCA realizations $\mathbf{m}_s^{\text{pca}} \in \mathbb{R}^{n_s}$ are constructed through application of $\mathbf{m}_s^{\text{pca}}=\boldsymbol\Phi \boldsymbol\xi + {\bar{\mathbf{m}}}_s$, where $\boldsymbol\Phi \in \mathbb{R}^{n_s \times l}$ is the basis matrix, ${\bar{\mathbf{m}}}_s \in \mathbb{R}^{n_s}$ is the mean of the SGeMS models, and $\boldsymbol\xi \in \mathbb{R}^l$ is a random variable, with each component sampled independently from $\mathcal{N}(0,1)$. The dimension of $\boldsymbol\xi$ ($l$) is set to 2000, which acts to retain 95\% of the total `energy' in the original set of 3000 realizations. This provides $\mathbf{m}_s^{\text{pca}}$ realizations that are essentially indistinguishable visually from SGeMS models while avoiding overfitting.

Note that, in \citep{tang2020jcp,tang2021deep}, a convolutional neural network (CNN)-based parameterization technique (called CNN-PCA) was used for realization generation instead of standalone PCA. CNN-PCA was required in those studies because the geomodels were channelized, and standalone PCA is not applicable for such systems. We can, however, directly apply PCA here since the geomodels are multi-Gaussian.

%The trained recurrent R-U-Nets are then applied to predict ground surface displacement, pressure and saturation fields of storage aquifer for half million PCA realizations. 

We now present the RS procedure applied in this work. This description follows that in \citep{tang2021deep}, though here we use PCA instead of CNN-PCA, and the observed data are surface displacements rather than flow rates at wells.

\begin{itemize}
    \item Sample each component of $\boldsymbol\xi \in \mathbb{R}^l$ from $\mathcal{N}(0,1)$. Construct $\mathbf{m}_s^{\text{pca}}(\boldsymbol\xi)$ as described above. 
    \item Sample a probability $p$ from a uniform distribution in $[0, 1]$.
    \item Compute the likelihood function $L(\mathbf{m}_s^{\text{pca}}(\boldsymbol\xi))$ using
    \begin{equation}
        L(\mathbf{m}_s^{\text{pca}}(\boldsymbol\xi)) = c \exp \left(-\frac{1}{2}[\hat{f}(\mathbf{m}_s^{\text{pca}}(\boldsymbol\xi)) - \mathbf{d}_{\text{obs}}]^T C_D^{-1} [\hat{f}(\mathbf{m}_s^{\text{pca}}(\boldsymbol\xi)) - \mathbf{d}_{\text{obs}}]\right),
    \end{equation}
    where $c$ is a normalization constant, $\hat{f}(\mathbf{m}_s^{\text{pca}}(\boldsymbol\xi))$ indicates the surrogate model predictions for surface displacement for geomodel $\mathbf{m}_s^{\text{pca}}(\boldsymbol\xi)$,
    $\mathbf{d}_{\text{obs}}$ denotes the observed surface displacement data, and $C_D$ is the covariance matrix of data measurement error. 
    \item Accept $\mathbf{m}_s^{\text{pca}}(\boldsymbol\xi)$ if $p \leq \frac{L(\mathbf{m}_{\text{pca}}(\boldsymbol\xi))}{S_L}$, where $S_L$ is the maximum likelihood value over all prior models considered. Otherwise, reject $\mathbf{m}_s^{\text{pca}}(\boldsymbol\xi)$.
\end{itemize}
As is evident from the description above, a randomly generated realization $\mathbf{m}_s^{\text{pca}}(\boldsymbol\xi)$ is much more likely to be accepted as a posterior sample if the predicted surface displacements are close to $\mathbf{d}_{\text{obs}}$. Similarly, if the mismatch is large, the model has a very low probability of acceptance.

\subsection{Problem setup and rejection sampling results}

A randomly selected SGeMS gemodel, which is shown in Fig.~\ref{fig:sgems-gaussian-reals}(c), is specified to be the `true' model. High-fidelity simulation is performed on this model, and the results are taken as the `true' data $\mathbf{d}_{\text{true}}$. The observed data $\mathbf{d}_{\text{obs}}$ are obtained by randomly perturbing $\mathbf{d}_{\text{true}}$ with measurement error $\boldsymbol\epsilon$
\begin{equation}
\mathbf{d}_{\text{obs}} = \mathbf{d}_{\text{true}} + \boldsymbol\epsilon,
\end{equation}
where $\boldsymbol\epsilon$ is sampled with $\mathbf{0}$ mean and covariance $C_D$. In this study the observed data correspond to surface displacement data at four observation locations (above O1 -- O4) at two different times (5~years and 8~years after the start of CO$_2$ injection). Thus we have a total of eight measurements. The standard deviation of the measurement error is set to 5\% of the corresponding true data. 

\begin{figure}
     \centering
     \begin{subfigure}[b]{0.45\textwidth}
         \centering
         \includegraphics[width=\textwidth]{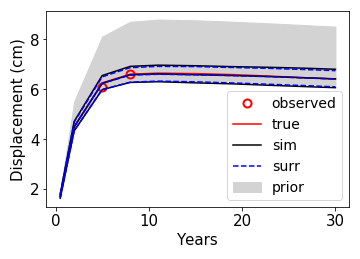}
         \caption{Displacement at the surface above O1}
         \label{hm-orate-w1}
     \end{subfigure}
     %\hfill
     \begin{subfigure}[b]{0.45\textwidth}
         \centering
          \includegraphics[width=\textwidth]{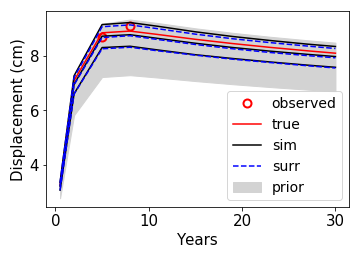}
         \caption{Displacement at the surface above O2}
         \label{hm-wrate-w1}
     \end{subfigure}
     %\hfill
     
     \begin{subfigure}[b]{0.45\textwidth}
         \centering
         \includegraphics[width=\textwidth]{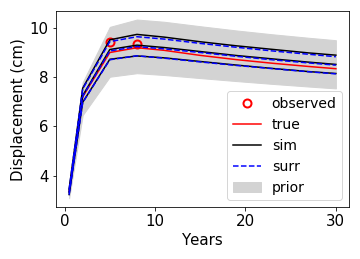}
         \caption{Displacement at the surface above O3}
         \label{hm-orate-w14}
     \end{subfigure}
     %\hspace{1mm}
     \begin{subfigure}[b]{0.45\textwidth}
         \centering
          \includegraphics[width=\textwidth]{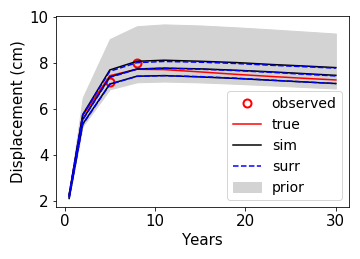}
         \caption{Displacement at the surface above O4}
         \label{hm-wrate-w14}
     \end{subfigure}
     
    \caption{History matching results for surface displacement at four observation locations. Gray regions represent the prior $\text{P}_{10}$--$\text{P}_{90}$ range, red points and red lines denote observed and true data, solid black and dashed blue curves denote the posterior $\text{P}_{10}$ (lower), $\text{P}_{50}$ (middle) and $\text{P}_{90}$ (upper) predictions obtained using the high-fidelity simulator (black) and the surrogate (blue).}
    \label{fig:hm-disp}
\end{figure}

\begin{figure}
     \centering
     \begin{subfigure}[b]{0.45\textwidth}
         \centering
         \includegraphics[width=\textwidth]{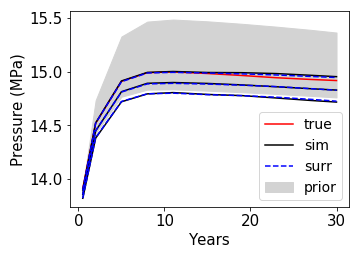}
         \caption{Pressure at O1}
         \label{hm-orate-w1}
     \end{subfigure}
     %\hfill
     \begin{subfigure}[b]{0.45\textwidth}
         \centering
          \includegraphics[width=\textwidth]{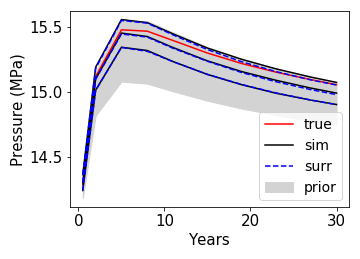}
         \caption{Pressure at O2}
         \label{hm-wrate-w1}
     \end{subfigure}
     %\hfill
     
     \begin{subfigure}[b]{0.45\textwidth}
         \centering
         \includegraphics[width=\textwidth]{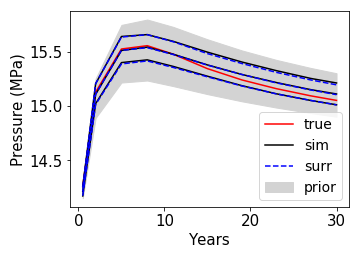}
         \caption{Pressure at O3}
         \label{hm-orate-w14}
     \end{subfigure}
     %\hspace{1mm}
     \begin{subfigure}[b]{0.45\textwidth}
         \centering
          \includegraphics[width=\textwidth]{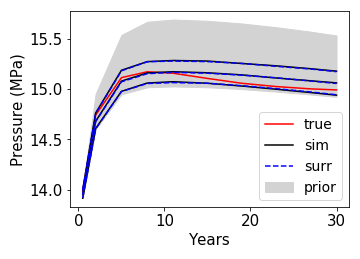}
         \caption{Pressure at O4}
         \label{hm-wrate-w14}
     \end{subfigure}
     
    \caption{History matching results for pressure at observation locations O1 -- O4. Note that only surface displacement data are used in the data assimilation. Gray regions represent the prior $\text{P}_{10}$--$\text{P}_{90}$ range, red lines denote true data, solid black and dashed blue curves denote the posterior $\text{P}_{10}$ (lower), $\text{P}_{50}$ (middle) and $\text{P}_{90}$ (upper) predictions obtained using the high-fidelity simulator (black) and the surrogate (blue).}
    \label{fig:hm-p}
\end{figure}

\begin{figure}
     \centering
     \begin{subfigure}[b]{0.45\textwidth}
         \centering
         \includegraphics[width=\textwidth]{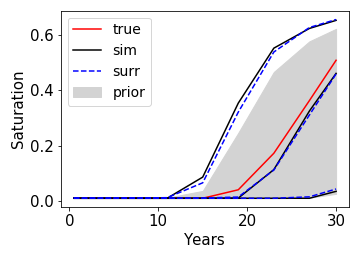}
         \caption{Saturation at O1}
         \label{hm-orate-w1}
     \end{subfigure}
     %\hfill
     \begin{subfigure}[b]{0.45\textwidth}
         \centering
          \includegraphics[width=\textwidth]{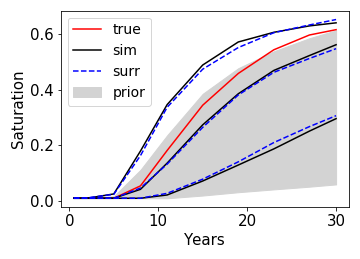}
         \caption{Saturation at O2}
         \label{hm-wrate-w1}
     \end{subfigure}
     %\hfill
     
     \begin{subfigure}[b]{0.45\textwidth}
         \centering
         \includegraphics[width=\textwidth]{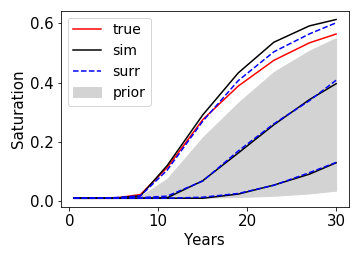}
         \caption{Saturation at O3}
         \label{hm-orate-w14}
     \end{subfigure}
     %\hspace{1mm}
     \begin{subfigure}[b]{0.45\textwidth}
         \centering
          \includegraphics[width=\textwidth]{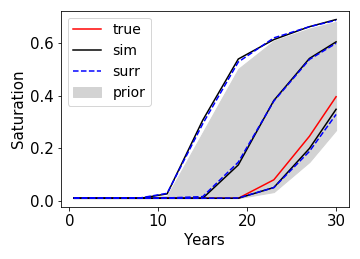}
         \caption{Saturation at O4}
         \label{hm-wrate-w14}
     \end{subfigure}
    \caption{History matching results for saturation at observation locations O1 -- O4. Note that only surface displacement data are used in the data assimilation. Gray regions represent the prior $\text{P}_{10}$--$\text{P}_{90}$ range, red lines denote true data, solid black and dashed blue curves denote the posterior $\text{P}_{10}$ (lower), $\text{P}_{50}$ (middle) and $\text{P}_{90}$ (upper) predictions obtained using the high-fidelity simulator (black) and the surrogate (blue).}
    \label{fig:hm-sat}
\end{figure}

Using a single Nvidia Tesla V100 GPU, it takes less than 1~hour for the surrogate model to provide the saturation, pressure and surface displacement predictions for the 500,000 geomodels evaluated during the RS procedure. The time required to generate the PCA realizations is negligible. High-fidelity numerical simulation, however, requires about 0.8~hours per run (using 32 CPU cores). Therefore, the total computation time to perform this RS assessment using HFS would be about 400,000~hours, which is clearly impractical.

With the setup described above, RS accepts 601 models out of the 500,000 considered. Data assimilation results for (ground-level) surface displacement are shown in Fig.~\ref{fig:hm-disp}. In this figure, the gray regions indicate the prior $\text{P}_{10}$-$\text{P}_{90}$ intervals, the red circles denote the observed data, and the red curves indicate the surface displacement response for the true model (the observed and true data deviate slightly due to measurement error $\boldsymbol\epsilon$). The blue dashed curves depict the $\text{P}_{10}$ (lower), $\text{P}_{50}$ (middle) and $\text{P}_{90}$ (upper) posterior RS results using the surrogate model. We also perform high-fidelity simulations on the 601 accepted geomodels. The resulting $\text{P}_{10}$, $\text{P}_{50}$ and $\text{P}_{90}$ curves are indicated by the solid black curves in Fig.~\ref{fig:hm-disp}. 

The results in Fig.~\ref{fig:hm-disp} demonstrate uncertainty reduction in surface displacement predictions at the four observation locations. The prior uncertainty above location O3 is relatively small, however, and is only modestly reduced by the data assimilation procedure. We observe close correspondence between the $\text{P}_{10}$, $\text{P}_{50}$ and $\text{P}_{90}$ predictions from the surrogate model (blue dashed curves) and those from the high-fidelity simulations (solid black curves). This is an important consistency, and suggests that the surrogate model is indeed appropriate for the many function evaluations required by RS.

Of particular interest in carbon storage operations are the pressure buildup at the top of the storage formation (caprock) and the location of the CO$_2$ plume. We now assess the degree to which uncertainty in these quantities is reduced through assimilation of surface displacement data. We reiterate that no data other than the eight surface displacement measurements are used for history matching. Results for prior and posterior pressure and saturation data at locations O1 -- O4 are shown in Figs.~\ref{fig:hm-p} and \ref{fig:hm-sat}. The surface displacement data are clearly informative in terms of pressure buildup, as we see substantial uncertainty reduction at all locations except O3 (consistent with Fig.~\ref{fig:hm-disp}). The surface displacement data are not very informative for saturation, and we see little uncertainty reduction in Fig.~\ref{fig:hm-sat}. We do, however, observe that the posterior distributions for saturation at locations O2 and O3 have shifted towards the true result (red curve), and that the true response is captured within the posterior $\text{P}_{10}$--$\text{P}_{90}$ interval. Finally, consistent with our observations in Fig.~\ref{fig:hm-disp}, the surrogate and HFS $\text{P}_{10}$, $\text{P}_{50}$ and $\text{P}_{90}$ curves continue to agree closely in Figs.~\ref{fig:hm-p} and \ref{fig:hm-sat}.

\begin{figure}
     \centering
     \begin{subfigure}[b]{0.32\textwidth}
         \centering
         \includegraphics[trim={3cm 0cm, 0cm, 0cm},clip,scale=0.42]{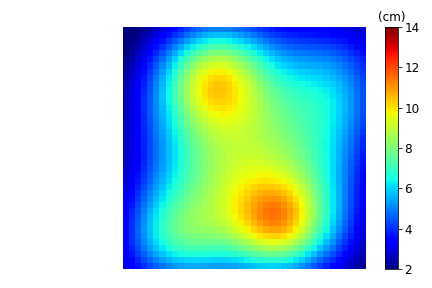}
         \caption{Prior P$_\text{10}$}
         \label{test-disp-pred-case-1}
     \end{subfigure}
     %\hfill
    \begin{subfigure}[b]{0.32\textwidth}
         \centering
         \includegraphics[trim={3cm 0cm, 0cm, 0cm},clip,scale=0.42]{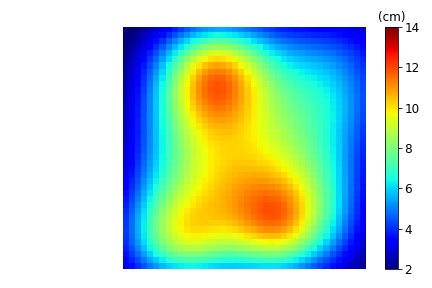}
         \caption{Prior P$_\text{50}$}
         \label{test-disp-pred-case-1}
     \end{subfigure}
     %\hfill
      \begin{subfigure}[b]{0.32\textwidth}
         \centering
         \includegraphics[trim={3cm 0cm, 0cm, 0cm},clip,scale=0.42]{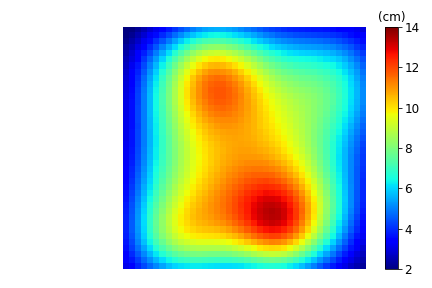}
         \caption{Prior P$_\text{90}$}
         \label{test-disp-pred-case-1}
     \end{subfigure}
     
          \begin{subfigure}[b]{0.32\textwidth}
         \centering
         \includegraphics[trim={3cm 0cm, 0cm, 0cm},clip,scale=0.42]{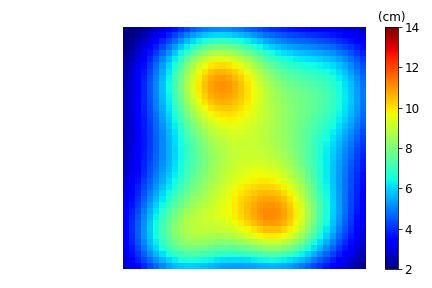}
         \caption{Posterior P$_\text{10}$}
         \label{test-disp-pred-case-1}
     \end{subfigure}
     %\hfill
    \begin{subfigure}[b]{0.32\textwidth}
         \centering
         \includegraphics[trim={3cm 0cm, 0cm, 0cm},clip,scale=0.42]{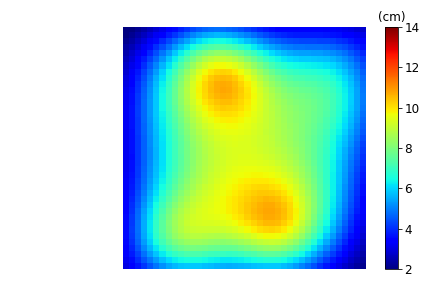}
         \caption{Posterior P$_\text{50}$}
         \label{test-disp-pred-case-1}
     \end{subfigure}
     %\hfill
      \begin{subfigure}[b]{0.32\textwidth}
         \centering
         \includegraphics[trim={3cm 0cm, 0cm, 0cm},clip,scale=0.42]{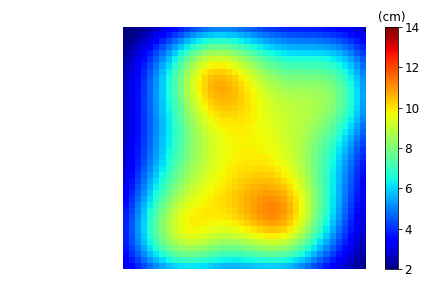}
         \caption{Posterior P$_\text{90}$}
         \label{test-disp-pred-case-1}
     \end{subfigure}
    
\caption{Prior (upper row) and posterior (lower row) P$_\text{10}$, P$_\text{50}$ and P$_\text{90}$ surface displacement maps from the surrogate model. Percentiles are based on average displacement at the surface, with the averages over locations directly above O1 -- O4, at 30~years.}
    \label{fig:prior-posterior-disp-maps}
\end{figure}

Finally, in Fig.~\ref{fig:prior-posterior-disp-maps}, we present surface displacement maps for prior and posterior models. The particular models displayed correspond to the P$_{10}$, P$_{50}$ and P$_{90}$ responses for average vertical displacement, with the average computed from the surface displacements above the four observation locations, at 30~years. In the upper row, which corresponds to prior results, we observe significant discrepancies between the three maps. The three maps on the lower row, however, are much more similar to one another. This clearly illustrates the reduction in uncertainty achieved by assimilating the surface displacement data.

\section{Concluding Remarks}
\label{sec:conl}

In this study, we extended our 3D recurrent R-U-Net surrogate model, originally developed for two-phase subsurface flow, to treat systems involving coupled flow and geomechanics. The methodology was then applied for the simulation of CO$_2$ storage operations. The recurrent R-U-Net involves the use of residual U-Nets and a convolutional long-short term memory network. The 3D solution domain for the full-order problem includes a storage aquifer, a large surrounding region, overburden, and bedrock. The storage-aquifer geomodel is viewed as uncertain, with grid-block porosity and log-permeability characterized by multi-Gaussian random fields. An advantage of the surrogate model is that it can be trained to predict key quantities only in regions of interest. In this work the recurrent R-U-Net was trained to predict CO$_2$ saturation (i.e., plume location) and pressure in the storage aquifer, and vertical displacement at the Earth's surface.

The problem setup entailed injection of about 4~Mta of supercritical CO$_2$ via four vertical injection wells. The overall model domain contained 133,200 cells, with the storage aquifer defined by 19,200 cells. A total of 2000 full-order models, each requiring 0.8~hours of simulation time on a 32-core CPU, were used for training. Following training, predictions of key quantities for new geomodels were achieved in about 0.01~seconds. A high degree of accuracy between the full-order HFS and the surrogate model was demonstrated for a set of 500 new test cases. Close agreement was observed for CO$_2$ saturation and pressure in the storage aquifer and for vertical displacement at the Earth's surface. Results were presented both for individual geomodels and for ensemble statistics (P$_{10}$, P$_{50}$, P$_{90}$ responses over the full set of test cases) at specific observation locations.

We then applied the recurrent R-U-Net model for data assimilation. A rigorous rejection sampling algorithm was used, which required the evaluation of a large number of prior models. Specifically, 500,000 prior geomodels were generated using PCA, and flow predictions were provided using the surrogate model (this assessment would be intractable using HFS). A small number of surface displacement measurements, derived in this case from numerical simulation results on a synthetic `true' model, were used for data assimilation. Significant uncertainty reduction in predictions for both surface displacement and storage-aquifer pressure was observed.

Future work in this area should address a number of topics. Larger and more complicated models, involving uncertainty in a wider range of properties including geomechanical parameters, should be considered. The use of other data assimilation algorithms, with a variety of data types, is also of interest. Because function evaluations with the surrogate model are so fast, methods and assessments (e.g., sensitivity of predictions to different data types or data precision) that would be impractical with full-order numerical models can now be considered. Finally, the use of the general recurrent R-U-Net framework for other coupled multi-physics problems, involving, e.g., some combination of flow, geomechanics, thermal effects and chemical reactions, should be explored.

\section{Code Availability}
The code used in this study is available at https://github.com/soloist96/Recurrent-R-U-Net-for-CO2-Storage.git. 

\section{Acknowledgements}
We are grateful to the Stanford Smart Fields Consortium and to Stanford--Chevron CoRE for partial funding of this work. We thank Pengcheng Fu from the Computational Geosciences Group at Lawrence Livermore National Laboratory for his help with the construction of the coupled flow-geomechanics models. We also acknowledge the Livermore Computing Center and the Stanford Center for Computational Earth \& Environmental Science (CEES) for providing the computational resources used in this work.

%\input{chap3}
% history matching
%\input{chap4}
% conclusions and acknowledgement 
%\input{chap5}
% appendix
%\input{appendix}

%\newpage
%\section*{References}

\bibliographystyle{elsarticle-num-names}
\bibliography{reference}

\begin{thebibliography}{43}
\providecommand{\natexlab}[1]{#1}
\providecommand{\url}[1]{\texttt{#1}}
\providecommand{\urlprefix}{URL }
\expandafter\ifx\csname urlstyle\endcsname\relax
  \providecommand{\doi}[1]{doi:\discretionary{}{}{}#1}\else
  \providecommand{\doi}[1]{doi:\discretionary{}{}{}\begingroup
  \urlstyle{rm}\url{#1}\endgroup}\fi
\providecommand{\bibinfo}[2]{#2}

\bibitem[{Tang et~al.(2020)Tang, Liu, and Durlofsky}]{tang2020jcp}
\bibinfo{author}{M.~Tang}, \bibinfo{author}{Y.~Liu}, \bibinfo{author}{L.~J.
  Durlofsky}, \bibinfo{title}{A deep-learning-based surrogate model for data
  assimilation in dynamic subsurface flow problems}, \bibinfo{journal}{Journal
  of Computational Physics} \bibinfo{volume}{413} (\bibinfo{year}{2020})
  \bibinfo{pages}{109456}.

\bibitem[{Tang et~al.(2021)Tang, Liu, and Durlofsky}]{tang2021deep}
\bibinfo{author}{M.~Tang}, \bibinfo{author}{Y.~Liu}, \bibinfo{author}{L.~J.
  Durlofsky}, \bibinfo{title}{Deep-learning-based surrogate flow modeling and
  geological parameterization for data assimilation in 3D subsurface flow},
  \bibinfo{journal}{Computer Methods in Applied Mechanics and Engineering}
  \bibinfo{volume}{376} (\bibinfo{year}{2021}) \bibinfo{pages}{113636}.

\bibitem[{Vilarrasa et~al.(2010)Vilarrasa, Bolster, Olivella, and
  Carrera}]{vilarrasa2010coupled}
\bibinfo{author}{V.~Vilarrasa}, \bibinfo{author}{D.~Bolster},
  \bibinfo{author}{S.~Olivella}, \bibinfo{author}{J.~Carrera},
  \bibinfo{title}{Coupled hydromechanical modeling of CO$_2$ sequestration in
  deep saline aquifers}, \bibinfo{journal}{International Journal of Greenhouse
  Gas Control} \bibinfo{volume}{4}~(\bibinfo{number}{6}) (\bibinfo{year}{2010})
  \bibinfo{pages}{910--919}.

\bibitem[{Shi et~al.(2013)Shi, Smith, Durucan, and Korre}]{shi2013coupled}
\bibinfo{author}{J.-Q. Shi}, \bibinfo{author}{J.~Smith},
  \bibinfo{author}{S.~Durucan}, \bibinfo{author}{A.~Korre}, \bibinfo{title}{A
  coupled reservoir simulation-geomechanical modelling study of the CO$_2$
  injection-induced ground surface uplift observed at Krechba, In Salah},
  \bibinfo{journal}{Energy Procedia} \bibinfo{volume}{37}
  (\bibinfo{year}{2013}) \bibinfo{pages}{3719--3726}.

\bibitem[{Talebian et~al.(2013)Talebian, Al-Khoury, and
  Sluys}]{talebian2013computational}
\bibinfo{author}{M.~Talebian}, \bibinfo{author}{R.~Al-Khoury},
  \bibinfo{author}{L.~Sluys}, \bibinfo{title}{A computational model for coupled
  multiphysics processes of CO$_2$ sequestration in fractured porous media},
  \bibinfo{journal}{Advances in Water Resources} \bibinfo{volume}{59}
  (\bibinfo{year}{2013}) \bibinfo{pages}{238--255}.

\bibitem[{Li and Laloui(2016)}]{li2016coupled}
\bibinfo{author}{C.~Li}, \bibinfo{author}{L.~Laloui}, \bibinfo{title}{Coupled
  multiphase thermo-hydro-mechanical analysis of supercritical CO$_2$
  injection: Benchmark for the In Salah surface uplift problem},
  \bibinfo{journal}{International Journal of Greenhouse Gas Control}
  \bibinfo{volume}{51} (\bibinfo{year}{2016}) \bibinfo{pages}{394--408}.

\bibitem[{Fuchs et~al.(2019)Fuchs, Espinoza, Lopano, Akono, and
  Werth}]{fuchs2019geochemical}
\bibinfo{author}{S.~J. Fuchs}, \bibinfo{author}{D.~N. Espinoza},
  \bibinfo{author}{C.~L. Lopano}, \bibinfo{author}{A.-T. Akono},
  \bibinfo{author}{C.~J. Werth}, \bibinfo{title}{Geochemical and geomechanical
  alteration of siliciclastic reservoir rock by supercritical CO$_2$-saturated
  brine formed during geological carbon sequestration},
  \bibinfo{journal}{International Journal of Greenhouse Gas Control}
  \bibinfo{volume}{88} (\bibinfo{year}{2019}) \bibinfo{pages}{251--260}.

\bibitem[{Fu et~al.(2021)Fu, Ju, Huang, Settgast, Liu, and Morris}]{futhermo}
\bibinfo{author}{P.~Fu}, \bibinfo{author}{X.~Ju}, \bibinfo{author}{J.~Huang},
  \bibinfo{author}{R.~R. Settgast}, \bibinfo{author}{F.~Liu},
  \bibinfo{author}{J.~P. Morris}, \bibinfo{title}{Thermo-poroelastic responses
  of a pressure-driven fracture in a carbon storage reservoir and the
  implications for injectivity and caprock integrity},
  \bibinfo{journal}{International Journal for Numerical and Analytical Methods
  in Geomechanics} \bibinfo{volume}{45}~(\bibinfo{number}{6})
  (\bibinfo{year}{2021}) \bibinfo{pages}{719--737}.

\bibitem[{Ju et~al.(2021)Ju, Fu, Settegast, and Morris}]{ju2021simple}
\bibinfo{author}{X.~Ju}, \bibinfo{author}{P.~Fu},
  \bibinfo{author}{R.~Settegast}, \bibinfo{author}{J.~Morris},
  \bibinfo{title}{A simple method to simulate thermo-hydro-mechanical processes
  in leakoff-dominated hydraulic fracturing with application to geological
  carbon storage}, \bibinfo{journal}{Earth and Space Science Open Archive}
  (\bibinfo{year}{2021}) \bibinfo{pages}{53}.

\bibitem[{Eshiet and Sheng(2014)}]{eshiet2014investigation}
\bibinfo{author}{K.~Eshiet}, \bibinfo{author}{Y.~Sheng},
  \bibinfo{title}{Investigation of geomechanical responses of reservoirs
  induced by carbon dioxide storage}, \bibinfo{journal}{Environmental Earth
  Sciences} \bibinfo{volume}{71}~(\bibinfo{number}{9}) (\bibinfo{year}{2014})
  \bibinfo{pages}{3999--4020}.

\bibitem[{Rutqvist et~al.(2019)Rutqvist, Rinaldi, Vilarrasa, and
  Cappa}]{rutqvist2019numerical}
\bibinfo{author}{J.~Rutqvist}, \bibinfo{author}{A.~P. Rinaldi},
  \bibinfo{author}{V.~Vilarrasa}, \bibinfo{author}{F.~Cappa},
  \bibinfo{title}{Numerical geomechanics studies of geological carbon storage
  (GCS)}, in: \bibinfo{booktitle}{Science of Carbon Storage in Deep Saline
  Formations}, \bibinfo{publisher}{Elsevier}, \bibinfo{pages}{237--252},
  \bibinfo{year}{2019}.

\bibitem[{Kim et~al.(2011)Kim, Tchelepi, and Juanes}]{kim2011stability}
\bibinfo{author}{J.~Kim}, \bibinfo{author}{H.~A. Tchelepi},
  \bibinfo{author}{R.~Juanes}, \bibinfo{title}{Stability and convergence of
  sequential methods for coupled flow and geomechanics: Fixed-stress and
  fixed-strain splits}, \bibinfo{journal}{Computer Methods in Applied Mechanics
  and Engineering} \bibinfo{volume}{200}~(\bibinfo{number}{13-16})
  (\bibinfo{year}{2011}) \bibinfo{pages}{1591--1606}.

\bibitem[{Florez and Gildin(2019)}]{florez2019model}
\bibinfo{author}{H.~Florez}, \bibinfo{author}{E.~Gildin},
  \bibinfo{title}{Model-order reduction of coupled flow and geomechanics in
  ultra-low permeability ULP reservoirs}, in: \bibinfo{booktitle}{SPE Reservoir
  Simulation Conference}, \bibinfo{organization}{Society of Petroleum
  Engineers}, \bibinfo{year}{2019}.

\bibitem[{Jin et~al.(2020)Jin, Garipov, Volkov, and Durlofsky}]{jin2020reduced}
\bibinfo{author}{Z.~L. Jin}, \bibinfo{author}{T.~Garipov},
  \bibinfo{author}{O.~Volkov}, \bibinfo{author}{L.~J. Durlofsky},
  \bibinfo{title}{Reduced-order modeling of coupled flow and quasistatic
  geomechanics}, \bibinfo{journal}{SPE Journal}
  \bibinfo{volume}{25}~(\bibinfo{number}{01}) (\bibinfo{year}{2020})
  \bibinfo{pages}{326--346}.

\bibitem[{Jiang et~al.(2021)Jiang, Tahmasebi, and Mao}]{jiang2021deep}
\bibinfo{author}{Z.~Jiang}, \bibinfo{author}{P.~Tahmasebi},
  \bibinfo{author}{Z.~Mao}, \bibinfo{title}{Deep residual U-Net convolution
  neural networks with autoregressive strategy for fluid flow predictions in
  large-scale geosystems}, \bibinfo{journal}{Advances in Water Resources}
  (\bibinfo{year}{2021}) \bibinfo{pages}{103878}.

\bibitem[{Mo et~al.(2019)Mo, Zhu, Zabaras, Shi, and Wu}]{mo2019deep}
\bibinfo{author}{S.~Mo}, \bibinfo{author}{Y.~Zhu},
  \bibinfo{author}{N.~Zabaras}, \bibinfo{author}{X.~Shi},
  \bibinfo{author}{J.~Wu}, \bibinfo{title}{Deep convolutional encoder-decoder
  networks for uncertainty quantification of dynamic multiphase flow in
  heterogeneous media}, \bibinfo{journal}{Water Resources Research}
  \bibinfo{volume}{55}~(\bibinfo{number}{1}) (\bibinfo{year}{2019})
  \bibinfo{pages}{703--728}.

\bibitem[{Wen et~al.(2021)Wen, Tang, and Benson}]{wen2021towards}
\bibinfo{author}{G.~Wen}, \bibinfo{author}{M.~Tang}, \bibinfo{author}{S.~M.
  Benson}, \bibinfo{title}{Towards a predictor for CO$_2$ plume migration using
  deep neural networks}, \bibinfo{journal}{International Journal of Greenhouse
  Gas Control} \bibinfo{volume}{105} (\bibinfo{year}{2021})
  \bibinfo{pages}{103223}.

\bibitem[{Gonz{\'a}lez-Nicol{\'a}s et~al.(2015)Gonz{\'a}lez-Nicol{\'a}s,
  Ba{\`u}, and Alzraiee}]{gonzalez2015detection}
\bibinfo{author}{A.~Gonz{\'a}lez-Nicol{\'a}s}, \bibinfo{author}{D.~Ba{\`u}},
  \bibinfo{author}{A.~Alzraiee}, \bibinfo{title}{Detection of potential leakage
  pathways from geological carbon storage by fluid pressure data assimilation},
  \bibinfo{journal}{Advances in Water Resources} \bibinfo{volume}{86}
  (\bibinfo{year}{2015}) \bibinfo{pages}{366--384}.

\bibitem[{Jung et~al.(2015)Jung, Zhou, and Birkholzer}]{jung2015detection}
\bibinfo{author}{Y.~Jung}, \bibinfo{author}{Q.~Zhou}, \bibinfo{author}{J.~T.
  Birkholzer}, \bibinfo{title}{On the detection of leakage pathways in
  geological CO$_2$ storage systems using pressure monitoring data: Impact of
  model parameter uncertainties}, \bibinfo{journal}{Advances in Water
  Resources} \bibinfo{volume}{84} (\bibinfo{year}{2015})
  \bibinfo{pages}{112--124}.

\bibitem[{Cameron et~al.(2016)Cameron, Durlofsky, and Benson}]{cameron2016use}
\bibinfo{author}{D.~A. Cameron}, \bibinfo{author}{L.~J. Durlofsky},
  \bibinfo{author}{S.~M. Benson}, \bibinfo{title}{Use of above-zone pressure
  data to locate and quantify leaks during carbon storage operations},
  \bibinfo{journal}{International Journal of Greenhouse Gas Control}
  \bibinfo{volume}{52} (\bibinfo{year}{2016}) \bibinfo{pages}{32--43}.

\bibitem[{Chen et~al.(2020)Chen, Harp, Lu, and Pawar}]{chen2020reducing}
\bibinfo{author}{B.~Chen}, \bibinfo{author}{D.~R. Harp},
  \bibinfo{author}{Z.~Lu}, \bibinfo{author}{R.~J. Pawar},
  \bibinfo{title}{Reducing uncertainty in geologic CO$_2$ sequestration risk
  assessment by assimilating monitoring data}, \bibinfo{journal}{International
  Journal of Greenhouse Gas Control} \bibinfo{volume}{94}
  (\bibinfo{year}{2020}) \bibinfo{pages}{102926}.

\bibitem[{Liu and Grana(2020)}]{liu2020petrophysical}
\bibinfo{author}{M.~Liu}, \bibinfo{author}{D.~Grana},
  \bibinfo{title}{Petrophysical characterization of deep saline aquifers for
  CO$_2$ storage using ensemble smoother and deep convolutional autoencoder},
  \bibinfo{journal}{Advances in Water Resources} \bibinfo{volume}{142}
  (\bibinfo{year}{2020}) \bibinfo{pages}{103634}.

\bibitem[{de~la Torre~Guzman et~al.(2014)de~la Torre~Guzman, Babaei, Shi,
  Korre, and Durucan}]{de2014coupled}
\bibinfo{author}{J.~de~la Torre~Guzman}, \bibinfo{author}{M.~Babaei},
  \bibinfo{author}{J.-Q. Shi}, \bibinfo{author}{A.~Korre},
  \bibinfo{author}{S.~Durucan}, \bibinfo{title}{Coupled flow-geomechanical
  performance assessment of CO$_2$ storage sites using the ensemble Kalman
  filter}, \bibinfo{journal}{Energy Procedia} \bibinfo{volume}{63}
  (\bibinfo{year}{2014}) \bibinfo{pages}{3475--3482}.

\bibitem[{Jahandideh et~al.(2021)Jahandideh, Hakim-Elahi, and
  Jafarpour}]{jahandideh2021inference}
\bibinfo{author}{A.~Jahandideh}, \bibinfo{author}{S.~Hakim-Elahi},
  \bibinfo{author}{B.~Jafarpour}, \bibinfo{title}{Inference of rock flow and
  mechanical properties from injection-induced microseismic events during
  geologic CO$_2$ storage}, \bibinfo{journal}{International Journal of
  Greenhouse Gas Control} \bibinfo{volume}{105} (\bibinfo{year}{2021})
  \bibinfo{pages}{103206}.

\bibitem[{Wilschut et~al.(2011)Wilschut, Peters, Visser, Fokker, and van
  Hooff}]{wilschut2011joint}
\bibinfo{author}{F.~Wilschut}, \bibinfo{author}{E.~Peters},
  \bibinfo{author}{K.~Visser}, \bibinfo{author}{P.~A. Fokker},
  \bibinfo{author}{P.~van Hooff}, \bibinfo{title}{Joint history matching of
  well data and surface subsidence observations using the ensemble Kalman
  filter: a field study}, in: \bibinfo{booktitle}{SPE Reservoir Simulation
  Symposium}, \bibinfo{organization}{Society of Petroleum Engineers},
  \bibinfo{year}{2011}.

\bibitem[{Jha et~al.(2015)Jha, Bottazzi, Wojcik, Coccia, Bechor, McLaughlin,
  Herring, Hager, Mantica, and Juanes}]{jha2015reservoir}
\bibinfo{author}{B.~Jha}, \bibinfo{author}{F.~Bottazzi},
  \bibinfo{author}{R.~Wojcik}, \bibinfo{author}{M.~Coccia},
  \bibinfo{author}{N.~Bechor}, \bibinfo{author}{D.~McLaughlin},
  \bibinfo{author}{T.~Herring}, \bibinfo{author}{B.~H. Hager},
  \bibinfo{author}{S.~Mantica}, \bibinfo{author}{R.~Juanes},
  \bibinfo{title}{Reservoir characterization in an underground gas storage
  field using joint inversion of flow and geodetic data},
  \bibinfo{journal}{International Journal for Numerical and Analytical Methods
  in Geomechanics} \bibinfo{volume}{39}~(\bibinfo{number}{14})
  (\bibinfo{year}{2015}) \bibinfo{pages}{1619--1638}.

\bibitem[{Zoccarato et~al.(2016)Zoccarato, Ba{\`u}, Ferronato, Gambolati,
  Alzraiee, and Teatini}]{zoccarato2016data}
\bibinfo{author}{C.~Zoccarato}, \bibinfo{author}{D.~Ba{\`u}},
  \bibinfo{author}{M.~Ferronato}, \bibinfo{author}{G.~Gambolati},
  \bibinfo{author}{A.~Alzraiee}, \bibinfo{author}{P.~Teatini},
  \bibinfo{title}{Data assimilation of surface displacements to improve
  geomechanical parameters of gas storage reservoirs},
  \bibinfo{journal}{Journal of Geophysical Research: Solid Earth}
  \bibinfo{volume}{121}~(\bibinfo{number}{3}) (\bibinfo{year}{2016})
  \bibinfo{pages}{1441--1461}.

\bibitem[{Tang(2018)}]{tang2018history}
\bibinfo{author}{M.~Tang}, \bibinfo{title}{History matching production and
  displacement data using derivative-free optimization}, Master's thesis,
  \bibinfo{school}{Stanford University}, \bibinfo{year}{2018}.

\bibitem[{Alghamdi et~al.(2020)Alghamdi, Hesse, Chen, and
  Ghattas}]{alghamdi2020bayesian}
\bibinfo{author}{A.~Alghamdi}, \bibinfo{author}{M.~A. Hesse},
  \bibinfo{author}{J.~Chen}, \bibinfo{author}{O.~Ghattas},
  \bibinfo{title}{Bayesian poroelastic aquifer characterization from InSAR
  surface deformation data. Part I: Maximum a posteriori estimate},
  \bibinfo{journal}{Water Resources Research}
  \bibinfo{volume}{56}~(\bibinfo{number}{10}) (\bibinfo{year}{2020})
  \bibinfo{pages}{e2020WR027391}.

\bibitem[{Alghamdi et~al.(2021)Alghamdi, Hesse, Chen, Villa, and
  Ghattas}]{alghamdi2021bayesian}
\bibinfo{author}{A.~Alghamdi}, \bibinfo{author}{M.~Hesse},
  \bibinfo{author}{J.~Chen}, \bibinfo{author}{U.~Villa},
  \bibinfo{author}{O.~Ghattas}, \bibinfo{title}{Bayesian poroelastic aquifer
  characterization from InSAR surface deformation data. Part II: Quantifying
  the uncertainty}, \bibinfo{journal}{arXiv preprint arXiv:2102.04577 (2021)} .

\bibitem[{Biot(1941)}]{biot1941general}
\bibinfo{author}{M.~A. Biot}, \bibinfo{title}{General theory of
  three-dimensional consolidation}, \bibinfo{journal}{Journal of Applied
  Physics} \bibinfo{volume}{12}~(\bibinfo{number}{2}) (\bibinfo{year}{1941})
  \bibinfo{pages}{155--164}.

\bibitem[{Coussy(2004)}]{coussy2004poromechanics}
\bibinfo{author}{O.~Coussy}, \bibinfo{title}{Poromechanics},
  \bibinfo{publisher}{John Wiley \& Sons}, \bibinfo{year}{2004}.

\bibitem[{Settgast et~al.(2017)Settgast, Fu, Walsh, White, Annavarapu, and
  Ryerson}]{settgast2017fully}
\bibinfo{author}{R.~R. Settgast}, \bibinfo{author}{P.~Fu},
  \bibinfo{author}{S.~D. Walsh}, \bibinfo{author}{J.~A. White},
  \bibinfo{author}{C.~Annavarapu}, \bibinfo{author}{F.~J. Ryerson},
  \bibinfo{title}{A fully coupled method for massively parallel simulation of
  hydraulically driven fractures in 3-dimensions},
  \bibinfo{journal}{International Journal for Numerical and Analytical Methods
  in Geomechanics} \bibinfo{volume}{41}~(\bibinfo{number}{5})
  (\bibinfo{year}{2017}) \bibinfo{pages}{627--653}.

\bibitem[{Ju et~al.(2020)Ju, Liu, Fu, White, Settgast, and Morris}]{ju2020gas}
\bibinfo{author}{X.~Ju}, \bibinfo{author}{F.~Liu}, \bibinfo{author}{P.~Fu},
  \bibinfo{author}{M.~D. White}, \bibinfo{author}{R.~R. Settgast},
  \bibinfo{author}{J.~P. Morris}, \bibinfo{title}{Gas production from hot water
  circulation through hydraulic fractures in methane hydrate-bearing sediments:
  THC-coupled simulation of production mechanisms}, \bibinfo{journal}{Energy \&
  Fuels} \bibinfo{volume}{34}~(\bibinfo{number}{4}) (\bibinfo{year}{2020})
  \bibinfo{pages}{4448--4465}.

\bibitem[{B{\"u}rgmann et~al.(2000)B{\"u}rgmann, Rosen, and
  Fielding}]{burgmann2000synthetic}
\bibinfo{author}{R.~B{\"u}rgmann}, \bibinfo{author}{P.~A. Rosen},
  \bibinfo{author}{E.~J. Fielding}, \bibinfo{title}{Synthetic aperture radar
  interferometry to measure Earth’s surface topography and its deformation},
  \bibinfo{journal}{Annual Review of Earth and Planetary Sciences}
  \bibinfo{volume}{28}~(\bibinfo{number}{1}) (\bibinfo{year}{2000})
  \bibinfo{pages}{169--209}.

\bibitem[{Remy et~al.(2009)Remy, Boucher, and Wu}]{remy2009applied}
\bibinfo{author}{N.~Remy}, \bibinfo{author}{A.~Boucher},
  \bibinfo{author}{J.~Wu}, \bibinfo{title}{Applied geostatistics with SGeMS: a
  user's guide}, \bibinfo{publisher}{Cambridge University Press},
  \bibinfo{year}{2009}.

\bibitem[{Ronneberger et~al.(2015)Ronneberger, Fischer, and
  Brox}]{ronneberger2015u}
\bibinfo{author}{O.~Ronneberger}, \bibinfo{author}{P.~Fischer},
  \bibinfo{author}{T.~Brox}, \bibinfo{title}{U-net: Convolutional networks for
  biomedical image segmentation}, in: \bibinfo{booktitle}{International
  Conference on Medical Image Computing and Computer-assisted Intervention},
  \bibinfo{organization}{Springer}, \bibinfo{pages}{234--241},
  \bibinfo{year}{2015}.

\bibitem[{He et~al.(2016)He, Zhang, Ren, and Sun}]{he2016deep}
\bibinfo{author}{K.~He}, \bibinfo{author}{X.~Zhang}, \bibinfo{author}{S.~Ren},
  \bibinfo{author}{J.~Sun}, \bibinfo{title}{Deep residual learning for image
  recognition}, in: \bibinfo{booktitle}{Proceedings of the IEEE Conference on
  Computer Vision and Pattern Recognition}, \bibinfo{pages}{770--778},
  \bibinfo{year}{2016}.

\bibitem[{Xingjian et~al.(2015)Xingjian, Chen, Wang, Yeung, Wong, and
  Woo}]{xingjian2015convolutional}
\bibinfo{author}{S.~Xingjian}, \bibinfo{author}{Z.~Chen},
  \bibinfo{author}{H.~Wang}, \bibinfo{author}{D.-Y. Yeung},
  \bibinfo{author}{W.-K. Wong}, \bibinfo{author}{W.-C. Woo},
  \bibinfo{title}{Convolutional LSTM network: A machine learning approach for
  precipitation nowcasting}, in: \bibinfo{booktitle}{Advances in Neural
  Information Processing Systems}, \bibinfo{pages}{802--810},
  \bibinfo{year}{2015}.

\bibitem[{Kingma and Ba(2014)}]{kingma2014adam}
\bibinfo{author}{D.~P. Kingma}, \bibinfo{author}{J.~Ba}, \bibinfo{title}{Adam:
  a method for stochastic optimization}, \bibinfo{journal}{arXiv preprint
  arXiv:1412.6980 (2014)} .

\bibitem[{Altunin(1968)}]{altunin1968thermophysical}
\bibinfo{author}{V.~Altunin}, \bibinfo{title}{Thermophysical Properties of
  Carbon Dioxide}, \bibinfo{publisher}{Wellingborough, Collets},
  \bibinfo{year}{1968}.

\bibitem[{Wagner and Kretzschmar(2008)}]{wagner2008iapws}
\bibinfo{author}{W.~Wagner}, \bibinfo{author}{H.-J. Kretzschmar},
  \bibinfo{title}{IAPWS industrial formulation 1997 for the thermodynamic
  properties of water and steam}, \bibinfo{journal}{International Steam Tables:
  Properties of Water and Steam based on the Industrial Formulation IAPWS-IF97}
   (\bibinfo{year}{2008}) \bibinfo{pages}{7--150}.

\bibitem[{Liu and Durlofsky(2021)}]{liu20203d}
\bibinfo{author}{Y.~Liu}, \bibinfo{author}{L.~J. Durlofsky}, \bibinfo{title}{3D
  CNN-PCA: A deep-learning-based parameterization for complex geomodels},
  \bibinfo{journal}{Computers \& Geosciences} \bibinfo{volume}{148}
  (\bibinfo{year}{2021}) \bibinfo{pages}{104676}.

\end{thebibliography}

\end{document}